\documentclass{article}

\PassOptionsToPackage{numbers, compress}{natbib}

\usepackage[main, final]{neurips_2026}

\usepackage{makecell}
\usepackage[utf8]{inputenc} %
\usepackage[T1]{fontenc}    %
\usepackage{hyperref}       %
\usepackage{url}            %
\usepackage{booktabs}       %
\usepackage{amsfonts}       %
\usepackage{nicefrac}       %
\usepackage{microtype}      %
\usepackage{wrapfig}
\usepackage[dvipsnames]{xcolor}         %
\usepackage[ruled]{algorithm2e}
\usepackage{graphicx}
\usepackage{subcaption}
\usepackage[rightcaption]{sidecap}
\usepackage{multirow}
\sidecaptionvpos{figure}{t}

\usepackage[acronym]{glossaries}
\usepackage{bbm} %
\makeglossaries

\newacronym{pc}{PC}{Probabilistic Circuit}
\newacronym{rg}{RG}{Region Graph}

\newacronym{pic}{PIC}{Probabilitic Integral Circuit}
\newacronym{sos}{SOS}{Sum of Square}
\newacronym{em}{EM}{Expectation Maximisation}
\newacronym{flops}{FLOPs}{Floating Point Operations}
\newacronym{bpd}{BPD}{Bits per Dimension}
\newacronym{llm}{LLM}{Large Language Model}

\newacronym{mar}{MAR}{Marginalisation}
\newacronym{map}{MAP}{Maximum A Posteriori}

\newcommand{\indepparadox}{independent patch conundrum\xspace}

\usepackage{macros}

\usepackage[capitalise,nameinlink]{cleveref}

\renewcommand{\paragraph}[1]{\textbf{#1}}

\hypersetup{
colorlinks=true,linkcolor=teal,citecolor=brown,urlcolor=orange
}

\title{Patching Up Circuit Learning via Parameter Sharing}

\author{%
    Charles Bricout \\
  University of Edinburgh\\
  \texttt{c.bricout@sms.ed.ac.uk} 
   \And 
   Gennaro Gala \\
  Independent Researcher
 \And
  Viacheslav Borovitskiy \\
  University of Edinburgh
   \And
  Antonio Vergari \\
  University of Edinburgh
}

\begin{document}

\maketitle

\begin{abstract}
Probabilistic circuits (PCs) have emerged 
as generative models guaranteeing tractable probabilistic inference.
However, tractability comes at the price of larger model sizes than in other generative models.
As such, to scale PCs to high-dimensional data such as images, previous work has devised several parameter sharing techniques as well as fast evaluation protocols that consider learning PCs on \textit{independent} image patches.
In this work, we revisit this practice to highlight a surprising learnability issue: models that assume independence counterintuitively can yield better likelihoods than models that do not.
To break this apparent bottleneck and to scale PCs further, we introduce a \textit{recursive parameter sharing construction} (\repa) that fosters efficiency without assuming independence.
We show empirically that \repa can greatly reduce time and memory consumption while yielding competitive likelihoods.
Inspired by these results, we further generalize the approach to data modalities that do not inherently exhibit patch-structure (\nopa), such as tabular data, showing similar efficiency improvements, with marginal changes in likelihoods and sample quality on seven tabular datasets.
Code available at \href{https://github.com/april-tools/RePA}{https://github.com/april-tools/RePA}.
\end{abstract}

\section{Introduction}\label{sec:intro}

Probabilistic circuits (PCs) \citep{darwiche2003differential,vergari2019tractable,choi2020pc} have emerged as a staple framework for tractable probabilistic modeling, as
they can guarantee the exact computation of probabilistic  inference routines such as sampling, and computing marginals or conditionals in linear time. 
As such, they have been successfully used in a number of applications such as lossless compression \citep{liu2022lossless}, generative modeling of biomedical data \citep{dang_tractable_2023}, learning and generation under symbolic constraints \citep{ahmed2022semantic, loconte2023turn,zhang2023tractable}, multi-token prediction in LLMs \citep{grivas_fast_2025}, time-series analysis \citep{thoma2021recowns,yu2021whittle,klotergens2026probabilistic}, and recently they have been shown to be competitive for tabular data generation \citep{scassola_sobering_2026}. 
\begin{figure*}[h]
    \centering
\includegraphics[width=0.32\linewidth]{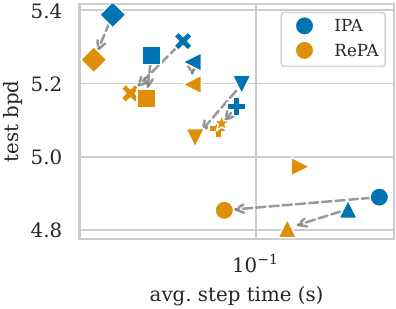}
\hfill
\includegraphics[width=0.32\linewidth]{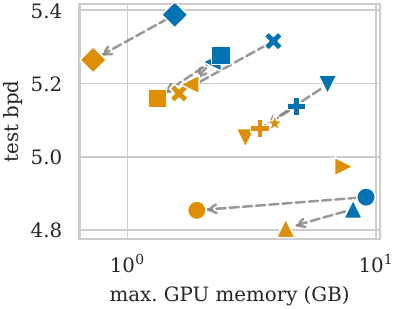}
\hfill
\includegraphics[width=.32\linewidth]{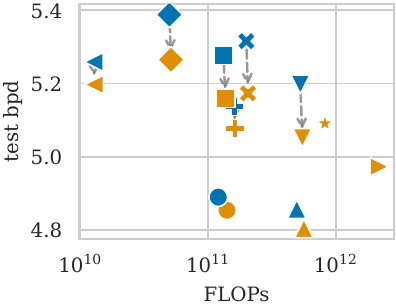}
\caption{\textbf{\repa delivers faster, smaller and more accurate circuits compared to PCs that assume patch independence (\indepatch)}. This is shown here in terms of average step time (seconds), GPU memory, FLOPs and bpd (i.e. normalized negative log-likelihoods) on ImageNet32 for several tensorized PC architectures.
\repa markers (yellow) that have no corresponding \indepatch markers (blue) are those for which \indepatch could not be trained on 40GB GPUs in under 2 days.
See \cref{app:fig:bpd_vs_efficiency} for similar trends on CelebA and ImageNet64 and the precise PC configuration associated to each marker.
}
    \label{fig:fig-1}
\end{figure*}
However, the structural constraints required for tractability imply PCs cannot easily incorporate the standard inductive biases (e.g., use convolutions or skip connections \citep{lecun2015deep,he_deep_2016}) employed by other generative models such as normalizing flows and diffusion models \citep{papamakarios2021flows, yang2023diffusion}.
As a consequence, to learn PCs of reasonable quality, one needs up to an order of magnitude more parameters than for their intractable counterparts such as flow and diffusion models \citep{lipman_flow_2023,yang2023diffusion,scassola_sobering_2026, loconte2025what}, which is possible by \textit{overparameterizing} the circuit \citep{peharz2019ratspn,peharz2020einsum,mari2023unifying,loconte2025what}. 

Not surprisingly, this becomes critical for high-dimensional data like images: scaling PCs to billions of parameters remains an open challenge, demanding steep trade-offs between model accuracy and computational resources \citep{peharz2020einsum,anji2022scaling-pcs-lvd,liu2023understanding,zhang_scaling_2025}.
To navigate these trade-offs, several recent papers construct smaller PCs over non-overlapping image \textit{patches} \citep{anji2022scaling-pcs-lvd,liu2023understanding,liu2024scaling,zhang_scaling_2025}.
We call this the \textit{independent patch assumption} (\indepatch).
The joint distribution over a whole image is then recovered by multiplying the probabilities of the disjoint patches of the image, \textit{as they are assumed to be independent}.
This accelerates experimentation under the seemingly safe presumption that model quality, measured through likelihood or sampling quality, will strictly increase once the patch independence assumption is dropped.
In this paper, we study this phenomenon and advance the research into scalable PCs by introducing  forms of \textit{parameter sharing} that can be applied to data modalities showing a patch-like structure as well as to those without it.
Specifically we make the following contributions.

First, \textbf{(C1)} we show that the aforementioned presumption
fails in practice: PCs trained on smaller independent patches can achieve equal or better likelihoods than those trained on larger patches, despite the latter being theoretically more expressive.  
We call this phenomenon the \textit{\indepparadox} (\cref{sec:indepatch}).
Similarly to previous works, this highlights how overparameterizing PCs can hinder learning \citep{liu2023understanding,dang2022sparse}.
Differently from previous works, we identify an expressiveness-learnability trade-off that depends on the domain dimensionality rather than just the number of parameters in a circuit, and we investigate principled ways to circumvent this specific bottleneck.

To this end, we build a new PC architecture \textbf{(C2)} that systematically exploits parameter sharing throughout all layers.
Specifically, we introduce \textit{recursive parameter sharing} (\repa) as a way to share sub-circuits defined over a hierarchy of coarser patches (\cref{sec:repa}).
It can naturally be trained end-to-end while drastically reducing parameter count.  
\repa generalizes the form of parameter sharing that arises when patches are assumed to be independent \citep{liu2023understanding} and is orthogonal to other forms of parameter sharing that have been used in training more expressive circuits \citep{gala2024pic,gala2024scaling,zhang_scaling_2025}.

\looseness=-1
Furthermore, we \textbf{(C3)} execute a systematic evaluation (\cref{sec:experiments}) of how \indepatch and \repa affect not only the quality of PCs as density estimators, but also their efficiency in terms of FLOPs, time and memory consumption, as illustrated in~\cref{fig:fig-1}.
Going beyond image data, we extend \repa to work on data modalities that do not exhibit any patch structure, such as tabular data, by exploiting the hierarchical structure of the PC \textbf{(C4)}. This technique, called \textit{no patch sharing} (\nopa), yields more efficient models in terms of memory and time at the cost of marginal changes in density estimation.

Finally, we \textbf{(C5)} evaluate the sample quality on tabular data and show that we get a minimal decrease in sample quality on tabular data when using \nopa.
Overall, we provide a stepping stone towards more efficient PC architectures and highlight that there is still much to be done to realize expressive and tractable probabilistic models.

\section{How to build and learn probabilistic circuits}
\label{sec:background-pc}

\paragraph{Notation.} We denote random variables with uppercase letters, e.g., $X$, and their realizations in lowercase, e.g., $x$.
Vectorized sets of variables and their realizations are denoted in bold, e.g., ${\X}=X_1,\ldots,X_n$ and $\x=x_1,\ldots,x_N$,
as well as matrices, e.g., $\A, \B, \W$.
For simplicity, we consider square images, each represented by $N=S\times S\times C$ variables, where $S$ is their height and width and $C$ the number of channels.
Patches are non-overlapping sub-images of size $D\times D\times C$ and $L=\nicefrac{N}{(D^2C)}$ denotes the number of patches in an image.

\looseness=-1
\paragraph{Probabilistic circuits.} 
A PC $c$ over variables $\X$ is a parameterized computational graph encoding a  probability mass function $c(\X)$ and comprising three fundamental unit types: \textit{input}, \textit{sum} and \textit{product} \citep{darwiche2003differential,vergari2019tractable,choi2020pc}. 
An input unit represents a tractable distribution $f_u(\X_u; \boldsymbol\theta_u)$ parameterized by $\boldsymbol\theta_u$.
For an input unit (or any other type of unit $u$), the subset of variables $\X_u \subset \X$ it operates upon is called its \textit{scope} and is denoted by $\scope(u)=\X_u$. 
Sum and product units operate over the outputs of other units that we will call \textit{inputs} of the unit, denoted by $\inp(u)$. 
A sum unit $s$ computes the convex combination of its inputs, i.e., $\sum_{v \in \inp(s)} w_{s,v}\cdot v(\X)$ where $\sum_{v \in \inp(s)}w_{s,v}=1$ and $w_{s,v}>0$ are learnable weights; and a product unit $p$ computes the product of its inputs, i.e., $\prod_{v\in\inp(p)}v(\X)$.
For image data, we consider categorical distributions as input distributions. 
The learnable parameters of a circuit $c$, denoted as $\boldsymbol\Omega=\{\{\boldsymbol\theta_u\}_u, \{w_{s,v}\}_{s,v}\}$, are all the parameters for input units $u$ and sum units $s$ in $c$.

The signature property of PCs is their ability to guarantee  \emph{tractable} probabilistic inference, i.e. computing probabilistic quantities of interest on the model \emph{exactly} and in \emph{polynomial time} \citep{vergari2019tractable}.
This is done by ensuring their computational graphs abide by certain \textit{structural properties}.
For example, to enable tractable sampling and the computation of every marginal and conditional in a single feedforward pass of the graph, all sum units in the PC need to be \textit{smooth} and all the products \textit{decomposable}. 
A sum unit $s$ is smooth if all its inputs have identical scopes: $\scope(u)=\scope(v) \; \forall u,v \in \inp(s)$.
A product unit $p$ is decomposable if all its inputs have disjoint scopes:
 $\scope(u)\cap\scope(v)=\emptyset \; \forall u,v \in \inp(p), u\neq v$.
Under these properties, product units encode local independence assumptions, while sum units introduce correlations between variables, as they define mixture models \citep{peharz2015theoretical,peharz_latent_2017,loconte2024subtractive}.

\begin{figure*}[!t]
    \centering
    \begin{subfigure}[t]{0.18\linewidth}
            \includegraphics[width=\linewidth]{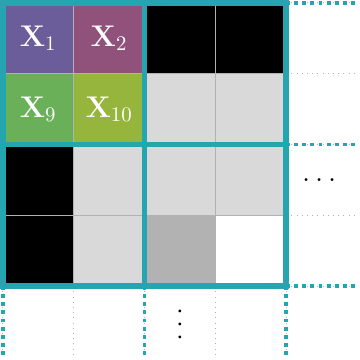}
            \caption{}\label{fig:rg-to-pc-a}
    \end{subfigure}
    \hfill
    \begin{subfigure}[t]{.22\linewidth}
        \includegraphics[width=\linewidth]{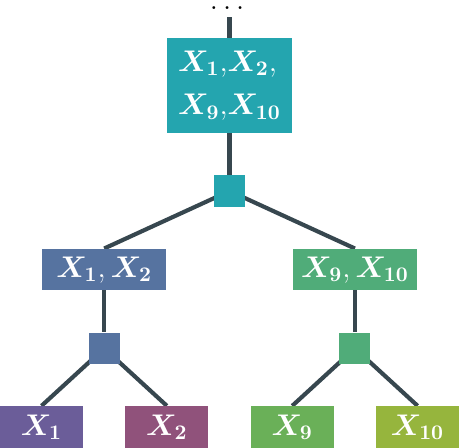}
        \caption{}\label{fig:rg-to-pc-b}
    \end{subfigure}
    \hfill
    \begin{subfigure}[t]{.28\linewidth}
        \includegraphics[width=\linewidth]{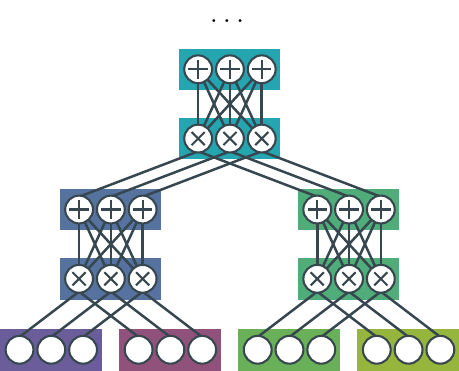}
        \caption{}\label{fig:rg-to-pc-c}
    \end{subfigure}
    \hfill
    \begin{subfigure}[t]{.26\linewidth}
        \includegraphics[width=\linewidth]{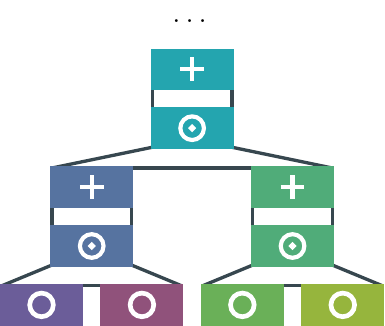}
        \caption{}\label{fig:rg-to-pc-d}
    \end{subfigure}
\caption{
\textbf{Example of a RG and a PC} for  a $2\times2$ patch of an $8\times8$ image (\ref{fig:rg-to-pc-a}) where the RG is a binary quad tree (\ref{fig:rg-to-pc-b}) resulting in the overparameterized ($K=3$) PC with CP-T layers (\ref{fig:rg-to-pc-c}). Abstracting away the units, we obtain a layer notation with input, Hadamard and sum layers (\ref{fig:rg-to-pc-d}).
RG regions and partitions are mapped to the corresponding  sum and product layers in the PC by color. 
}
    \label{fig:rg-to-pc}
\end{figure*}

We follow previous works considering \textit{tensorized} PCs \citep{vergari2019visualizing,peharz2019ratspn,peharz2020einsum,mari2023unifying,loconte2025what,zhang_scaling_2025}, i.e., circuits where units are grouped into \textit{layers} to exploit GPU parallelism.
As \citet{loconte2025what} showed, it is possible to abstract many PC algorithms to build smooth and decomposable tensorized PCs into three stages: (i) creating  a hierarchical partitioning of the input variables, also called a \emph{region graph (RG)} \citep{dennis2012learning}, (ii) specifying a \emph{level of overparameterization}, and (iii) choosing a \emph{sum-product layer} type.

\setlength{\intextsep}{3pt}
\begin{wrapfigure}{r}{0.3\textwidth}
  \begin{center}
    \includegraphics[width=0.3\textwidth]{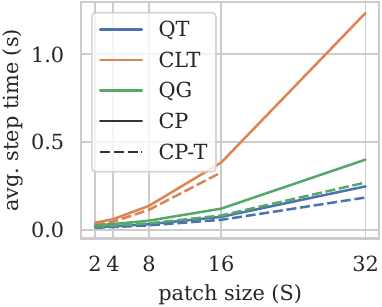}
  \end{center}
  \caption{Training time for different RGs with $K=512$ on images of increasing size.}
  \label{fig:rg_step_time}
\end{wrapfigure}
\paragraph{Region graphs} (RGs) enforce smoothness and decomposability by design by providing a hierarchical partitioning of the input variables in the form of a directed acyclic graph of variable scopes, called \textit{regions}, which are recursively \textit{partitioned} into smaller ones---see~\cref{fig:rg-to-pc}. 
Each region and its corresponding partitions are then mapped to a tensorized sum-product layer comprising many computational units \citep{mari2023unifying,loconte2025what}. 
Many RGs have been proposed in the literature, but here we will focus on \textit{modality-aware} RGs for images \citep{loconte2025what} and \textit{learnable} RGs.

The first category comprises binary quad trees (QTs) \citep{mari2023unifying} that recursively split an image into two axis-aligned sub-images; and quad graphs (QGs) that do the same but cache previously partitioned regions.
For a fixed image size, QTs and QGs are uniquely defined. 
On the other hand, it is possible to learn a RG from a specific dataset, as it is the case for Chow-Liu tree (CLT) RGs \citep{dang2021strudel} which use the Chow-Liu algorithm \citep{chow1968clts}.
Thanks to their ability to capture data distribution, CLTs have been shown to yield PCs with better likelihoods \citep{garg2025bitblasting} and have proven to be a good choice for tabular data \citep{scassola_sobering_2026}.
However, at the same time they require a considerable computational effort to be learned \citep{anji2022scaling-pcs-lvd}.
For example, they cannot be learned directly on ImageNet, requiring approximations \citep{severo2025enhancing}.
Importantly, in contrast to QTs and QGs, they come  with an irregular structure that \textit{is not patch-aligned}.
Furthermore, CLTs are harder to scale.
Indeed, as shown in \cref{fig:rg_step_time}, CLT RGs yield deeper and thus more costly circuits, compared to QTs and QGs. 
As we scale to images of size $64\times64\times3$ they become hard to fit in memory for a circuit defined over the whole image.

\paragraph{Overparameterization.} One could instantiate a single unit for every partition and region in the RG and build a valid PC this way.
However, this would result in limited expressiveness \citep{vergari2015simplifying}.
To avoid this, it is customary to create layers of $K$ parallel units \citep{peharz2019ratspn}.
This layer design allows using GPU acceleration to dramatically speed up training and inference \citep{peharz2020einsum,liu2024scaling}.
Generally, PCs on images of size $32\times32\times3$ and larger use $K$ in the order of $10^3$ \cite{gala2024pic, loconte_sum_2025,liu2024scaling}.
Although larger values may be desirable, they tend to hit the memory bottleneck of common commercial GPUs \citep{zhang_scaling_2025}.

\paragraph{Sum-product layers} are compositions of tensorized layers $\layer(\x)\in[0,1]^{K}$, each comprising $K$ units that correspond to partitions and regions in a RG \citep{loconte2025what}.
Each layer has $H$ other layers $\layer_i$ as inputs, with $i=1,\ldots H$ (\cref{fig:rg-to-pc}). 
We focus on two layers that have been shown to yield the most competitive PCs in terms of likelihood: 
(i) CP \citep{mari2023unifying}, 
and (ii) CP-T \citep{liu2021regularization}
A CP layer encodes $\layer(\x) =\bigodot_{i=1}^{H} \W_i \layer_i(\x_i)$ where $\odot$ denotes the {Hadamard} (elementwise) vector product and $\W_i$ are matrices obtained by concatenating all the weights $w$ associated to the $K$ sum units in the $i$-th input sum layer (\cref{fig:rg-to-pc}).
CP-T encodes $\layer(\x) =\W \bigodot_{i=1}^{H} \layer_i(\x_i) $.

\section{Assuming patch independence and the expressiveness conundrum}
\label{sec:indepatch}

\begin{SCfigure}
    \centering
    \includegraphics[width=0.2\linewidth]{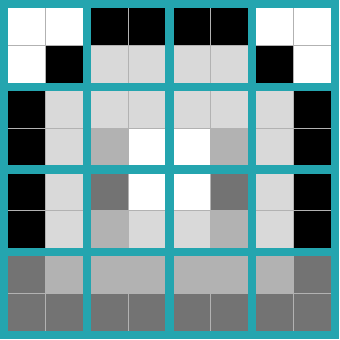}
    \hspace{15pt}
    \includegraphics[width=0.4\linewidth]{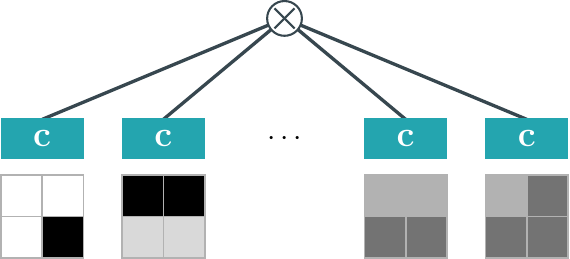}
    \hspace{10pt}
    \caption{\textbf{\indepatch PC} $c^{\prime}$ built on  $2\times2$ independent patches ($L\!=\!16$), where the same circuit $c$ (e.g., like in \cref{fig:rg-to-pc}) is shared across all patches. The final likelihood is obtained by multiplying all values together (\cref{eq:indepatch}).}
    \label{fig:ipa}
\end{SCfigure}

Training a PC on high dimensional patch-structured modalities, such as image data, can be challenging due to the high number of parameters, and a full model might not fit into memory. 
To alleviate this problem, several works \citep{anji2022scaling-pcs-lvd, liu2023understanding,liu2024scaling,zhang_scaling_2025,garg2025bitblasting} split images into $L$ non overlapping patches of size $D\times D\times C$ and trained a single PC $c$ on these patches.
To retrieve a probability distribution defined over all pixels (and channels) $\X$, they make the \textit{independent patch assumption} (\indepatch), i.e., compute: 
\begin{equation}\label{eq:indepatch}
\prob_{\indepatch}(\X) = \indepatch_{D}(c, \X) = \prod\nolimits_{i=1}^{L(D)}c(\X_{i}; \boldsymbol{\Omega})
\end{equation}
where  $\X=\X_1 \sqcup\ldots\sqcup\X_L$ denotes the partitioning of pixels  into disjoint patches.
This setting can be seen as a form of patch-level parameter sharing \citep{liu2023understanding}, as it implicitly represents a circuit $c^{\prime}$ defined over the whole scope $\X$ and formed by a decomposable product unit over $L$ \textit{shallow copies} of $c$, each taking input over a different patch $\X_i$ (\cref{fig:ipa}).
\footnote{Note that in the way \indepatch has been introduced in the literature, patches are both independent \textit{and} modelled by a single circuit. We investigate in \cref{app:ipa_noshare} what happens when the independence holds but not patch-sharing. We find that the conundrum still holds.}
Note that these copies do not need to be kept in memory together: the same base circuit $c$ with fixed parameters $\boldsymbol{\Omega}$ is ``swept'' across the whole image, while its input units are \textit{rescoped}, i.e., their distributions are assumed to operate on different pixel scopes for every patch.
We refer to the implicit circuit $c^{\prime}$ as an \indepatch, 
and by contrast, we call \fullpc a classic PC defined and instantiated on the whole image.

With \indepatch, one can effectively train smaller circuits
that are more computationally efficient, and at the same time retrieve a joint distribution for the whole image.
For example, while a \fullpc using QT RGs with $K=512$ units trained on ImageNet32 (having $32\times 32\times 3$ pixels) requires 670M parameters and $16.6$ GB of memory in the cirkit library \citep{The_APRIL_Lab_cirkit_2024},
an \indepatch PC trained over  $8\times 8\times 3$ patches requires only 41M parameters and $1.5$ GB of memory 
to process the same batch of 32 images. 

At the same time, \cref{eq:indepatch} imposes a strong assumption that is very likely violated on real-world images and has clear limitations:
sampling an image from an \indepatch PC would yield fragmented samples with no global coherence (see \cref{fig:all-samples-qt}) as each instance of the base PC is sampled independently.
As such, one could safely presume that the expressiveness of a PC not making the \indepatch or using larger patch sizes can only improve. 
However, this is not always the case in practice, as we discuss next.

\begin{figure*}[!t]
    \centering
    \hfill\includegraphics[width=0.31\linewidth]{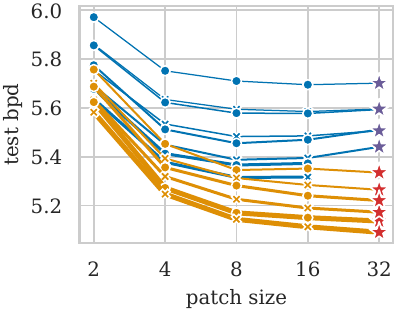}\hfill
    \includegraphics[width=0.31\linewidth]{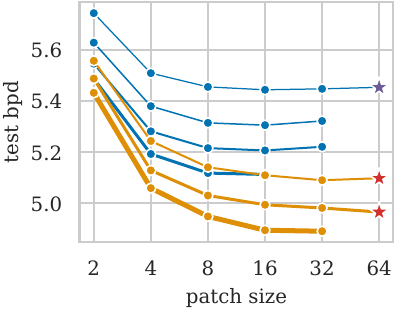}\hfill
    \includegraphics[width=0.31\linewidth]{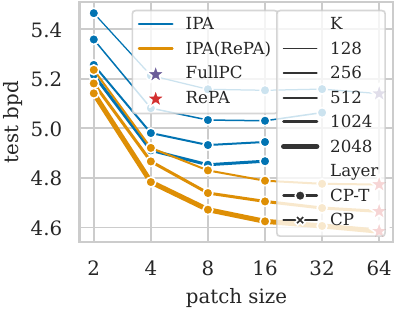}\hfill
    \caption{\textbf{\repa PCs improve over \indepatch and \fullpc} while greatly alleviating the \indepparadox  as  measured in bits-per-dimension (bpd), for increasing  patch sizes and QT RGs on ImageNet32 (left), ImageNet64 (center), and CelebA (right).
    More results in \cref{fig:bpd_vs_patch}.
    }
    \label{fig:bpd_vs_patch2}
\end{figure*}

\paragraph{The \indepparadox.} 
Contrary to our intuition, a systematic evaluation of how \indepatch circuits scale w.r.t. patch size ($D$)
reveals an opposite trend: performance significantly degrades or stagnates when $D$ increases beyond 8.
\cref{fig:bpd_vs_patch2} shows this in terms of bits-per-dimension (bpd)---pixel-normalized negative log-likelihoods---for several PC architectures learned on ImageNet32, CelebA and ImageNet64.
This puzzling behavior implies that there is an expressiveness-learnability trade-off in PCs: \textit{in high-dimensional spaces, models that are more expressive in theory can be harder to learn in practice}.
Similar conclusions have been advanced when talking about increasing $K$ for overparameterized circuits \citep{anji2022scaling-pcs-lvd}.
\cref{fig:bpd_vs_patch2} reveals that this consistently happens for every $K$, hinting at a different bottleneck in scalability that pertains to the circuit input dimensionality more than scaling the number of parameters through $K$.
All in all, we are adding another ``piece'' to the puzzling evidence that training large PCs is hard.

We note that this trend consistently appears for QT and QG RGs and all layers, and only the performance of CLT RGs (cf. \cref{fig:treerg2pc}) seems to degrade slightly more gracefully (see \cref{fig:bpd_vs_patch}).
At the same time, we confirmed the presence of this phenomenon also for more advanced PC variants, i.e., probabilistic integral circuits \citep{gala2024pic,gala2024scaling}
and sum of squared circuits \citep{loconte2024subtractive,loconte_sum_2025,loconte2026square}, as we discuss in \cref{app:pics-sos}.
We now introduce our way to introduce dependencies back in \indepatch PCs while retaining the efficiency gains that parameter sharing offers.

\section{Recursive patch-aware (\repa) parameter sharing}
\label{sec:repa}

\paragraph{How to reintroduce dependencies?}
A natural approach to consider is to replace the single product unit of \cref{eq:indepatch} with a deeper circuit  $\toppatch$ parameterized by $\boldsymbol{\Omega}_{\mathsf{top}}$.
Then, the sum units in $\toppatch$ could be used to recover the missing correlations across patches.
To do so, one can collect all the output activations of the  base circuit $c_{\mathsf{base}}$ operating on patches $\X_i \in [0,255]^{D\times D\times C}$, and feed them as inputs to $\toppatch$, i.e.,
\begin{align}
\label{eq:top-patch}
   \prob_{\mathsf{top}}(\X)= \toppatch( \Out^{(1)}; \boldsymbol{\Omega}_{\mathsf{top}}) & \;\;
   \text{where}\ \Out^{(1)}= \Out^{(1)}_1,\ldots,\Out^{(1)}_L \;\;
   \text{and}\ \Out^{(1)}_i = c_{\mathsf{base}}(\X_{i}; \boldsymbol{\Omega}_{\mathsf{base}})
\end{align}
where $\Out^{(1)}_i\in [0,1]^{K}$ for $i=1,\ldots,L$ is the output of the base circuit $c_{\mathsf{base}}$ on the $i$-th patch.\footnote{We note that $\toppatch$, as well as all higher-level PCs in \repa, contain only sum and product layers.}
This idea has been explored by \citet{liu2023understanding} in combination with their latent variable distillation (LVD) pipeline. LVD learns the parameters $\boldsymbol{\Omega}_{\mathsf{base}}$ of a PC with CLT RG by mimicking the latent variables of a deep generative model \citep{anji2022scaling-pcs-lvd}. 
They employed a two-stage approach: first the base circuit $c_{\mathsf{base}}$ is learned by maximum likelihood estimation (MLE) over image patches while conditioned on the latent variables of a deep generative model. 
Second, the outputs $\Out^{(1)}$ are collected, and the Chow-Liu algorithm (see \cref{app:rg}) is used to heuristically  learn the RG for $\toppatch$.
Its parameters $\boldsymbol{\Omega}_{\mathsf{top}}$ are subsequently fine-tuned by MLE.

\looseness=-1
However, this two-stage approach is unnecessary for patch-aligned RGs such as QTs and QGs (\cref{sec:background-pc}).
Because they already recursively split the data into patches, they naturally enable end-to-end training, as we discuss next.
Furthermore, although a monolithic $\toppatch$ circuit can recover patch dependencies, it can still hinder scalability, especially when $c_{\mathsf{base}}$ is much smaller.
For now, we focus on a systematic parameter sharing scheme that improves upon the $\toppatch$ idea and enables end-to-end training.

\begin{SCfigure}
    \centering
    \includegraphics[width=0.2\linewidth]{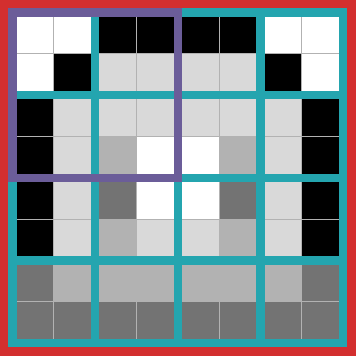}
    \hspace{15pt}
    \includegraphics[width=0.4\linewidth]{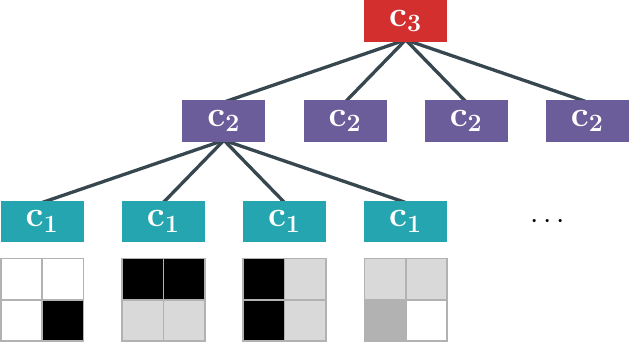}
    \hspace{10pt}
    \caption{\textbf{\repa recursively splits the image into a hierarchy of patches.} 
    The patches at each layer $i$ of the hierarchy (color-coded) share the same circuit $c_i$, which is rescoped according to each patch it is being evaluated on.  
    }
    \label{fig:repa}
\end{SCfigure}

\paragraph{A hierarchy of patches.}
The main idea behind \repa is to apply parameter sharing at every level of a hierarchy in which coarser patches are then split into smaller ones in a regular way.
For example, consider the $8\times8\times1$ image shown in \cref{fig:repa}.
At the lowest level, we define a fine grid (in blue) comprising $L^{(1)}=16$ patches of size $2\times 2\times 1$, similarly to \indepatch PCs (\cref{fig:ipa}).
Unlike \indepatch, and unlike using a single circuit $\toppatch$ as above, we recursively merge patches, creating circuit layers that are shared at higher levels of the hierarchy.
For example, by grouping $2\times 2$ contiguous patches, we first form $L^{(2)}=4$ purple patches at level $2$, each covering $4 \times 4\times 1$ pixels, and finally model a single patch at the third level (in red), comprising the whole image.

More generally, given a patch grouping size, a \repa PC is a collection of $T$ shared circuits $c_{1:T}$, each parameterized by $\boldsymbol{\Omega}_{t}$, where $t=1,\ldots,T$ and $T$ is the height of the hierarchy induced by the grouping.
For a $32\times32\times3$ image, the same $2\times 2$ patch grouping used in the above example yields a hierarchy of $T=5$ levels, comprising $L^{(1)}=256,L^{(2)}=64,L^{(3)}=16,L^{(4)}=4,L^{(5)}=1$ patches each.
For a patch hierarchy of $T$ levels, and a base patch grouping $D\times D\times C$, a \repa PC defines a distribution $\prob_{\repa}(\X)$ over the whole image as follows:
\begin{align}  
   \repa_D(c_{1:T}, \X) = c_T(\Out^{(T-1)}; \boldsymbol{\Omega}_{T}) \label{eq:repa} \;\; &\text{where}\ \Out^{(t)}= \Out^{(t)}_1,\ldots,\Out^{(t)}_{L^{(t)}}\\
    \notag\text{and} \;\;\Out^{(t)}_i = c_t(\textsc{getPatches}_t(\Out^{(t-1)}, i); \boldsymbol{\Omega}_{t}) \;\; &\text{and} \;\;\Out_i^{(1)} = c_1(\X_i; \boldsymbol{\Omega}_{1}).
\end{align}
In words, the first level of the hierarchy behaves as \cref{eq:top-patch}: the base circuit $c_1$ with $K^{(1)}$ units outputs a tensor $\Out_i^{(1)}\in[0,1]^{D^{(1)}\times D^{(1)}\times K^{(1)}}$ for each input patch $\X_i$, where $i=1,\ldots,L^{(1)}$.
Recursively, each level-$t$ circuit $c_t$ takes as input the $D^{(t)}\times D^{(t)}\times K^{(t-1)}$ tensor aggregating the outputs of the circuits at level $t-1$, where $K^{(t-1)}$ denotes the number of units per layer in $c_{t-1}$. To aggregate the outputs, we define the function $\textsc{getPatches}_t(\Out^{(t-1)}, i)$ which, for a level $t$, collects the outputs of the previous level $t-1$ to build the $i$-th patch of the current level.
This process continues until the last level $T$, which consists of a circuit with $K^{(T)}$ units except for the last layer which has a single output unit (i.e., $K=1$). 
\cref{app:alg:inference_repa} details this procedure.
The resulting construction is visualized in \cref{fig:repa}.
For simplicity, in all our experiments, we keep the patch grouping size $D\times D$ and overparameterization level $K$ constant across levels of the hierarchy.
Note that, under this assumption, all higher-level circuits $c_{2:T}$ have the same number of parameters and size.

\looseness=-1
\paragraph{Memory and time gains.} Similarly to \indepatch PCs, \repa does not require explicitly materializing the circuit created by the recursive procedure and storing it in memory; instead, it is enough to simply store the sequence $c_{1:T}$.
Therefore, to evaluate \cref{eq:repa} and implement the feedforward computation of probabilities in the PC, we can simply evaluate each circuit in the hierarchy for each intermediate patch.
As a result, the overall number of FLOPs is on par with a \fullpc.
Nevertheless, parameter sharing in \repa allows us to greatly scale to large images and numbers of units. 
As we show in \cref{fig:fig-1,sec:experiments}, this yields circuits that are not only more efficient than a \fullpc, but also than \indepatch PCs.
Recall from \cref{sec:indepatch} that a \fullpc using QT RGs with $K=512$ units on $32\times 32\times 3$ images requires 670M parameters and $16.6$ GB of memory, while an \indepatch PC trained over  $8\times 8\times 3$ patches reduces this to 41M parameters and $1.5$ GB of memory.
A \repa with $2\times 2$ patch grouping, the same $K=512$ units and QT RG, amounts to only $5$M parameters and $0.73$ GB.
Furthermore, \repa is orthogonal to other scaling techniques like Monarch layers \citep{zhang_scaling_2025} allowing us to improve efficiency further. \cref{app:repa-monarch} demonstrates the positive interaction between both techniques.

While retaining and even improving upon the efficiency gains of \indepatch PCs, \repa circuits crucially break the independence assumption and model the dependencies across pixels appearing in the disjoint patches.
It achieves this by introducing $K$ sum units per layer in the circuit hierarchy above the first level.
These additional sum units act as latent variables that increase the model's expressiveness \citep{peharz2015theoretical,peharz_latent_2017, gala2024pic} effectively counterbalancing the capacity restrictions introduced by parameter sharing.
Up to this point we have assumed that the hierarchical patch partitioning of \repa is compatible with the circuit's RG.
While this holds naturally for QT and QG RGs \citep{loconte2025what}, it is not the case for CLTs. 
To this end, \cref{app:repa-clt} demonstrates how \repa can be adapted for non-patch-aligned RGs with similar benefits. However, we chose to focus on end-to-end compatible parameter sharing. As such, we are interested in extending parameter sharing beyond patches.

\section{Can we exploit another source of structure than patches? (\nopa)}
\label{sec:tabsha}

Image data is a practical candidate for parameter sharing as it displays an easily exploitable structure; however, PCs are not yet competitive on images. On the other hand, PCs have been shown to be very competitive for tabular data generation \citep{scassola_sobering_2026}, a modality that does not display any obvious structure. Indeed, the organization of the columns is completely arbitrary and has no exploitable meaning.

Abstracting from the data, we notice that the PC itself is already a source of structure. Using the hierarchy described by the circuit, we can create a simple heuristic that applies parameter sharing for any modality. 
We use a simple recipe, starting from an arbitrary PC $c$, (1) mark the depth of each layer as the largest number of layers between itself and an input layer, (2) group every sum layer with the same depth, and (3) share a single weight tensor across the group of layers. We call this \textit{no patch sharing} (\nopa). 

We note here that we do not share any input parameters and that the number of parameters that are shared depends on the region graph.
Intuitively, we rely on the idea that, even on unstructured data, the mixtures defined by the sum layer at the same depth will have a similar effect on their input probability.
A similar idea has been explored previously while factorizing CP layers at the same topological depth \cite{loconte2025what} 
(see \cref{app:factorized_cp} for an in-depth comparison). 
While \nopa poses indeed a stronger expressiveness bottleneck than \repa, we will show that it will still be effective on low-dimensional tabular data (\cref{sec:experiments}).

\begin{figure}[t]
    \centering
    \hfill\includegraphics[height=3.25cm]{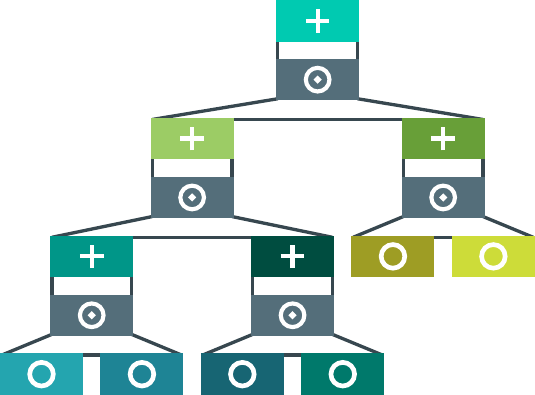}\hfill
    \includegraphics[height=3.25cm]{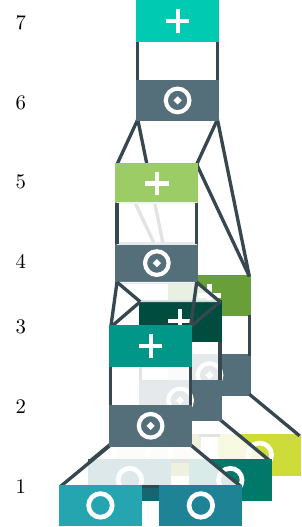}\hfill
    \includegraphics[height=3.25cm]{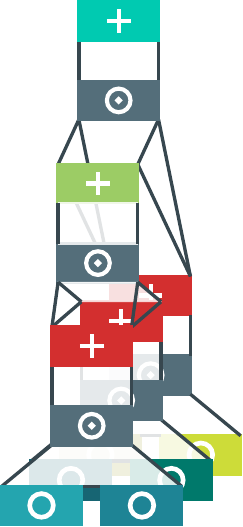}\hfill
    \caption{\textbf{\nopa shares parameters across the topological depth}. The layers of an unfolded circuit (left) can be grouped by their topological ordering (center) to \textit{fold} the circuit, making it more amenable to GPU accelerators. \nopa exploits this grouping by sharing parameters across sum layers at the same topological depth (right). Each color represents a different set of parameters, we highlight in red the three sum layers that share the same weight tensor, we use grey for layers without parameters. 
    }
    \label{fig:nopa}
\end{figure}

\section{Experiments}
\label{sec:experiments}
We are interested in answering the following research questions.
(\textbf{RQ1}) What is the impact of the independence assumption (IPA) on distribution estimation quality? 
(\textbf{RQ2}) How does \repa improve on \indepatch  regarding quality (bpd) versus efficiency (time, memory, FLOPs)?
(\textbf{RQ3}) Is the impact of parameter sharing similar on tabular data (\nopa)?
And finally, (\textbf{RQ4}) how does parameter sharing impacys PCs as generative models?
We remark that our aim is \textit{not} to set the new SotA bpds for image or tabular datasets, but to \textit{understand} when and how parameter sharing can make PCs effective and efficient distribution estimators and generative models.

\paragraph{Datasets.} Following prior work evaluating PCs on image data \citep{gala2024scaling,loconte2025what,mari2023unifying,zhang_scaling_2025}, we train and evaluate on two resolutions of ImageNet \cite{jia2009imagenet}: $32\times32\times3$ and $64\times64\times3$, and on CelebA \cite{liu2015deep}, which has been  cropped and resized to $64\times64\times3$ (see \cref{app:dataset}).  
We employ the YCoCg-R color transform as defined in \citet{malvar_color_2003} as it was introduced in \citet{gala2024scaling} for PC learning.\footnote{We note that the transform of the same name implemented in \citet{zhang_scaling_2025,garg2025bitblasting} is different from the original specification \citep{malvar_color_2003} and the one used in \citet{gala2024scaling}.} For tabular data, we used the same seven UCI datasets as TabPC: \textit{Adult} \cite{adult_2}, \textit{Beijing} \cite{beijing_pm2.5_381}, \textit{Default} \cite{default_of_credit_card_clients_350}, \textit{Diabetes} \cite{diabetes_130-us_hospitals_for_years_1999-2008_296}, \textit{Magic} \cite{magic_gamma_telescope_159}, \textit{News} \cite{online_news_popularity_332}, and \textit{Shoppers} \cite{online_shoppers_purchasing_intention_dataset_468}. 
Each dataset is split into train, validation, and test sets (see \cref{app:tab:split_per_dataset}).

\paragraph{PC architectures.} For our experiments on image data, we conduct an extensive hyperparameter search across different layers (CP, CP-T), RGs (QT, QG and CLT, see \cref{sec:background-pc}) and a number of units $K$ per layer that goes from 128 up to 4096.
We first evaluate a large grid of configurations on ImageNet32 to identify the best configurations to run on ImageNet64 and CelebA, which are larger and more computationally expensive.
For example, we quickly observed that CP and CP-T layers yield similar dynamics; therefore, we primarily use CP-T in our experiments unless otherwise specified. 
For tabular data, we use the same setup as \textsc{TabPC} \citep{scassola_sobering_2026}, which applies PCs to tabular data. Because these models use CLTs, the exact number of layers that benefit from parameter sharing can vary across datasets and random seeds. We detail all the configurations we explored in \cref{app:hyperparams}.

\paragraph{Training.} Models for image data are optimized with Adam \citep{kingma2014adam}. We briefly experimented with EM, but found that i) it is still affected by the \indepparadox and ii) it does not provide substantially better bpds than Adam when we use the same large batch size by leveraging gradient accumulation; see \cref{app:fig:em_vs_adam} and \cref{app:tab:batch_size}.
We use a fixed learning rate of $0.05$, use 300 max epochs and do early stopping when the validation loss does not improve for 3 epochs. \cref{app:tab:lr} shows that this fixed learning rate performed best for all tested architectures. 
The minimum delta for the early stopping is set per dataset (see \cref{app:tab:hyperparams_per_dataset}).
We use a single patch per image during training. 

For tabular data, we followed the configuration of TabPC by using RAdam \citep{liu_variance_2021} and a learning rate scheduler that reduces the learning rate when the training does not progress. We report all the configurations in \cref{app:tab:hyper_grid_tabpc}. 

\paragraph{Hardware and time limit.} We use one GPU with 40 GB of memory, and we reserve 128 GB of RAM and 16 CPU cores for each experiment. No experiments were allowed to run for more than 2 days. Any experiment lasting more than 2 days was killed and not included in our analysis. More details can be found in \cref{app:hw}.

\paragraph{RQ1: The effect of the \indepparadox on distribution estimation.}

We train \indepatch PCs for several patch sizes $D\times D\times C$, with $D$ starting from 2 and reaching the full image size (either 32 or 64, depending on the dataset), at which point we obtain a \fullpc. 
We compare against a \repa PC defined over the full image using a $2\times 2$ patch grouping (see \cref{fig:bpd_vs_patch2}).
\cref{fig:bpd_vs_patch} shows the test bpd obtained by all configurations of \indepatch, \repa and \fullpc on QT and QG RGs.
The \indepparadox is clearly visible: \fullpc obtains worse or marginally better bpd than \indepatch models that assume independence.
We see a similar trend across different optimizers (\cref{app:fig:em_vs_adam}), with only slight improvements when using more sophisticated PC variants (\cref{app:fig:sos_vs_pics}) or CLT RGs (\cref{fig:bpd_vs_patch}).
In contrast, \repa PCs consistently outperform both \fullpc and \indepatch across architectures and variants, illustrating that recursive parameter sharing does improve learnability.
For completeness, we report the bpd of \repa architectures defined on intermediate patches rather than the whole image.
In this case, we recover the joint distribution by assuming patch independence, denoted as \textsc{IPA}(\repa) as in \cref{eq:ipa-repa}.
Although \textsc{IPA}(\repa) PCs outperform \indepatch PCs, assuming independence at the last layers still leads to saturating bpd in some configurations.
Therefore, to RQ1, we conclude that parameter sharing does significantly help.
However, it does not entirely resolve the learnability bottleneck in PCs. \cref{app:tab:full_bpd_res} provides the complete list of results for \indepatch and \repa.

\paragraph{RQ2: Estimation quality vs efficiency in \indepatch and \repa.}
\label{subsec:efficiency_repa}

Following \citep{mari2023unifying,loconte2025what,zhang_scaling_2025}, we argue that efficiency is as crucial as distribution estimation quality, especially when we have limited computational resources and need to scale to large images. %
We use torch CUDA APIs \cite{pytorchTorchcudax2014} to compute peak GPU memory and average step time.
Additionally, we compute the FLOPs using torch's \texttt{FlopCounterMode} \cite{githubPytorchtorchutilsflop_counterpyMain}.
All efficiency metrics are computed on an NVIDIA L40 GPU for a batch of 32 images over 200 steps, where a step consists of a forward and a backward pass.
We focus on QG and QT RGs, as they allow for end-to-end learning with \repa. 
\cref{fig:fig-1} shows that using \repa systematically pushes the models toward the Pareto front between model quality and step time or GPU memory. 
Full results for all datasets are provided in \cref{app:fig:bpd_vs_efficiency}.
As expected, FLOPs remain the same, since our parameter sharing approach primarily reduces memory consumption (\cref{sec:repa}).

\paragraph{RQ3: Adapting parameter sharing for tabular data (\nopa).}

\begin{figure*}[!t]
    \centering
    \includegraphics[width=0.32\linewidth]{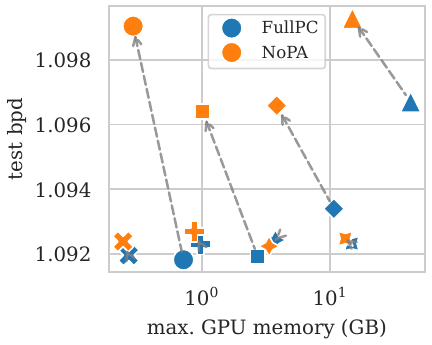}
    \includegraphics[width=0.32\linewidth]{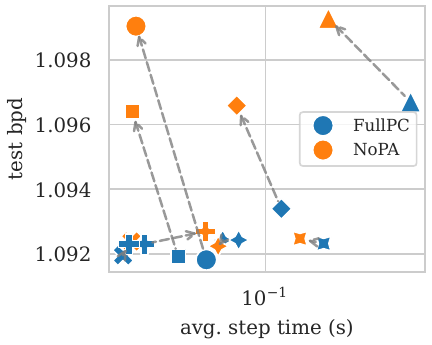}
    \includegraphics[width=0.32\linewidth]{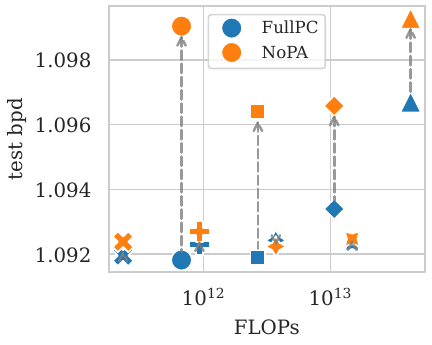}
    \caption{\textbf{Our \nopa (\cref{sec:tabsha}) setting on the \textit{Adult} dataset reduces training time and memory consumption} with minimal increase in bpd. Each shape is a pair (layer type, $K$).}
    \label{fig:tabpc_res}
\end{figure*}

\cref{fig:tabpc_res} shows that our \nopa setting provides substantial memory and time gains with minimal loss in density estimation. On Adult, the largest bpd loss between \fullpc and \nopa is $0.006$. \cref{app:fig:full_tabpc_bpd} shows the same tendency for the other datasets.
While there are no definite trends showing an improvement in bpd, the significant efficiency boost gained with this  simple approach paves the way for more complex parameter sharing on tabular data. These results are quite surprising: we get similar bpds while using a very naive heuristic, assuming that layers sharing the same topological depth have similar functions. This prompts for more extensive research into how PCs use their parameters.

\begin{figure*}[!t]
    \centering
    \includegraphics[width=0.32\linewidth]{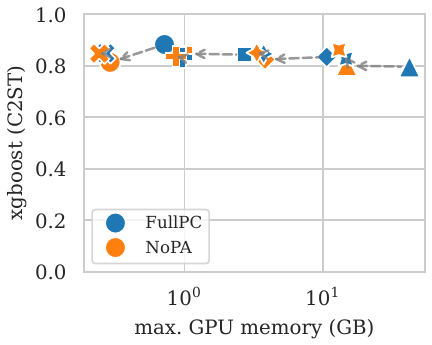}
    \includegraphics[width=0.32\linewidth]{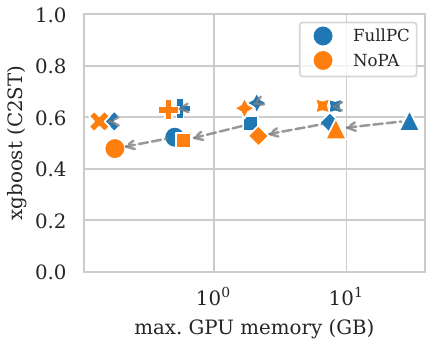}
    \includegraphics[width=0.32\linewidth]{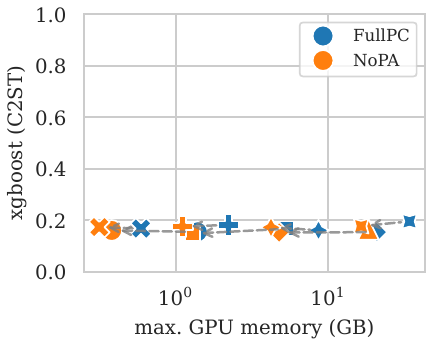}
    \caption{\textbf{Our \nopa (\cref{sec:tabsha}) setting on the \textit{Adult} (left), \textit{Beijing} (center) and \textit{Default} (right) datasets shows minimal decrease in sample quality.} Each shape is a pair (layer type, $K$).}
    \label{fig:tabpc_res_c2st}
\end{figure*}

\paragraph{RQ4: How does parameter sharing impacts PCs as generative models?}
\citet{scassola_sobering_2026} provides us with a metric to quantify sampling quality called the \textit{classifier two-sample test} (C2ST) \cite{lopez-paz2017revisiting} score, using an XGBoost \cite{chen_xgboost_2016} model.
\cref{fig:tabpc_res_c2st} shows that there is no substantial decrease in sample quality when using \nopa compared to \fullpc. \cref{app:fig:full_tabpc_c2st} shows the same results for the seven datasets and all efficiency metrics.  
This further demonstrates that parameter sharing has little negative effect on the quality of the model.

\begin{figure}[!t]
\setlength{\tabcolsep}{0.5pt}    
\centering
        \begin{tabular}{ccccc}
        \multicolumn{5}{c}{\small\indepatch (bpd = 5.00)}\\
        \includegraphics[width=0.06\linewidth]{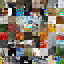} &
        \includegraphics[width=0.06\linewidth]{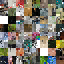} &
        \includegraphics[width=0.06\linewidth]{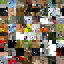} &
        \includegraphics[width=0.06\linewidth]{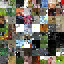} &
        \includegraphics[width=0.06\linewidth]{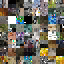} \\[-2pt]
        \includegraphics[width=0.06\linewidth]{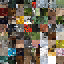} &
        \includegraphics[width=0.06\linewidth]{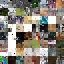} &
        \includegraphics[width=0.06\linewidth]{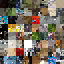} &
        \includegraphics[width=0.06\linewidth]{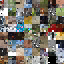} &
        \includegraphics[width=0.06\linewidth]{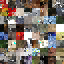} \\[-2pt]
        \includegraphics[width=0.06\linewidth]{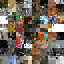} &
        \includegraphics[width=0.06\linewidth]{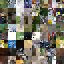} &
        \includegraphics[width=0.06\linewidth]{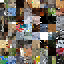} &
        \includegraphics[width=0.06\linewidth]{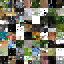} &
        \includegraphics[width=0.06\linewidth]{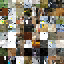} \\[-2pt]
        \includegraphics[width=0.06\linewidth]{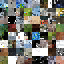} &
        \includegraphics[width=0.06\linewidth]{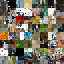} &
        \includegraphics[width=0.06\linewidth]{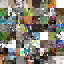} &
        \includegraphics[width=0.06\linewidth]{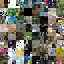} &
        \includegraphics[width=0.06\linewidth]{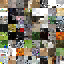} \\[-2pt]
        \includegraphics[width=0.06\linewidth]{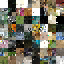} &
        \includegraphics[width=0.06\linewidth]{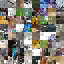} &
        \includegraphics[width=0.06\linewidth]{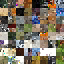} &
        \includegraphics[width=0.06\linewidth]{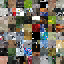} &
        \includegraphics[width=0.06\linewidth]{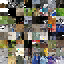} \\[-2pt]
    \end{tabular}
    \begin{tabular}{ccccc}
        \multicolumn{5}{c}{\small\repa (bpd = 4.99)}\\
        \includegraphics[width=0.06\linewidth]{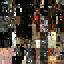} &
        \includegraphics[width=0.06\linewidth]{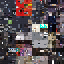} &
        \includegraphics[width=0.06\linewidth]{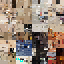} &
        \includegraphics[width=0.06\linewidth]{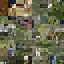} &
        \includegraphics[width=0.06\linewidth]{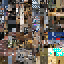} \\[-2pt]
        \includegraphics[width=0.06\linewidth]{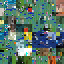} &
        \includegraphics[width=0.06\linewidth]{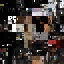} &
        \includegraphics[width=0.06\linewidth]{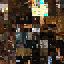} &
        \includegraphics[width=0.06\linewidth]{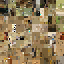} &
        \includegraphics[width=0.06\linewidth]{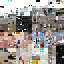} \\[-2pt]
        \includegraphics[width=0.06\linewidth]{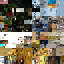} &
        \includegraphics[width=0.06\linewidth]{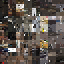} &
        \includegraphics[width=0.06\linewidth]{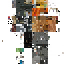} &
        \includegraphics[width=0.06\linewidth]{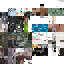} &
        \includegraphics[width=0.06\linewidth]{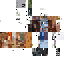} \\[-2pt]
        \includegraphics[width=0.06\linewidth]{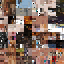} &
        \includegraphics[width=0.06\linewidth]{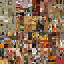} &
        \includegraphics[width=0.06\linewidth]{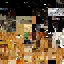} &
        \includegraphics[width=0.06\linewidth]{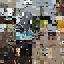} &
        \includegraphics[width=0.06\linewidth]{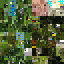} \\[-2pt]
        \includegraphics[width=0.06\linewidth]{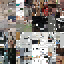} &
        \includegraphics[width=0.06\linewidth]{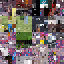} &
        \includegraphics[width=0.06\linewidth]{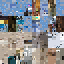} &
        \includegraphics[width=0.06\linewidth]{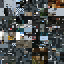} &
        \includegraphics[width=0.06\linewidth]{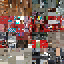} \\[-2pt]
    \end{tabular}
        \begin{tabular}{ccccc}
        \multicolumn{5}{c}{\small\fullpc (bpd = 5.45)}\\
        \includegraphics[width=0.06\linewidth]{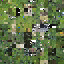} &
        \includegraphics[width=0.06\linewidth]{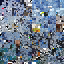} &
        \includegraphics[width=0.06\linewidth]{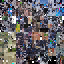} &
        \includegraphics[width=0.06\linewidth]{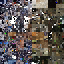} &
        \includegraphics[width=0.06\linewidth]{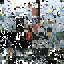} \\[-2pt]
        \includegraphics[width=0.06\linewidth]{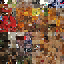} &
        \includegraphics[width=0.06\linewidth]{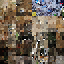} &
        \includegraphics[width=0.06\linewidth]{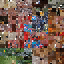} &
        \includegraphics[width=0.06\linewidth]{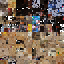} &
        \includegraphics[width=0.06\linewidth]{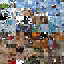} \\[-2pt]
        \includegraphics[width=0.06\linewidth]{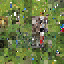} &
        \includegraphics[width=0.06\linewidth]{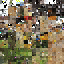} &
        \includegraphics[width=0.06\linewidth]{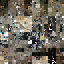} &
        \includegraphics[width=0.06\linewidth]{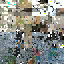} &
        \includegraphics[width=0.06\linewidth]{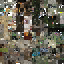} \\[-2pt]
        \includegraphics[width=0.06\linewidth]{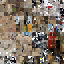} &
        \includegraphics[width=0.06\linewidth]{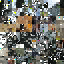} &
        \includegraphics[width=0.06\linewidth]{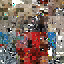} &
        \includegraphics[width=0.06\linewidth]{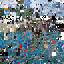} &
        \includegraphics[width=0.06\linewidth]{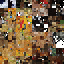} \\[-2pt]
        \includegraphics[width=0.06\linewidth]{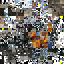} &
        \includegraphics[width=0.06\linewidth]{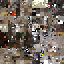} &
        \includegraphics[width=0.06\linewidth]{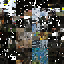} &
        \includegraphics[width=0.06\linewidth]{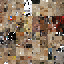} &
        \includegraphics[width=0.06\linewidth]{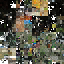} \\[-2pt]
    \end{tabular}

    \caption{
    \textbf{\repa samples retrieve global patch dependencies} and are closer to \fullpc on ImageNet64 where we show \indepatch $8\times 8\times3$ and $K=1024$ (left),  \repa (center) with $K=512$ and \fullpc with $K=128$ (right). All PCs use a QG RG.
    More samples in \cref{fig:all-samples-qg,fig:all-samples-qt}. We take modes of the input unit distributions rather than sampling from them, see details in \cref{app:sampling}.
    }
    \label{fig:some-samples}
\end{figure}

Regarding images, the quality of image samples from PCs is notoriously behind that of other deep generative models for image data \citep{lang2022elevating,braun2025tractable}.
Nevertheless, we qualitatively inspect samples generated by \indepatch, \repa and \fullpc circuits to understand if \repa can break the independence assumption bottleneck and recover some of the correlations across patches that \fullpc can capture, but \indepatch cannot by its very definition.
\cref{fig:some-samples,fig:all-samples-qg,fig:all-samples-qt} show how \repa PCs successfully recover some correlations, such as colour continuity, similar to \fullpc.
This yields much less ``patchy'' samples than those produced by \indepatch. 
We highlight that the bpd achieved by these models does not strictly correlate with sample quality \citep{theis_note_2016}. In fact, while \fullpc scores a test bpd of $5.45$, the \indepatch scores $5.00$, and the \repa PC achieves an even better $4.99$.
Coupled with the \indepparadox, this issue further highlights the need to report not only bpds for learned PCs, but also to inspect the visual quality of their samples.

\section{Conclusion and future work}

In this work, we focused on the independent patch assumption (\indepatch) when learning PCs, which allows scaling PC experiments by improving efficiency \citep{anji2022scaling-pcs-lvd,liu2023understanding,liu2024scaling,zhang_scaling_2025,garg2025bitblasting}.
Unexpectedly, \indepatch PCs yield similar or better bpds than PCs trained on whole images.
This highlights a systemic learnability problem when scaling PCs to high resolution images.
We introduced \repa, a recursive patch sharing technique that alleviates this issue, offering lower bpd, while significantly improving efficiency. 
We also devised \nopa, a technique that exploits the structure of the PC itself to enable parameter sharing on any modality. 
Surprisingly, our naive heuristic yields considerable efficiency gains at the cost of a marginal decrease in density estimation and sampling quality.
While PCs are still far behind models like flows and diffusion on images, this is an important stepping stone towards scaling PCs as efficient distribution estimators and accurate generative models.
Furthermore, on tabular data generation, a task where PCs are known to be competitive, we opened the way for significant efficiency improvements using more complex parameter sharing techniques.
Future work should focus on understanding the learnability bottleneck and on developing more advanced parameter sharing techniques, while extending those to other modalities such as text, and time-series data.

\section*{Author Contributions}

CB and AV conceived the original idea. They later discussed it with VB, who joined them in developing the experimental design. CB implemented the training and validation code, as well as \repa, \indepatch, and \nopa and is responsible for all plots, algorithms and experiments in the paper, with the following exception. GG implemented the efficient backward-sampling algorithm and ran the image-sampling experiments. CB led the writing of the paper, with contributions from GG, VB, and AV. AV and VB supervised all phases of the project and provided feedback throughout.

\section*{Acknowledgements}

CB and AV are supported by the "UNREAL: Unified Reasoning Layer for Trustworthy ML" project (EP/Y023838/1) selected by the ERC and funded by UKRI EPSRC.

\bibliographystyle{plainnat}
\bibliography{referomnia,clean_charles_ref}

\newpage
\appendix
\thispagestyle{empty}

\crefalias{section}{appendix}
\crefalias{subsection}{appendix}

\section{Building tensorized PCs from CLT region graphs}
\label{app:arch}

\label{app:rg}
\begin{figure}[h]
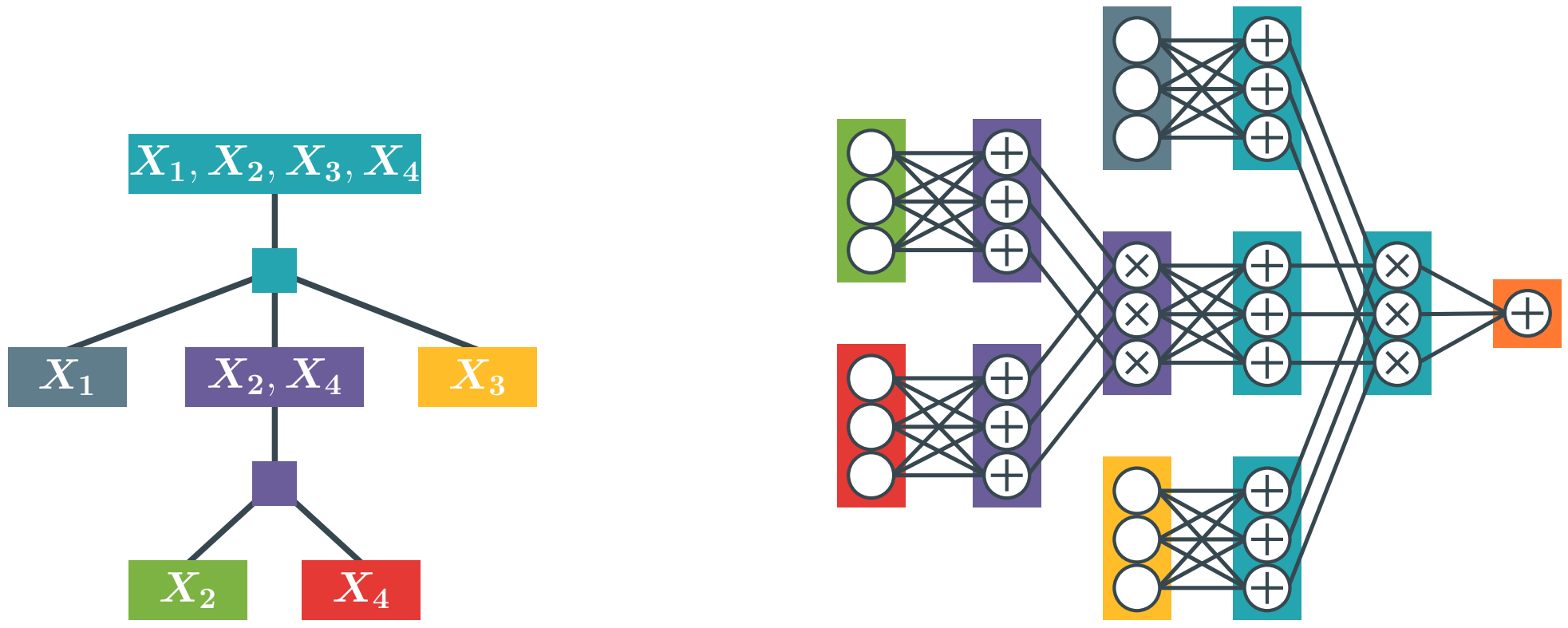

    \begin{minipage}[b]{0.4\textwidth}
        \centering
        \includegraphics[page=3,scale=0.70]{figures/clt.pdf}
        \hspace{5em}
        \subcaption{tree RG}
    \end{minipage}
    \begin{minipage}[b]{0.5\textwidth}
        \centering
        \includegraphics[page=4,scale=0.8]{figures/clt.pdf}
        \hspace{5em}
        \subcaption{Tensorized PC}
    \end{minipage}
    \hspace{10em}
    \caption{\textbf{Tree RG $\rightarrow$ Tensorized PC.}
    The Chow-Liu tree algorithm \citep{chow1968clts} can be used to build a tree region graph \citep{dang2021strudel} (a) which can then be used to build a tensorized PC parameterized with CP layers (b) or with alternative CP-T \citep{liu2021regularization,loconte2025what} or Monarch layers \citep{zhang_scaling_2025}.}
    \label{fig:treerg2pc}
\end{figure}

We implement CLT PCs by building a RG out of the tree structure learned by the Chow-Liu algorithm \citep{chow1968clts}.
Specifically, to convert such a tree  into a RG, one traverses it bottom-up and systematically alternates between region and partition nodes. Each node in the tree induces a region whose scope is the union of variables in its subtree, while leaves become singleton regions.
For every internal node, a partition node is introduced that connects the regions of its children as inputs to the parent region as output, encoding a decomposition of the parent scope into disjoint child scopes. The resulting region graph alternates regions and partitions and preserves the hierarchical structure of the original tree.
In \cref{fig:treerg2pc} we show a tree region graph and a tensorized circuit built based on it.

\section{Algorithms}
\label{app:algos}

\subsection{Algorithms for patch sharing}
\label{app:patch_algos}

\cref{app:alg:inference_ipa} provides a simple algorithm to compute the output of \indepatch. We simply multiply the outputs of each individual patch $\X_i$.

\begin{algorithm}
\SetKwComment{Comment}{/* }{ */}
\caption{ \texttt{InferenceIPA}($c$, $\X$)}\label{app:alg:inference_ipa}
\textbf{Input:} Tensorized PC $c$ over $\X = \{X_i\}_{i=1}^{L(D)}$\;
\textbf{Output:} Likelihood of $\X$. \\

\Begin{
        $\textsf{out} = 1$\;
        \For{$i \in L$}{
           $\textsf{out}= \textsf{out} \times c(\X_i)$\;
       }
        \Return{ $\textsf{out}$}
}
\end{algorithm}

\cref{app:alg:inference_repa} explains how to compute the likelihood of a sample using \repa. For each level $t\in[1,T]$, we compute the output of the $t$-th circuit for each of the $L^{(t)}$ patches, using the previous outputs as the new input and using $\textsc{GetPatches}_t$ to collect the relevant elements.
\begin{algorithm}
\SetKwComment{Comment}{/* }{ */}
\caption{ \texttt{InferenceRePA}($c_1$,\dots, $c_T$, $\X$)}\label{app:alg:inference_repa}
\textbf{Input:} $T$ tensorized PCs $c_{1:T}$ over $\X = \{X_i\}_{i=1}^{L(D^{(1)})}$\;
\textbf{Output:} Likelihood of $\X$. \\

\Begin{
        $\Out^{(0)}=\X$\;
        \For{$t \in [1,T]$}{ 
            \For{$i \in L^{(t)}$}{
           $\Out^{(t)}_i=c_t(\textsc{GetPatches}_t(\Out^{(t-1)},i))$\;
           }
        }
        \Return{ $\Out^{(T)}$}
}
\end{algorithm}

\subsection{Multi stage training for \repa}
\label{app:subsec:multi-stage-train}
As explained in \cref{app:repa-clt}, \cref{app:alg:trainrepaclt} recursively builds the $T$ circuit by (i) learning a CLT on previous level's outputs $\Out^{(t-1)}$, (ii) training all $c_{1:t}$ circuits on the train data $\X$, (iii) retrieving the output for each patch to build $\Out^{(t)}$, and finally (iv) we introduce $K-1$ new units to the last layers of $c_t$ by simply copying the same unit, with the exact same weight. This last step is important to ensure that the next level receives inputs of shape $D\times D\times K$.

By copying the same weight for every output unit, we guarantee that, at initialization, the circuit of level $t+1$, $\indepatch_{D^{T-(t+1)}}(\repa(c_{1:t+1}, \X))$, will have the exact same likelihood as $\indepatch_{D^{T-t}}(\repa(c_{1:t}, \X))$, the circuit of level $t$. Indeed, the convex sum of a constant vector returns the very same constant value. Hence, at initialization, $c_{t+1}$ essentially computes the product of its input. We note that this should be taken into account if one wants to freeze the previous circuit while training  $c_{t+1}$.

\begin{algorithm}
\SetKwComment{Comment}{/* }{ */}
\caption{ \texttt{TrainRePACLT}($T$, $D$, $\X$)}\label{app:alg:trainrepaclt}
\textbf{Input:} The number $T$ of PCs in the \repa, the size of patches $D$ and the data $\X$\;
\textbf{Output:} list of $T$ PCs that make the \repa PC. \\
\Begin{
        $\mathsf{c\_list}=[]$\;
        $\Out^{(0)}=\X$\;
        \For{$t \in [1,T]$}{ 
           $\mathsf{c\_list}.\mathsf{append}(\textsc{BuildCLTPC}(\Out^{(t-1)}, D))$\;
            $\textsc{TrainPC}(\mathsf{c\_list},\X)$\;
            $\Out^{(t)}=\textsc{InferenceRePA}(\mathsf{c\_list}, \Out^{(t-1)})$\;
            $\textsc{AugmentOutputs}(\mathsf{c\_list}[t])$\;
        }
        \Return{ $\mathsf{c\_list}$}
}
\end{algorithm}

\subsection{Sampling}
\label{app:sampling}

To introduce sampling in tensorized PCs, we view their layers at a finer granularity, namely as a collection of sum layers (matrix multiplications) and Hadamard layers (element-wise products) instead of sum-product layers.
\cref{fig:treerg2pc}(b) shows an example of tensorized PC with sum and Hadamard layers.
In \cref{alg:sampling} we detail how to sample from such structures.
We interpret the entries of each non-negative matrix $\W^{S\times K}$ of the sum layers as the parameters of categorical distributions associated to $S$ latent variables, each taking values in $\{1,\ldots,K\}$.
Note that we can always normalize a PC s.t.\ its normalization constant is equal to 1 thus yielding parameter matrices that sum up to 1 along every row, as detailed in \citep{peharz2015theoretical}.

Then, sampling a data point $\x$ translates to iteratively sampling from such latent variables according to the hierarchical structure of the circuit, i.e.\ following a topological order like a breadth-first search (BFS).
Note that sampling the latent variables corresponding to a sum layer corresponds to choosing (i) a selection of the input layers on which to recursively continue sampling, and (ii) a particular computational unit within each selected layer.
The information (i) and (ii) for each layer is stored in dictionaries.

Due to decomposability, sampling from a product layer $\layer$ translates to choosing a selection of the input computational units, as they will be defined on different variables.
Unlike sum layers where we sample from Categoricals to select such units, in product layers they are unequivocally determined by which product unit of $\layer$ has been previously selected.
We sample all sum and product layers as explained below.
Finally, we sample from the input (categorical) layers and assign values to the variables the PC is defined on.
We sample from an input layer $\layer$ when at least one input unit within $\layer$ has been selected by the sampling procedure above for sum and product layers.
That is, given $X\in\X$ the variable on which $\layer$ depends and $n_k$ the $k$-th input unit to sample from, we sample an assignment to $X$ from $n_k$.

\begin{figure}[!t]
\setlength{\tabcolsep}{0.5pt}    
\centering
\begin{tabular}{ccccc}
    \multicolumn{5}{c}{\indepatch}\\
        \includegraphics[width=0.06\linewidth]{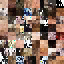} &
        \includegraphics[width=0.06\linewidth]{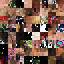} &
        \includegraphics[width=0.06\linewidth]{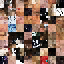} &
        \includegraphics[width=0.06\linewidth]{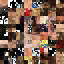} &
        \includegraphics[width=0.06\linewidth]{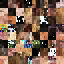} \\[-2pt]
        \includegraphics[width=0.06\linewidth]{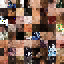} &
        \includegraphics[width=0.06\linewidth]{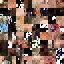} &
        \includegraphics[width=0.06\linewidth]{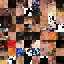} &
        \includegraphics[width=0.06\linewidth]{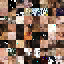} &
        \includegraphics[width=0.06\linewidth]{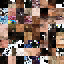} \\[-2pt]
        \includegraphics[width=0.06\linewidth]{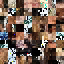} &
        \includegraphics[width=0.06\linewidth]{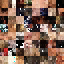} &
        \includegraphics[width=0.06\linewidth]{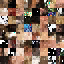} &
        \includegraphics[width=0.06\linewidth]{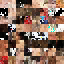} &
        \includegraphics[width=0.06\linewidth]{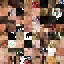} \\
\end{tabular}
    \begin{tabular}{ccccc}
    \multicolumn{5}{c}{\repa}\\
        \includegraphics[width=0.06\linewidth]{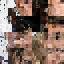} &
        \includegraphics[width=0.06\linewidth]{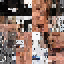} &
        \includegraphics[width=0.06\linewidth]{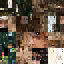} &
        \includegraphics[width=0.06\linewidth]{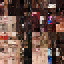} &
        \includegraphics[width=0.06\linewidth]{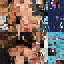} \\[-2pt]
        \includegraphics[width=0.06\linewidth]{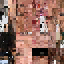} &
        \includegraphics[width=0.06\linewidth]{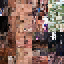} &
        \includegraphics[width=0.06\linewidth]{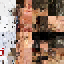} &
        \includegraphics[width=0.06\linewidth]{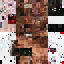} &
        \includegraphics[width=0.06\linewidth]{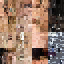} \\
\end{tabular}
        \begin{tabular}{ccccc}
        \multicolumn{5}{c}{\fullpc}\\
        \includegraphics[width=0.06\linewidth]{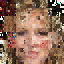} &
        \includegraphics[width=0.06\linewidth]{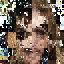} &
        \includegraphics[width=0.06\linewidth]{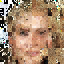} &
        \includegraphics[width=0.06\linewidth]{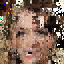} &
        \includegraphics[width=0.06\linewidth]{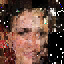} \\[-2pt]
        \includegraphics[width=0.06\linewidth]{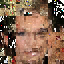} &
        \includegraphics[width=0.06\linewidth]{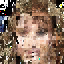} &
        \includegraphics[width=0.06\linewidth]{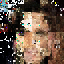} &
        \includegraphics[width=0.06\linewidth]{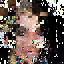} &
        \includegraphics[width=0.06\linewidth]{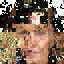} \\[-2pt]
        \includegraphics[width=0.06\linewidth]{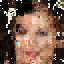} &
        \includegraphics[width=0.06\linewidth]{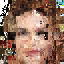} &
        \includegraphics[width=0.06\linewidth]{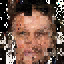} &
        \includegraphics[width=0.06\linewidth]{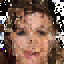} &
        \includegraphics[width=0.06\linewidth]{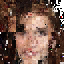} \\
\end{tabular}
        \begin{tabular}{ccccc}
        \includegraphics[width=0.06\linewidth]{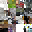} &
        \includegraphics[width=0.06\linewidth]{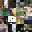} &
        \includegraphics[width=0.06\linewidth]{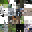} &
        \includegraphics[width=0.06\linewidth]{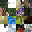} &
        \includegraphics[width=0.06\linewidth]{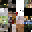} \\[-2pt]
        \includegraphics[width=0.06\linewidth]{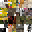} &
        \includegraphics[width=0.06\linewidth]{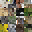} &
        \includegraphics[width=0.06\linewidth]{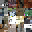} &
        \includegraphics[width=0.06\linewidth]{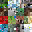} &
        \includegraphics[width=0.06\linewidth]{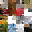} \\[-2pt]
        \includegraphics[width=0.06\linewidth]{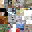} &
        \includegraphics[width=0.06\linewidth]{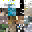} &
        \includegraphics[width=0.06\linewidth]{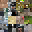} &
        \includegraphics[width=0.06\linewidth]{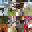} &
        \includegraphics[width=0.06\linewidth]{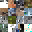} \\
\end{tabular}
    \begin{tabular}{ccccc}
        \includegraphics[width=0.06\linewidth]{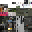} &
        \includegraphics[width=0.06\linewidth]{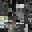} &
        \includegraphics[width=0.06\linewidth]{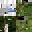} &
        \includegraphics[width=0.06\linewidth]{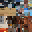} &
        \includegraphics[width=0.06\linewidth]{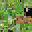} \\[-2pt]
        \includegraphics[width=0.06\linewidth]{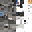} &
        \includegraphics[width=0.06\linewidth]{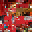} &
        \includegraphics[width=0.06\linewidth]{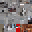} &
        \includegraphics[width=0.06\linewidth]{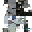} &
        \includegraphics[width=0.06\linewidth]{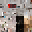} \\
\end{tabular}
        \begin{tabular}{ccccc}
        \includegraphics[width=0.06\linewidth]{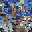} &
        \includegraphics[width=0.06\linewidth]{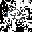} &
        \includegraphics[width=0.06\linewidth]{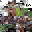} &
        \includegraphics[width=0.06\linewidth]{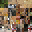} &
        \includegraphics[width=0.06\linewidth]{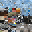} \\[-2pt]
        \includegraphics[width=0.06\linewidth]{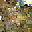} &
        \includegraphics[width=0.06\linewidth]{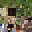} &
        \includegraphics[width=0.06\linewidth]{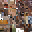} &
        \includegraphics[width=0.06\linewidth]{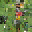} &
        \includegraphics[width=0.06\linewidth]{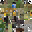} \\
\end{tabular}
        \begin{tabular}{ccccc}
        \includegraphics[width=0.06\linewidth]{media/samples_export/IPA/imagenet64/quad-graph/patch-8x8-units-1024/sample_1.png} &
        \includegraphics[width=0.06\linewidth]{media/samples_export/IPA/imagenet64/quad-graph/patch-8x8-units-1024/sample_2.png} &
        \includegraphics[width=0.06\linewidth]{media/samples_export/IPA/imagenet64/quad-graph/patch-8x8-units-1024/sample_3.png} &
        \includegraphics[width=0.06\linewidth]{media/samples_export/IPA/imagenet64/quad-graph/patch-8x8-units-1024/sample_4.png} &
        \includegraphics[width=0.06\linewidth]{media/samples_export/IPA/imagenet64/quad-graph/patch-8x8-units-1024/sample_5.png} \\[-2pt]
        \includegraphics[width=0.06\linewidth]{media/samples_export/IPA/imagenet64/quad-graph/patch-8x8-units-1024/sample_6.png} &
        \includegraphics[width=0.06\linewidth]{media/samples_export/IPA/imagenet64/quad-graph/patch-8x8-units-1024/sample_7.png} &
        \includegraphics[width=0.06\linewidth]{media/samples_export/IPA/imagenet64/quad-graph/patch-8x8-units-1024/sample_8.png} &
        \includegraphics[width=0.06\linewidth]{media/samples_export/IPA/imagenet64/quad-graph/patch-8x8-units-1024/sample_9.png} &
        \includegraphics[width=0.06\linewidth]{media/samples_export/IPA/imagenet64/quad-graph/patch-8x8-units-1024/sample_10.png} \\[-2pt]
        \includegraphics[width=0.06\linewidth]{media/samples_export/IPA/imagenet64/quad-graph/patch-8x8-units-1024/sample_11.png} &
        \includegraphics[width=0.06\linewidth]{media/samples_export/IPA/imagenet64/quad-graph/patch-8x8-units-1024/sample_12.png} &
        \includegraphics[width=0.06\linewidth]{media/samples_export/IPA/imagenet64/quad-graph/patch-8x8-units-1024/sample_13.png} &
        \includegraphics[width=0.06\linewidth]{media/samples_export/IPA/imagenet64/quad-graph/patch-8x8-units-1024/sample_14.png} &
        \includegraphics[width=0.06\linewidth]{media/samples_export/IPA/imagenet64/quad-graph/patch-8x8-units-1024/sample_15.png} \\
\end{tabular}
    \begin{tabular}{ccccc}
        \includegraphics[width=0.06\linewidth]{media/samples_export/RePA/imagenet64/quad-graph/patch-64x64-units-512/sample_1.png} &
        \includegraphics[width=0.06\linewidth]{media/samples_export/RePA/imagenet64/quad-graph/patch-64x64-units-512/sample_2.png} &
        \includegraphics[width=0.06\linewidth]{media/samples_export/RePA/imagenet64/quad-graph/patch-64x64-units-512/sample_3.png} &
        \includegraphics[width=0.06\linewidth]{media/samples_export/RePA/imagenet64/quad-graph/patch-64x64-units-512/sample_4.png} &
        \includegraphics[width=0.06\linewidth]{media/samples_export/RePA/imagenet64/quad-graph/patch-64x64-units-512/sample_5.png} \\[-2pt]
        \includegraphics[width=0.06\linewidth]{media/samples_export/RePA/imagenet64/quad-graph/patch-64x64-units-512/sample_6.png} &
        \includegraphics[width=0.06\linewidth]{media/samples_export/RePA/imagenet64/quad-graph/patch-64x64-units-512/sample_7.png} &
        \includegraphics[width=0.06\linewidth]{media/samples_export/RePA/imagenet64/quad-graph/patch-64x64-units-512/sample_8.png} &
        \includegraphics[width=0.06\linewidth]{media/samples_export/RePA/imagenet64/quad-graph/patch-64x64-units-512/sample_9.png} &
        \includegraphics[width=0.06\linewidth]{media/samples_export/RePA/imagenet64/quad-graph/patch-64x64-units-512/sample_10.png} \\
\end{tabular}
        \begin{tabular}{ccccc}
        \includegraphics[width=0.06\linewidth]{media/samples_export/IPA/imagenet64/quad-graph/patch-64x64-units-128/sample_1.png} &
        \includegraphics[width=0.06\linewidth]{media/samples_export/IPA/imagenet64/quad-graph/patch-64x64-units-128/sample_2.png} &
        \includegraphics[width=0.06\linewidth]{media/samples_export/IPA/imagenet64/quad-graph/patch-64x64-units-128/sample_3.png} &
        \includegraphics[width=0.06\linewidth]{media/samples_export/IPA/imagenet64/quad-graph/patch-64x64-units-128/sample_4.png} &
        \includegraphics[width=0.06\linewidth]{media/samples_export/IPA/imagenet64/quad-graph/patch-64x64-units-128/sample_5.png} \\[-2pt]
        \includegraphics[width=0.06\linewidth]{media/samples_export/IPA/imagenet64/quad-graph/patch-64x64-units-128/sample_6.png} &
        \includegraphics[width=0.06\linewidth]{media/samples_export/IPA/imagenet64/quad-graph/patch-64x64-units-128/sample_7.png} &
        \includegraphics[width=0.06\linewidth]{media/samples_export/IPA/imagenet64/quad-graph/patch-64x64-units-128/sample_8.png} &
        \includegraphics[width=0.06\linewidth]{media/samples_export/IPA/imagenet64/quad-graph/patch-64x64-units-128/sample_9.png} &
        \includegraphics[width=0.06\linewidth]{media/samples_export/IPA/imagenet64/quad-graph/patch-64x64-units-128/sample_10.png} \\
\end{tabular}

    \caption{Samples for CelebA (top row), ImageNet32 (middle row) and ImageNet64 (bottom row) using \indepatch $8\times 8\times3$ and $K=1024$ (left),  \repa (center) with $K=1024$ for CelebA and $K=512$ for ImageNet64 and \fullpc with $K=128$ (right). All the PCs were built with a \textbf{QG} RG.}
    \label{fig:all-samples-qg}
\end{figure}

\begin{figure}[!t]
\setlength{\tabcolsep}{0.5pt}    
\centering
    \begin{tabular}{ccccc}
    \multicolumn{5}{c}{\indepatch}\\
        \includegraphics[width=0.06\linewidth]{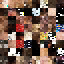} &
        \includegraphics[width=0.06\linewidth]{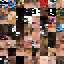} &
        \includegraphics[width=0.06\linewidth]{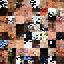} &
        \includegraphics[width=0.06\linewidth]{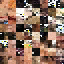} &
        \includegraphics[width=0.06\linewidth]{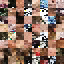} \\[-2pt]
        \includegraphics[width=0.06\linewidth]{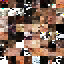} &
        \includegraphics[width=0.06\linewidth]{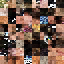} &
        \includegraphics[width=0.06\linewidth]{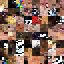} &
        \includegraphics[width=0.06\linewidth]{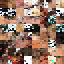} &
        \includegraphics[width=0.06\linewidth]{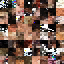} \\[-2pt]
        \includegraphics[width=0.06\linewidth]{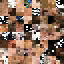} &
        \includegraphics[width=0.06\linewidth]{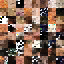} &
        \includegraphics[width=0.06\linewidth]{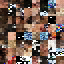} &
        \includegraphics[width=0.06\linewidth]{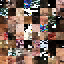} &
        \includegraphics[width=0.06\linewidth]{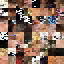} \\
\end{tabular}
    \begin{tabular}{ccccc}
    \multicolumn{5}{c}{\repa}\\
        \includegraphics[width=0.06\linewidth]{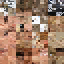} &
        \includegraphics[width=0.06\linewidth]{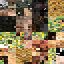} &
        \includegraphics[width=0.06\linewidth]{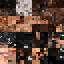} &
        \includegraphics[width=0.06\linewidth]{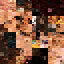} &
        \includegraphics[width=0.06\linewidth]{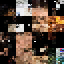} \\[-2pt]
        \includegraphics[width=0.06\linewidth]{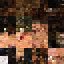} &
        \includegraphics[width=0.06\linewidth]{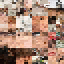} &
        \includegraphics[width=0.06\linewidth]{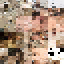} &
        \includegraphics[width=0.06\linewidth]{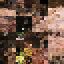} &
        \includegraphics[width=0.06\linewidth]{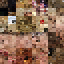} \\[-2pt]
        \includegraphics[width=0.06\linewidth]{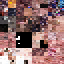} &
        \includegraphics[width=0.06\linewidth]{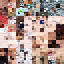} &
        \includegraphics[width=0.06\linewidth]{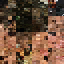} &
        \includegraphics[width=0.06\linewidth]{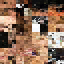} &
        \includegraphics[width=0.06\linewidth]{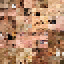} \\
\end{tabular}
        \begin{tabular}{ccccc}
        \multicolumn{5}{c}{\fullpc}\\
        \includegraphics[width=0.06\linewidth]{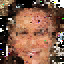} &
        \includegraphics[width=0.06\linewidth]{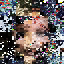} &
        \includegraphics[width=0.06\linewidth]{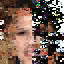} &
        \includegraphics[width=0.06\linewidth]{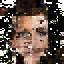} &
        \includegraphics[width=0.06\linewidth]{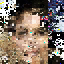} \\[-2pt]
        \includegraphics[width=0.06\linewidth]{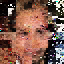} &
        \includegraphics[width=0.06\linewidth]{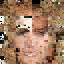} &
        \includegraphics[width=0.06\linewidth]{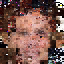} &
        \includegraphics[width=0.06\linewidth]{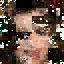} &
        \includegraphics[width=0.06\linewidth]{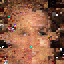} \\[-2pt]
        \includegraphics[width=0.06\linewidth]{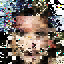} &
        \includegraphics[width=0.06\linewidth]{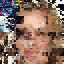} &
        \includegraphics[width=0.06\linewidth]{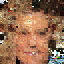} &
        \includegraphics[width=0.06\linewidth]{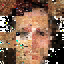} &
        \includegraphics[width=0.06\linewidth]{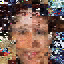} \\
\end{tabular}
        \begin{tabular}{ccccc}
        \includegraphics[width=0.06\linewidth]{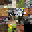} &
        \includegraphics[width=0.06\linewidth]{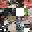} &
        \includegraphics[width=0.06\linewidth]{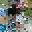} &
        \includegraphics[width=0.06\linewidth]{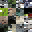} &
        \includegraphics[width=0.06\linewidth]{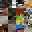} \\[-2pt]
        \includegraphics[width=0.06\linewidth]{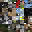} &
        \includegraphics[width=0.06\linewidth]{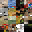} &
        \includegraphics[width=0.06\linewidth]{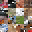} &
        \includegraphics[width=0.06\linewidth]{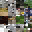} &
        \includegraphics[width=0.06\linewidth]{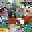} \\[-2pt]
        \includegraphics[width=0.06\linewidth]{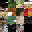} &
        \includegraphics[width=0.06\linewidth]{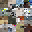} &
        \includegraphics[width=0.06\linewidth]{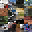} &
        \includegraphics[width=0.06\linewidth]{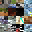} &
        \includegraphics[width=0.06\linewidth]{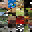} \\
\end{tabular}
    \begin{tabular}{ccccc}
        \includegraphics[width=0.06\linewidth]{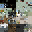} &
        \includegraphics[width=0.06\linewidth]{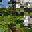} &
        \includegraphics[width=0.06\linewidth]{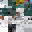} &
        \includegraphics[width=0.06\linewidth]{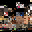} &
        \includegraphics[width=0.06\linewidth]{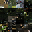} \\[-2pt]
        \includegraphics[width=0.06\linewidth]{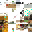} &
        \includegraphics[width=0.06\linewidth]{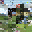} &
        \includegraphics[width=0.06\linewidth]{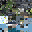} &
        \includegraphics[width=0.06\linewidth]{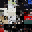} &
        \includegraphics[width=0.06\linewidth]{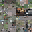} \\
\end{tabular}
        \begin{tabular}{ccccc}
        \includegraphics[width=0.06\linewidth]{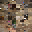} &
        \includegraphics[width=0.06\linewidth]{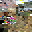} &
        \includegraphics[width=0.06\linewidth]{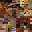} &
        \includegraphics[width=0.06\linewidth]{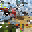} &
        \includegraphics[width=0.06\linewidth]{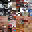} \\[-2pt]
        \includegraphics[width=0.06\linewidth]{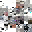} &
        \includegraphics[width=0.06\linewidth]{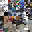} &
        \includegraphics[width=0.06\linewidth]{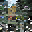} &
        \includegraphics[width=0.06\linewidth]{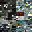} &
        \includegraphics[width=0.06\linewidth]{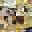} \\[-2pt]
        \includegraphics[width=0.06\linewidth]{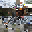} &
        \includegraphics[width=0.06\linewidth]{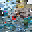} &
        \includegraphics[width=0.06\linewidth]{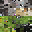} &
        \includegraphics[width=0.06\linewidth]{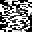} &
        \includegraphics[width=0.06\linewidth]{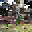} \\
\end{tabular}
        \begin{tabular}{ccccc}
        \includegraphics[width=0.06\linewidth]{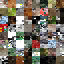} &
        \includegraphics[width=0.06\linewidth]{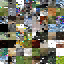} &
        \includegraphics[width=0.06\linewidth]{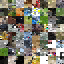} &
        \includegraphics[width=0.06\linewidth]{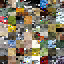} &
        \includegraphics[width=0.06\linewidth]{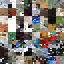} \\[-2pt]
        \includegraphics[width=0.06\linewidth]{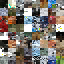} &
        \includegraphics[width=0.06\linewidth]{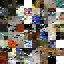} &
        \includegraphics[width=0.06\linewidth]{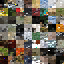} &
        \includegraphics[width=0.06\linewidth]{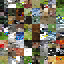} &
        \includegraphics[width=0.06\linewidth]{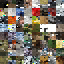} \\[-2pt]
        \includegraphics[width=0.06\linewidth]{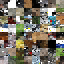} &
        \includegraphics[width=0.06\linewidth]{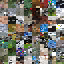} &
        \includegraphics[width=0.06\linewidth]{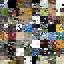} &
        \includegraphics[width=0.06\linewidth]{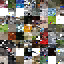} &
        \includegraphics[width=0.06\linewidth]{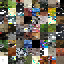} \\
\end{tabular}
    \begin{tabular}{ccccc}
        \includegraphics[width=0.06\linewidth]{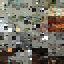} &
        \includegraphics[width=0.06\linewidth]{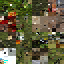} &
        \includegraphics[width=0.06\linewidth]{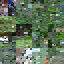} &
        \includegraphics[width=0.06\linewidth]{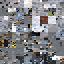} &
        \includegraphics[width=0.06\linewidth]{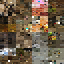} \\[-2pt]
        \includegraphics[width=0.06\linewidth]{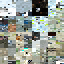} &
        \includegraphics[width=0.06\linewidth]{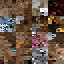} &
        \includegraphics[width=0.06\linewidth]{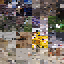} &
        \includegraphics[width=0.06\linewidth]{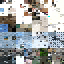} &
        \includegraphics[width=0.06\linewidth]{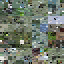} \\
\end{tabular}
        \begin{tabular}{ccccc}
        \includegraphics[width=0.06\linewidth]{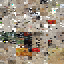} &
        \includegraphics[width=0.06\linewidth]{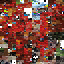} &
        \includegraphics[width=0.06\linewidth]{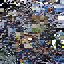} &
        \includegraphics[width=0.06\linewidth]{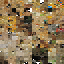} &
        \includegraphics[width=0.06\linewidth]{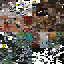} \\[-2pt]
        \includegraphics[width=0.06\linewidth]{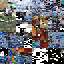} &
        \includegraphics[width=0.06\linewidth]{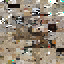} &
        \includegraphics[width=0.06\linewidth]{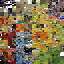} &
        \includegraphics[width=0.06\linewidth]{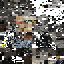} &
        \includegraphics[width=0.06\linewidth]{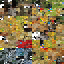} \\[-2pt]
        \includegraphics[width=0.06\linewidth]{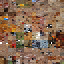} &
        \includegraphics[width=0.06\linewidth]{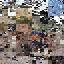} &
        \includegraphics[width=0.06\linewidth]{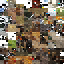} &
        \includegraphics[width=0.06\linewidth]{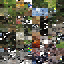} &
        \includegraphics[width=0.06\linewidth]{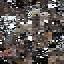} \\
\end{tabular}

    \caption{Samples for CelebA (top row), ImageNet32 (middle row) and ImageNet64 (bottom row) using \indepatch $8\times 8\times3$ and $K=1024$ (left),  \repa (center) with $K=1024$ for CelebA and $K=512$ for ImageNet64 and \fullpc with $K=128$ (right). All the PCs were built with a \textbf{QT} RG.}
    \label{fig:all-samples-qt}
\end{figure}

Note that, to improve sample quality, we take the mode of the input distribution rather than sampling from it. 
Although in this way we are not allowed to use these samples for probabilistic inference, as they are not i.i.d. from the circuit distribution, they are valid for generative modeling purposes.
In fact, this is very common practice in deep generative models like VAEs, and also common for PCs \citep{lang2022elevating,braun2025tractable}, see Appendix D of \citet{peharz2020einsum} for a discussion.
At the same time, some of the black and white artifacts we see in our samples (\cref{fig:all-samples-qg,fig:all-samples-qt}) are due to selecting the mode of the inputs.

Sampling from \repa circuits still follows \cref{alg:sampling} once we materialize the shared structure and parameters of such models.

\begin{algorithm}
\small
\SetKwComment{Comment}{/* }{ */}
\caption{\texttt{samplingTensorizedPC}$(c, N)$}
\label{alg:sampling}

\textbf{Input:} Tensorized PC $c$ over $\X = \{X_i\}_{i=1}^{D}$, number of desired samples $N$ \;
\textbf{Output:} Samples $\mathbf{S} \in \mathbb{R}^{N \times D}$ drawn from $c$. \\

\textbf{Assumptions:} (1) $c$ is normalized: all sum layer parameters sum up to 1 over the columns; (2) each input layer is defined over a single RV; (3) the width of a layer is a multiple of $K$. \\

\textbf{Notes:} (1) All assignments preceded by the symbol $\forall$ can be parallelized; (2) $\textsf{unravel-index}$ is the corresponding numpy function but whose indexing starts from 1 instead of 0.

\Begin{
    $\mathcal{S}[c] \leftarrow [N]$\;
    $\mathcal{U}[c] \leftarrow \mathbf{1}_N$\;
    \For{each inner layer $\layer \in \textsf{BFS}(c)$}{
        $\mathcal{L} \leftarrow$ input layers of $\layer$\;

        \If{$\mathcal{S}[\layer]$ is empty}{
            \textbf{continue}\\
        }
        \ElseIf{$\layer$ is a sum layer with $\W \in \mathbb{R}^{K \times KL}$}{
            $v \leftarrow$ vector of size $|\mathcal{S}[\layer]|$ in $[KL]$\\
            $v_i \sim \text{Categorical}(\W_{(\mathcal{U}[\layer])_i,:})$\\
            $(\textsf{idx1}, \textsf{idx2}) \leftarrow \textsf{unravel-index}(v, (L,K))$\\

            \For{$i \in [L]$}{
                $\mathcal{S}[\mathcal{L}[i]].\textsf{extend}(\mathcal{S}[\layer][\textsf{idx1} == i])$\\
                $\mathcal{U}[\mathcal{L}[i]].\textsf{extend}(\textsf{idx2}[\textsf{idx1} == i])$\\
            }
        }
        \ElseIf{$\layer$ is Hadamard product layer}{
            \For{$i \in [L]$}{
                $\mathcal{S}[\mathcal{L}[i]].\textsf{extend}(\mathcal{S}[\layer])$\\
                $\mathcal{U}[\mathcal{L}[i]].\textsf{extend}(\mathcal{U}[\layer])$\\
            }
        }
    }
    $\mathbf{S} \leftarrow \mathbb{R}^{N \times D}$\\

    \For{each input layer $\layer \in c$ s.t. $\mathcal{S}[\layer] \neq [\,]$}{
        $j \leftarrow \scope(\layer)$\\
        $\textsf{pairs} \leftarrow \textsf{vstack}(\mathcal{S}[\layer], \mathcal{U}[\layer])$\\

        \For{$(i,k) \in \textsf{pairs}$}{
            $\mathbf{S}_{ij} \leftarrow$ sample from or take the mode of the $k$-th unit of $\layer$\\
        }
    }
    \Return{$\mathbf{S}$}\;
}
\end{algorithm}

\clearpage
\section{Parallels with other methods}

\subsection{Monarch matrices}
\label{app:monarch}

A \emph{Monarch matrix} \citep{dao_monarch_2022} factorizes a matrix
$\mathbf{W} \in \mathbb{R}^{m_1 m_2 \times n_1 n_2}$ as
\[
  \mathbf{W} = \mathbf{P}_B \, \mathsf{blockdiag}(\mathbf{B}_1,\dots,\mathbf{B}_{m_1}) \, \mathbf{P}_A \, \mathsf{blockdiag}(\mathbf{A}_1,\dots,\mathbf{A}_{n_2}),
\]
where $\mathbf{A}_k \in \mathbb{R}^{m_1 \times n_1}$ for $k=1,\dots,n_2$,
$\mathbf{B}_k \in \mathbb{R}^{m_2 \times n_2}$ for $k=1,\dots,m_1$,
and $\mathbf{P}_A$, $\mathbf{P}_B$ are fixed permutation matrices.
Specifically, $\mathbf{P}_A$ treats a vector of length $n_2 m_1$ as an $(n_2 \times m_1)$
matrix stored in row-major order and transposes it in column-major
order (i.e.\ transposes it to $m_1 \times n_2$).
$\mathbf{P}_B$ does the same for a vector of length $m_1 m_2$: it views it as
an $(m_1 \times m_2)$ matrix and transposes it to $(m_2 \times m_1)$.
In both cases the permutation is parameter-free and amounts to a
\texttt{reshape} followed by a \texttt{transpose}.

A monarch matrix reduces the parameter count from $m_1 m_2 n_1 n_2$ to
$n_2 m_1 n_1 + m_1 m_2 n_2$.
When $\mathbf{W}$ is monarch, the product $\mathbf{y} = \mathbf{W}\mathbf{x}$ can be computed efficiently without forming $\mathbf{W}$.
Specifically, one needs to reshape $\mathbf{x} \in \mathbb{R}^{n_1 n_2}$ into a matrix
$\mathbf{X} \in \mathbb{R}^{n_2 \times n_1}$ and compute two batched
multiplications:
\[
  \mathbf{Z}_{k} = \sum_{j} \mathbf{A}_{k,j}\, \mathbf{X}_{k,j}, \qquad
  \mathbf{Y}_{k} = \sum_{j} \mathbf{B}_{k,j}\, \mathbf{Z}_{k,j},
\]
or, in Einstein notation,
$\mathbf{Z}_{i,n} = \sum_j \mathbf{A}_{n,i,j}\, \mathbf{X}_{n,j}$ followed by
$\mathbf{Y}_{i,n} = \sum_j \mathbf{B}_{n,i,j}\, \mathbf{Z}_{n,j}$,
after which we flatten $\mathbf{Y}$, i.e.\ $\mathbf{y} = \operatorname{vec}(\mathbf{Y})$.
Each step costs $O(m_1 n_1 n_2)$ and $O(m_1 m_2 n_2)$ respectively,
giving an overall complexity of $O\!\bigl(m_1 n_2 (n_1 + m_2)\bigr)$
versus $O(m_1 m_2 n_1 n_2)$ for a dense matrix--vector product.

\subsection{Factorized CP layers}
\label{app:factorized_cp}
As briefly illustrated in \cref{fig:nopa}, we can group layers based on their topological depth into groups called \textit{folds}. For example, a folded CP layer would compute:
$$\layer^{(n)}(\x) =\bigodot_{i=1}^{H} \W^{(n)}_i \layer^{(n)}_i(\x_i)$$ 
Where $F$ is the \textit{fold dimension}, $n\in [F]$ and $\W^{(n)}_i\in\mathbb{R}^{F\times K\times K}$. While this grouping allows frameworks to parallelize PCs more efficiently on GPUs, prior work has already attempted parameter sharing by exploiting the \textit{fold dimension}. \citet{loconte2025what} proposed a new layer, CP$^{XS}$, to use the same weights matrix across the fold when using CP layers:
$$\layer^{(n)}(\x) =\bigodot_{i=1}^{H} \W_i \layer^{(n)}_i(\x_i)$$ 
Where $\W_i\in\mathbb{R}^{K\times K}$. While this setting is clearly reminiscent of \nopa, we note a significant difference. \nopa shares across the fold dimension of the sum layers directly, regardless of their original sum-product abstraction. For example, a CP layer with \nopa uses a single  $\W\in\mathbb{R}^{K\times K}$ matrix because the CP layer is decomposed into $H$ sums followed by a single product:
$$\layer^{(n)}(\x) =\bigodot_{i=1}^{H} \W \layer^{(n)}_i(\x_i)$$ 
This also makes our formulation more general as it operates on the operations themselves and not the abstraction that creates them.

\clearpage
\section{Experimental details}
\label{app:exp-details}

\subsection{Hardware \& software specification}
\label{app:hw}

Every experiment was run using the Cirkit library \citep{The_APRIL_Lab_cirkit_2024} with PyTorch \citep{paszke_pytorch_2019} as the backend. We used Lightning \cite{falcon_pytorchlightningpytorch-lightning_2020} to handle our training pipelines.

\cref{app:tab:hardware_req} specifies the hardware used for every parameter grid, grouped by dataset, class and region graph. For the Fubini cluster, we did not put any restriction on the amount of RAM and the number of CPUs used as it does not use any scheduling or resource allocation. The raw result folder of each single run contains a \texttt{summary.yaml} file that specifies the exact GPU specification and the name of the nodes used.

\begin{table}[]
    \centering
    \caption{\textbf{Hardware used for training experiments} detailed in \cref{app:tab:hyper_grid_ipa}, \cref{app:tab:hyper_grid_repa}, \cref{app:tab:hyper_grid_repa_clt}. We report either \textit{reserved} resources for SLURM jobs or \textit{available} resources when not using a SLURM cluster. }
    \begin{tabular}{l l l c c c c}
    \toprule
    
      Dataset  & Class & RG & \#CPUs & \#GPU & GPU Types & RAM  \\
     \midrule    
      
    ImageNet32 & \indepatch & [CLT, QG, QT-2] &  16 & 1 & [L40s, H200] &128GB\\
    ImageNet32 & \repa & [QG, QT-2] &  20 & 1 & [A100] &120GB\\
    ImageNet32 & \repa & [CLT] &  127 & 1 & [RTX A6000] &504GB\\

    ImageNet64 & \indepatch & [CLT, QG, QT-2] & 16 & 1 & [L40s, H200] &128GB\\
    ImageNet64 & \repa & [QG, QT-2] &  16 & 1 & [L40s, H200] &128GB\\
    ImageNet64 & \repa & [CLT] &  127 & 1 & [RTX A6000] &504GB\\

    CelebA & \indepatch & [CLT, QG, QT-2] &  16 & 1 & [L40s, H200] &128GB\\
    CelebA & \repa & [QG, QT-2] &  16 & 1 & [L40s, H200] &128GB\\
    CelebA & \repa & [CLT] &  127 & 1 & [RTX A6000] &504GB\\
    
    \bottomrule
    \end{tabular}
    \label{app:tab:hardware_req}
\end{table}

\subsection{Dataset}
\label{app:dataset}
\paragraph{Data Split.} For each image dataset, we create a new train and validation set by splitting the original train set and we keep the original validation set for testing. For tabular data, we refer to \citet{scassola_sobering_2026}.
\begin{table}[h]
    \centering
    \caption{\textbf{Train, validation and test splits for each dataset.}}
    \begin{tabular}{l c c c}
    \toprule
         Dataset &   Train & Validation & Test \\
     \midrule    
         ImageNet32 &  960,875 & 320,292 & 50,000\\
         ImageNet64 &  1,217,108 & 64,059 & 50,000\\
         CelebA & 162,770 & 19,867 & 19,962\\
    \bottomrule
    \end{tabular}
    \label{app:tab:split_per_dataset}
\end{table}

\paragraph{CelebA.} Following \citet{gala2024scaling}, we obtain the $64\times64\times3$ version of CelebA by first cropping the center of the image to a shape of $140\times140\times3$ and then resizing it to the final $64\times64\times3$.

\subsection{Hyperparameters}
\label{app:hyperparams}

\begin{table}[h]
    \centering
    \caption{\textbf{Dataset-dependent hyperparameters.} We use different effective batch size and early stopping delta values to balance quality and training time.}
    
    \begin{tabular}{l c c}
    \toprule
         Dataset &   \makecell{Effective \\ Batch Size} & \makecell{Early Stopping\\ Delta} \\
     \midrule    
         ImageNet32 &  20,000 & 10\\
         ImageNet64 &  20,000 & 30\\
         CelebA & 5,000 & 10\\
    \bottomrule
    \end{tabular}
    \label{app:tab:hyperparams_per_dataset}
\end{table}

We first ran an experiment to test the \indepparadox. We detail each experiment for \indepatch in \cref{app:tab:hyper_grid_ipa}.
\begin{table}[h]
    \centering
    \caption{\textbf{Hyperparameter grid used for \indepatch}, values in brackets indicate each value that this parameter can take. We train the full grid of all possible combinations.}
    
    \begin{tabular}{l l c c c c}
    \toprule
        Class & Dataset &  Layer type & $K$ & RG & $D$ \\
     \midrule    
         \indepatch & ImageNet32 &  [CP, CP-T] & [128,256,512,1024] & [QT-2, QG, CLT] & [2,4,8,16,32] \\
         \indepatch & ImageNet64 &  [CP-T] & [128,256,512,1024] & [QT-2, QG, CLT] & [2,4,8,16,32,64] \\
         \indepatch & CelebA &  [CP-T] & [128,256,512,1024] & [QT-2, QG, CLT] & [2,4,8,16,32,64] \\
         \indepatch & ImageNet32 &  [Mona] & [1024, 4096] & [QT-2, QG] & [8,16,32] \\
         
    \bottomrule
    \end{tabular}
  
    \label{app:tab:hyper_grid_ipa}
\end{table}

We define a similar grid for \repa (or \indepatch(\repa)) in \cref{app:tab:hyper_grid_repa} using only QT and QG as they do not require multi-stage training. We separately give the configurations used for each stage when doing \repa for CLT RG in \cref{app:tab:hyper_grid_repa_clt}.

\begin{table}[h]
    \centering
    \caption{\textbf{Hyperparameter grid used for \repa}. When $D^t$ is smaller than the image $S$, $t$ being the recursion level, we use \indepatch(\repa).}
    
    \begin{tabular}{l l c c c c c}
    \toprule
        Class & Dataset &  Layer type & $K$ & RG & $D$ & $D^t$ \\
     \midrule    
         \repa & ImageNet32 &  [CP, CP-T] & [512,1024,2048] & [QT, QG] & [2] & [2,4,8,16,32]\\
         \repa & ImageNet64 &  [CP-T] & [512,1024,2048,4096] & [QT, QG] & [2]& [2,4,8,16,32,64]\\
         \repa & CelebA &  [CP-T] & [512,1024,2048,4096] & [QT, QG] & [2]& [2,4,8,16,32,64]\\
         \repa & ImageNet32 &  [Mona] & [1024, 4096] & [QT, QG] & [2] & [8,16,32]\\

    \bottomrule
    \end{tabular}
  
    \label{app:tab:hyper_grid_repa}
\end{table}
\begin{table}[h]
    \centering
    \caption{\textbf{Hyperparameter grid used for \repa on CLT.} When $D^t$ is smaller than the image $S$, $t$ being the recursion level, we use \indepatch(\repa). Each row represents the configuration for one of the three circuits that constitute the final \repa.}
    
    \begin{tabular}{l l l c c c c c}
    \toprule
        $c_i$& Class & Dataset &  Layer type & $K$ & RG & $D$ & $D^t$ \\
     \midrule    
         $c_1$ & \repa & ImageNet32 &  [CP-T] & [512,1024] & [CLT] & 4 & 4\\
         $c_2$ & \repa & ImageNet32 &  [CP-T] & [512,1024] & [CLT] & 4 & 16\\
         $c_3$ & \repa & ImageNet32 &  [CP-T] & [512,1024] & [CLT] & 2 & 32\\
     \midrule    
         $c_1$ & \repa & ImageNet64 &  [CP-T] & [512,1024] & [CLT] & 4 & 4\\
         $c_2$ & \repa & ImageNet64 &  [CP-T] & [512,1024] & [CLT] & 4 & 16\\
         $c_3$ & \repa & ImageNet64 &  [CP-T] & [512,1024] & [CLT] & 4 & 64\\
     \midrule    
         $c_1$ & \repa & CelebA &  [CP-T] & [512,1024] & [CLT] & 4 & 4\\
         $c_2$ & \repa & CelebA &  [CP-T] & [512,1024] & [CLT] & 4 & 16\\
         $c_3$ & \repa & CelebA &  [CP-T] & [512,1024] & [CLT] & 4 & 64\\
    \bottomrule
    \end{tabular}
  
    \label{app:tab:hyper_grid_repa_clt}
\end{table}
For our experiments on \textsc{TabPC} \cite{scassola_sobering_2026}, we designed a hyperparameter grid (see \cref{app:tab:hyper_grid_tabpc}) that includes the best configurations found in the paper. We used the same grid for \fullpc and \nopa.
Following the original paper, we trained using RAdam \cite{liu_variance_2021} and early stopping with a patience of 10 epochs and no minimum delta. When in use, the ReduceLROnPlateau scheduler uses a patience of 0, a factor of 0.85 and a minimum learning rate of $2\times10^{-4}$. We used the same seven UCI datasets: \textit{Adult} \cite{adult_2}, \textit{Beijing} \cite{beijing_pm2.5_381}, \textit{Default} \cite{default_of_credit_card_clients_350}, \textit{Diabetes} \cite{diabetes_130-us_hospitals_for_years_1999-2008_296}, \textit{Magic} \cite{magic_gamma_telescope_159}, \textit{News} \cite{online_news_popularity_332}, and \textit{Shoppers} \cite{online_shoppers_purchasing_intention_dataset_468}.

\begin{table}[h]
    \centering
    \caption{\textbf{Hyperparameter grid used for tabular data}}
    
    \begin{tabular}{l c}
    \toprule
    Parameter & Possible values\\
    \midrule
     Layer type & CP, CP-T\\
     $K$ &  1024, 2048, 4096, 8192 \\
     Learning rate & $0.1$, $0.25$, $0.5$ \\
     Scheduler & ReduceLROnPlateau, None\\
     Batch size & 256, 512 \\
    \bottomrule
    \end{tabular}
  
    \label{app:tab:hyper_grid_tabpc}
\end{table}

\section{Additional Results}
\label{app:more-results}

\subsection{Significance of results}

The results presented in this paper do not use multiple seeds and use a single learning rate / batch size configuration based on a  preliminary hyperparameter optimization which was done across all patch sizes for IPA on ImageNet32 (including FullPC). To address potential concerns regarding the significance of these results and the variance when changing hyperparameters, we performed three experiments where we changed the seed, the learning rate, and the batch size, showing minimal change.

\paragraph{Seeds} We ran 6 seeds of the best configuration for each class (\indepatch/\repa/\fullpc) on ImageNet32. \cref{app:tab:seeds} shows tiny standard deviations for all configurations. This strongly supports that the model rankings are robust to different random seeds. Furthermore, we did a similar analysis using the default TabPC configuration \citep{scassola_sobering_2026} with similar results. \cref{app:tab:seeds_tabpc} shows small standard deviations across seeds.

\begin{table}[h]
    \centering
    \caption{Variation of our best models using a quad-graph over different random seeds on ImageNet32. }
    
    \begin{tabular}{c c c c}
    \toprule
      class   & test bpd (mean) &  test bpd (std) & count\\
      \midrule
      \fullpc &	5.347784 	&0.001272 &6\\
      \indepatch 	&5.200556 	&0.000694 	&6\\
\repa(K=1024) 	&5.053292 	&0.005864 	&6\\
      \bottomrule
    \end{tabular}
    \label{app:tab:seeds}
\end{table}

\begin{table}[h]
    \centering
    \caption{Variation of our TabPC on different random seeds on the Adult dataset.}
    
    \begin{tabular}{c c c c}
    \toprule
      class   & test bpd (mean) &  test bpd (std) & count\\
      \midrule
      \fullpc &	1.097372 	&0.001769 &6\\
      \nopa 	& 1.099097 	&0.003449 	&6\\
      \bottomrule
    \end{tabular}
    \label{app:tab:seeds_tabpc}
\end{table}

\paragraph{Learning rate}
We tried 6 different learning rates on the best QG configurations for our three model classes (IPA, RePA, FullPC) on ImageNet32. \cref{app:tab:lr} shows a clear downward trend where the minimum is systematically obtained when the learning rate is 0.05, the value we used.

\begin{table}[h]
    \centering
    \caption{Variation of our best models using a quad-graph over different learning rates on ImageNet32. }
    
    \begin{tabular}{r l l l l l l}
    \toprule
    lr & 0.0001 & 0.0005 & 0.001 & 0.0050 & 0.01 & 0.05\\
      \midrule
      \fullpc & Timed out (T/O) & T/O & 5.70 & 5.57 & 5.50 & 5.35\\
      \indepatch & 6.89 & 5.72 & 5.66 & 5.49 & 5.38 & 5.20\\
\repa(K=1024) 	& T/O & T/O & T/O & 5.21 & 5.15 & 5.04\\
      \bottomrule
    \end{tabular}
    \label{app:tab:lr}
\end{table}

We further note that previous works highlighted how a single learning rate value was the best across other hyperparameters such as patch size and number of sum units, and even across datasets \citep{loconte2025what}.

\paragraph{Batch size}
Regarding our EM vs Adam comparison, we found during our initial hyperparameter research that Adam with a large batch size got better results than EM. As an illustration, \cref{app:tab:batch_size} shows how the bpd changes as we scale the batch size for our best \fullpc configuration on ImageNet32, we realized that, for our models, Adam with a very high batch size was the best configuration.

\begin{table}[h]
    \centering
    \caption{Variation of batch size on our best \fullpc model while using Adam or EM. The model uses a quad-graph, CP-T layers and $K=512$. }
    
    \begin{tabular}{r l l l l }
    \toprule
    batch size & 512 & 1024 & 8192 & 16384\\
      \midrule
      \fullpc(Adam) &5.40 & 5.40 & 5.37 & 5.35\\
      \fullpc(EM) & 6.88  & 6.60 & 5.85 &5.59\\
      \bottomrule
    \end{tabular}
    \label{app:tab:batch_size}
\end{table}

\subsection{Dropping parameter sharing in \indepatch} \label{app:ipa_noshare}

In the paper, we assume that \indepatch processes all independent patches using the same circuit; in essence, this is parameter sharing. In this section, we briefly try to disentangle the effects of the independent-patch assumption and parameter sharing. To do so, we define \indepatch(\textsc{NoShare}), a variant of \indepatch in which each patch uses its own PC.
We ran an experiment comparing these two settings on ImageNet32 for different patch sizes, with the configuration $K=512$, QG, CP-T. We did not use $K=1024$ because \indepatch(\textsc{NoShare}) requires too much memory. \cref{tab:ipa_noshare} shows that \indepatch(\textsc{NoShare}) is marginally worse than \indepatch. Parameter sharing does not seem to have a significant impact on the independence assumption conundrum, although it still delivers much more efficient models.

\begin{table}[h!]
    \centering
        \caption{\textbf{Dropping parameter sharing for \indepatch does not improve test bpds}, although it improves theoretical expressivity.}
    \begin{tabular}{l c c c c c}
    \toprule
        patch size & 2 & 4 & 8 & 16 & 32\\
        \midrule
    \indepatch & 5.605210 & 5.328382 & 5.277076 & 5.293465 & 5.346852\\
    \indepatch(\textsc{NoShare}) & 5.611599 & 5.334445 & 5.278990 & 5.293443 & 5.346852\\
    \bottomrule
    \end{tabular}

    \label{tab:ipa_noshare}
\end{table}

\subsection{The independent patch conundrum on PICs and SOS PCs}
\label{app:pics-sos}

We test the \indepparadox on sum of squares (SOS) \cite{loconte_sum_2025} circuits and probabilistic integral circuits (PIC) \cite{gala2024pic} to see if this variant also presents the same phenomenon. We trained different variations of both circuits on QT using the \texttt{cirkit} library. We used $K\in[128,256,512,1024]$ and we compared CP and CP-T layers. For each model, we used the optimization hyperparameters reported in their respective article. For SOS, we used Adam with a learning rate of $10^{-2}$, a weight decay of $10^{-2}$ and a batch size of $128$. For PIC, we used a cosine annealing scheduler with warm restarts every 500 steps going from $5\times10^{-3}$ to $10^{-4}$ with a batch size of $256$.

\begin{figure}
    \centering
    \includegraphics[width=0.5\linewidth]{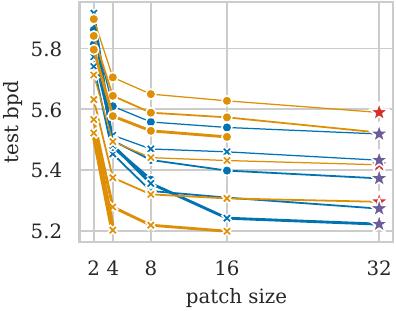}
    \includegraphics[width=0.22\linewidth]{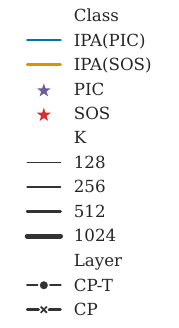}
    
    \caption{\textbf{Sum of squares (SOS) PC and probabilistic integral circuits (PIC) also show signs of plateau.} All PCs are built using the QT region graph.}
    \label{app:fig:sos_vs_pics}
\end{figure}

\cref{app:fig:sos_vs_pics} shows that both models seem to be more resilient than normal PC although they do show diminishing returns.

\subsection{The independent patch conundrum across optimizers}
\label{app:em-adam}

\begin{figure}
    \centering
    \includegraphics[width=0.3\linewidth]{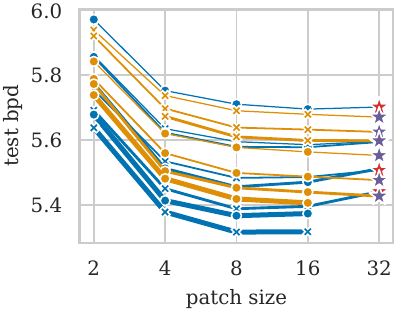}
    \includegraphics[width=0.3\linewidth]{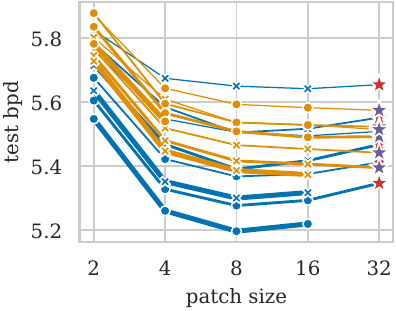}
    \includegraphics[width=0.3\linewidth]{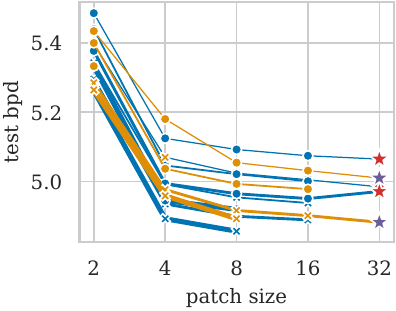}
    
    \includegraphics[width=0.8\linewidth]{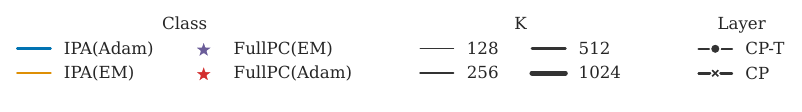}
    
    \caption{\textbf{Both Adam and EM can be affected by the \indepparadox when optimizing PCs over different patch sizes.} We note that Adam provides as good a BPD as, or better than, EM when we use the same effective batch size and learning rate scheduler. Lines in the plot that do not have points for all patch sizes indicate that we could not fit the corresponding circuit in memory.}
    \label{app:fig:em_vs_adam}
\end{figure}

We test the \indepparadox across two optimizers using Adam and expectation maximization (EM), implemented for \texttt{cirkit}. We trained the same configurations that were previously trained with Adam --- see \cref{app:hyperparams}. Following \citet{zhang_scaling_2025} we use the same cosine annealing scheduler with warm restart, starting with a step size of 1 and decreasing to 0, restarting every 20 epochs. \cref{app:fig:em_vs_adam} shows that EM does not get worse as the patch size increases, although it displays diminishing returns, but for the same architecture it often reaches worse bpd than the same PC using Adam. 

\subsection{Efficiency of \repa compared to \indepatch}

\begin{figure}
    \centering
    \includegraphics[width=0.32\linewidth]{media/efficiency_scatter_pdf/Imagenet_32__train_step_time__vs__test_bpd.pdf}
    \hfill
    \includegraphics[width=0.32\linewidth]{media/efficiency_scatter_pdf/Imagenet_32__max_train_memory__vs__test_bpd.pdf}
    \hfill
    \includegraphics[width=0.32\linewidth]{media/efficiency_scatter_pdf/Imagenet_32__flops_train__vs__test_bpd.pdf}

     \includegraphics[width=0.32\linewidth]{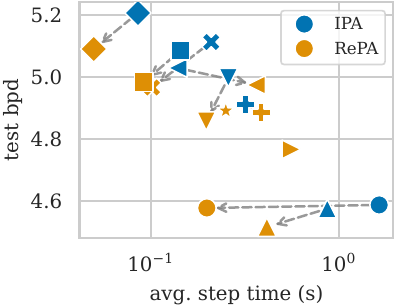}
    \hfill
    \includegraphics[width=0.32\linewidth]{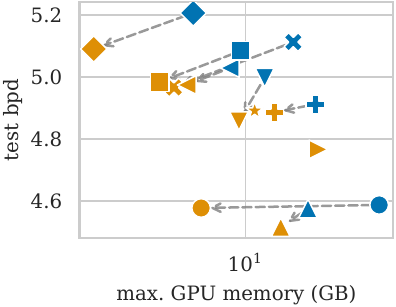}
    \hfill
    \includegraphics[width=0.32\linewidth]{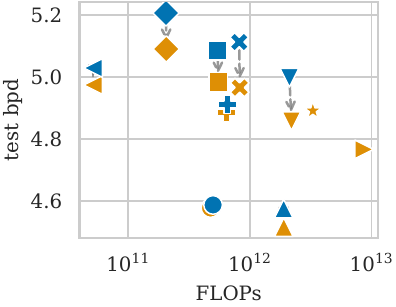}

 \includegraphics[width=0.32\linewidth]{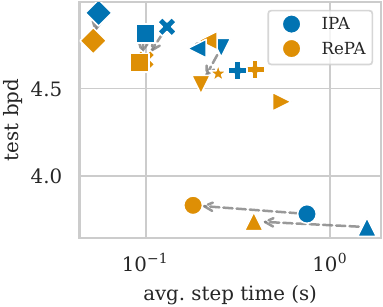}
    \hfill
    \includegraphics[width=0.32\linewidth]{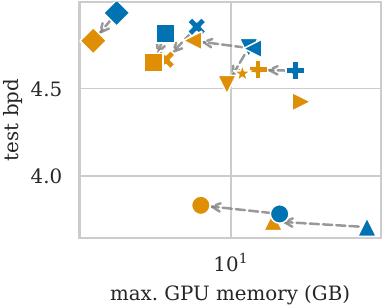}
    \hfill
    \includegraphics[width=0.32\linewidth]{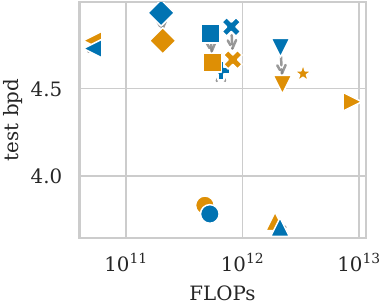}
    
    \includegraphics[width=0.8\linewidth]{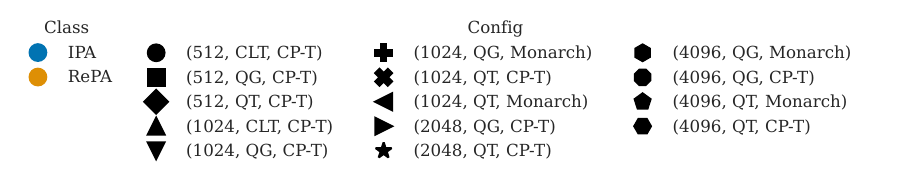}
    
    \caption{\textbf{\repa models are more Pareto efficient for all datasets tested}. We compare bpd with regards to training step time (right), GPU memory (center) and FLOPs (left) on ImageNet 32 (top), ImageNet 64 (middle) and CelebA (bottom).}
    \label{app:fig:bpd_vs_efficiency}
\end{figure}

\cref{app:fig:bpd_vs_efficiency} is the extended version of \cref{fig:fig-1} with all datasets represented and the detailed legend.

\subsection{Full results: bpd vs patch size}

\cref{app:tab:full_bpd_res} compiles the density estimation results for every configuration presented in the paper. 
    {\scriptsize

\begin{longtable}{l l c c c c c | l l c c c c c}
\caption{\textbf{Complete density estimation results} for every model configuration presented in the paper. We exclude PICs and SOS as they are only presented in appendix.}\label{app:tab:full_bpd_res}\\
\toprule
Dataset & Class & $K$ & Layer type & RG & $D$ & Test bpd & Dataset & Class & $K$ & Layer type & RG & $D$ & Test bpd \\
\midrule
\endfirsthead
\toprule
Dataset & Class & $K$ & Layer type & RG & $D$ & Test bpd & Dataset & Class & $K$ & Layer type & RG & $D$ & Test bpd \\
\midrule
\endhead
\midrule
\multicolumn{14}{r}{Continued on next page} \\
\midrule
\endfoot
\bottomrule
\endlastfoot
CelebA & IPA & 128 & CP-T & CLT & 2 & 4.70 & Imagenet 32 & IPA & 512 & CP-T & QT & 16 & 5.39 \\
CelebA & IPA & 128 & CP-T & CLT & 4 & 4.19 & Imagenet 32 & IPA & 512 & CP-T & QT & 32 & 5.44 \\
CelebA & IPA & 128 & CP-T & CLT & 8 & 4.06 & Imagenet 32 & IPA & 1024 & CP & CLT & 2 & 5.33 \\
CelebA & IPA & 128 & CP-T & CLT & 16 & 4.02 & Imagenet 32 & IPA & 1024 & CP & CLT & 4 & 4.95 \\
CelebA & IPA & 128 & CP-T & CLT & 32 & 3.99 & Imagenet 32 & IPA & 1024 & CP & CLT & 8 & 4.92 \\
CelebA & IPA & 128 & CP-T & QG & 2 & 5.38 & Imagenet 32 & IPA & 1024 & CP & QG & 2 & 5.64 \\
CelebA & IPA & 128 & CP-T & QG & 4 & 5.12 & Imagenet 32 & IPA & 1024 & CP & QG & 4 & 5.35 \\
CelebA & IPA & 128 & CP-T & QG & 8 & 5.07 & Imagenet 32 & IPA & 1024 & CP & QG & 8 & 5.30 \\
CelebA & IPA & 128 & CP-T & QG & 16 & 5.06 & Imagenet 32 & IPA & 1024 & CP & QG & 16 & 5.32 \\
CelebA & IPA & 128 & CP-T & QG & 32 & 5.07 & Imagenet 32 & IPA & 1024 & CP & QT & 2 & 5.68 \\
CelebA & IPA & 128 & CP-T & QG & 64 & 5.06 & Imagenet 32 & IPA & 1024 & CP & QT & 4 & 5.41 \\
CelebA & IPA & 128 & CP-T & QT & 2 & 5.46 & Imagenet 32 & IPA & 1024 & CP & QT & 8 & 5.37 \\
CelebA & IPA & 128 & CP-T & QT & 4 & 5.21 & Imagenet 32 & IPA & 1024 & CP & QT & 16 & 5.37 \\
CelebA & IPA & 128 & CP-T & QT & 8 & 5.16 & Imagenet 32 & IPA & 1024 & CP-T & CLT & 2 & 5.26 \\
CelebA & IPA & 128 & CP-T & QT & 16 & 5.15 & Imagenet 32 & IPA & 1024 & CP-T & CLT & 4 & 4.89 \\
CelebA & IPA & 128 & CP-T & QT & 32 & 5.16 & Imagenet 32 & IPA & 1024 & CP-T & CLT & 8 & 4.86 \\
CelebA & IPA & 128 & CP-T & QT & 64 & 5.14 & Imagenet 32 & IPA & 1024 & CP-T & QG & 2 & 5.55 \\
CelebA & IPA & 256 & CP-T & CLT & 2 & 4.63 & Imagenet 32 & IPA & 1024 & CP-T & QG & 4 & 5.26 \\
CelebA & IPA & 256 & CP-T & CLT & 4 & 4.06 & Imagenet 32 & IPA & 1024 & CP-T & QG & 8 & 5.20 \\
CelebA & IPA & 256 & CP-T & CLT & 8 & 3.93 & Imagenet 32 & IPA & 1024 & CP-T & QG & 16 & 5.22 \\
CelebA & IPA & 256 & CP-T & CLT & 16 & 3.88 & Imagenet 32 & IPA & 1024 & CP-T & QT & 2 & 5.64 \\
CelebA & IPA & 256 & CP-T & CLT & 32 & 3.86 & Imagenet 32 & IPA & 1024 & CP-T & QT & 4 & 5.38 \\
CelebA & IPA & 256 & CP-T & QG & 2 & 5.25 & Imagenet 32 & IPA & 1024 & CP-T & QT & 8 & 5.32 \\
CelebA & IPA & 256 & CP-T & QG & 4 & 4.96 & Imagenet 32 & IPA & 1024 & CP-T & QT & 16 & 5.32 \\
CelebA & IPA & 256 & CP-T & QG & 8 & 4.92 & Imagenet 32 & RePA & 512 & CP & QG & 2 & 5.68 \\
CelebA & IPA & 256 & CP-T & QG & 16 & 4.93 & Imagenet 32 & RePA & 512 & CP & QG & 4 & 5.38 \\
CelebA & IPA & 256 & CP-T & QG & 32 & 4.96 & Imagenet 32 & RePA & 512 & CP & QG & 8 & 5.27 \\
CelebA & IPA & 256 & CP-T & QT & 2 & 5.36 & Imagenet 32 & RePA & 512 & CP & QG & 16 & 5.25 \\
CelebA & IPA & 256 & CP-T & QT & 4 & 5.08 & Imagenet 32 & RePA & 512 & CP & QG & 32 & 5.24 \\
CelebA & IPA & 256 & CP-T & QT & 8 & 5.03 & Imagenet 32 & RePA & 512 & CP & QT & 2 & 5.76 \\
CelebA & IPA & 256 & CP-T & QT & 16 & 5.03 & Imagenet 32 & RePA & 512 & CP & QT & 4 & 5.45 \\
CelebA & IPA & 256 & CP-T & QT & 32 & 5.06 & Imagenet 32 & RePA & 512 & CP & QT & 8 & 5.35 \\
CelebA & IPA & 512 & CP-T & CLT & 2 & 4.59 & Imagenet 32 & RePA & 512 & CP & QT & 16 & 5.35 \\
CelebA & IPA & 512 & CP-T & CLT & 4 & 3.95 & Imagenet 32 & RePA & 512 & CP & QT & 32 & 5.34 \\
CelebA & IPA & 512 & CP-T & CLT & 8 & 3.82 & Imagenet 32 & RePA & 512 & CP-T & CLT & 4 & 4.95 \\
CelebA & IPA & 512 & CP-T & CLT & 16 & 3.78 & Imagenet 32 & RePA & 512 & CP-T & CLT & 16 & 4.87 \\
CelebA & IPA & 512 & CP-T & QG & 2 & 5.19 & Imagenet 32 & RePA & 512 & CP-T & CLT & 32 & 4.85 \\
CelebA & IPA & 512 & CP-T & QG & 4 & 4.86 & Imagenet 32 & RePA & 512 & CP-T & QG & 2 & 5.60 \\
CelebA & IPA & 512 & CP-T & QG & 8 & 4.81 & Imagenet 32 & RePA & 512 & CP-T & QG & 4 & 5.27 \\
CelebA & IPA & 512 & CP-T & QG & 16 & 4.83 & Imagenet 32 & RePA & 512 & CP-T & QG & 8 & 5.19 \\
CelebA & IPA & 512 & CP-T & QT & 2 & 5.26 & Imagenet 32 & RePA & 512 & CP-T & QG & 16 & 5.17 \\
CelebA & IPA & 512 & CP-T & QT & 4 & 4.98 & Imagenet 32 & RePA & 512 & CP-T & QG & 32 & 5.16 \\
CelebA & IPA & 512 & CP-T & QT & 8 & 4.93 & Imagenet 32 & RePA & 512 & CP-T & QT & 2 & 5.70 \\
CelebA & IPA & 512 & CP-T & QT & 16 & 4.94 & Imagenet 32 & RePA & 512 & CP-T & QT & 4 & 5.39 \\
CelebA & IPA & 1024 & CP-T & CLT & 2 & 4.55 & Imagenet 32 & RePA & 512 & CP-T & QT & 8 & 5.31 \\
CelebA & IPA & 1024 & CP-T & CLT & 4 & 3.88 & Imagenet 32 & RePA & 512 & CP-T & QT & 16 & 5.28 \\
CelebA & IPA & 1024 & CP-T & CLT & 8 & 3.74 & Imagenet 32 & RePA & 512 & CP-T & QT & 32 & 5.26 \\
CelebA & IPA & 1024 & CP-T & CLT & 16 & 3.71 & Imagenet 32 & RePA & 1024 & CP & QG & 2 & 5.63 \\
CelebA & IPA & 1024 & CP-T & QG & 2 & 5.12 & Imagenet 32 & RePA & 1024 & CP & QG & 4 & 5.31 \\
CelebA & IPA & 1024 & CP-T & QG & 4 & 4.78 & Imagenet 32 & RePA & 1024 & CP & QG & 8 & 5.21 \\
CelebA & IPA & 1024 & CP-T & QG & 8 & 4.73 & Imagenet 32 & RePA & 1024 & CP & QG & 16 & 5.16 \\
CelebA & IPA & 1024 & CP-T & QT & 2 & 5.22 & Imagenet 32 & RePA & 1024 & CP & QG & 32 & 5.15 \\
CelebA & IPA & 1024 & CP-T & QT & 4 & 4.91 & Imagenet 32 & RePA & 1024 & CP & QT & 2 & 5.69 \\
CelebA & IPA & 1024 & CP-T & QT & 8 & 4.85 & Imagenet 32 & RePA & 1024 & CP & QT & 4 & 5.36 \\
CelebA & IPA & 1024 & CP-T & QT & 16 & 4.87 & Imagenet 32 & RePA & 1024 & CP & QT & 8 & 5.28 \\
CelebA & RePA & 512 & CP-T & CLT & 4 & 3.95 & Imagenet 32 & RePA & 1024 & CP & QT & 16 & 5.24 \\
CelebA & RePA & 512 & CP-T & CLT & 16 & 3.85 & Imagenet 32 & RePA & 1024 & CP & QT & 32 & 5.22 \\
CelebA & RePA & 512 & CP-T & CLT & 64 & 3.83 & Imagenet 32 & RePA & 1024 & CP-T & CLT & 4 & 4.91 \\
CelebA & RePA & 512 & CP-T & QG & 2 & 5.16 & Imagenet 32 & RePA & 1024 & CP-T & CLT & 16 & 4.82 \\
CelebA & RePA & 512 & CP-T & QG & 4 & 4.79 & Imagenet 32 & RePA & 1024 & CP-T & CLT & 32 & 4.80 \\
CelebA & RePA & 512 & CP-T & QG & 8 & 4.69 & Imagenet 32 & RePA & 1024 & CP-T & QG & 2 & 5.54 \\
CelebA & RePA & 512 & CP-T & QG & 16 & 4.66 & Imagenet 32 & RePA & 1024 & CP-T & QG & 4 & 5.19 \\
CelebA & RePA & 512 & CP-T & QG & 32 & 4.66 & Imagenet 32 & RePA & 1024 & CP-T & QG & 8 & 5.10 \\
CelebA & RePA & 512 & CP-T & QG & 64 & 4.65 & Imagenet 32 & RePA & 1024 & CP-T & QG & 16 & 5.06 \\
CelebA & RePA & 512 & CP-T & QT & 2 & 5.24 & Imagenet 32 & RePA & 1024 & CP-T & QG & 32 & 5.05 \\
CelebA & RePA & 512 & CP-T & QT & 4 & 4.92 & Imagenet 32 & RePA & 1024 & CP-T & QT & 2 & 5.63 \\
CelebA & RePA & 512 & CP-T & QT & 8 & 4.83 & Imagenet 32 & RePA & 1024 & CP-T & QT & 4 & 5.32 \\
CelebA & RePA & 512 & CP-T & QT & 16 & 4.79 & Imagenet 32 & RePA & 1024 & CP-T & QT & 8 & 5.23 \\
CelebA & RePA & 512 & CP-T & QT & 32 & 4.78 & Imagenet 32 & RePA & 1024 & CP-T & QT & 16 & 5.19 \\
CelebA & RePA & 512 & CP-T & QT & 64 & 4.77 & Imagenet 32 & RePA & 1024 & CP-T & QT & 32 & 5.17 \\
CelebA & RePA & 1024 & CP-T & CLT & 4 & 3.87 & Imagenet 32 & RePA & 2048 & CP & QG & 2 & 5.56 \\
CelebA & RePA & 1024 & CP-T & CLT & 16 & 3.76 & Imagenet 32 & RePA & 2048 & CP & QG & 4 & 5.23 \\
CelebA & RePA & 1024 & CP-T & CLT & 64 & 3.74 & Imagenet 32 & RePA & 2048 & CP & QG & 8 & 5.13 \\
CelebA & RePA & 1024 & CP-T & QG & 2 & 5.10 & Imagenet 32 & RePA & 2048 & CP & QG & 16 & 5.09 \\
CelebA & RePA & 1024 & CP-T & QG & 4 & 4.70 & Imagenet 32 & RePA & 2048 & CP & QG & 32 & 5.07 \\
CelebA & RePA & 1024 & CP-T & QG & 8 & 4.58 & Imagenet 32 & RePA & 2048 & CP & QT & 2 & 5.62 \\
CelebA & RePA & 1024 & CP-T & QG & 16 & 4.55 & Imagenet 32 & RePA & 2048 & CP & QT & 4 & 5.27 \\
CelebA & RePA & 1024 & CP-T & QG & 32 & 4.53 & Imagenet 32 & RePA & 2048 & CP & QT & 8 & 5.17 \\
CelebA & RePA & 1024 & CP-T & QG & 64 & 4.52 & Imagenet 32 & RePA & 2048 & CP & QT & 16 & 5.15 \\
CelebA & RePA & 1024 & CP-T & QT & 2 & 5.18 & Imagenet 32 & RePA & 2048 & CP & QT & 32 & 5.14 \\
CelebA & RePA & 1024 & CP-T & QT & 4 & 4.87 & Imagenet 32 & RePA & 2048 & CP-T & QG & 2 & 5.51 \\
CelebA & RePA & 1024 & CP-T & QT & 8 & 4.74 & Imagenet 32 & RePA & 2048 & CP-T & QG & 4 & 5.12 \\
CelebA & RePA & 1024 & CP-T & QT & 16 & 4.71 & Imagenet 32 & RePA & 2048 & CP-T & QG & 8 & 5.01 \\
CelebA & RePA & 1024 & CP-T & QT & 32 & 4.68 & Imagenet 32 & RePA & 2048 & CP-T & QG & 16 & 4.98 \\
CelebA & RePA & 1024 & CP-T & QT & 64 & 4.67 & Imagenet 32 & RePA & 2048 & CP-T & QG & 32 & 4.97 \\
CelebA & RePA & 2048 & CP-T & QG & 2 & 5.05 & Imagenet 32 & RePA & 2048 & CP-T & QT & 2 & 5.58 \\
CelebA & RePA & 2048 & CP-T & QG & 4 & 4.62 & Imagenet 32 & RePA & 2048 & CP-T & QT & 4 & 5.25 \\
CelebA & RePA & 2048 & CP-T & QG & 8 & 4.50 & Imagenet 32 & RePA & 2048 & CP-T & QT & 8 & 5.14 \\
CelebA & RePA & 2048 & CP-T & QG & 16 & 4.47 & Imagenet 32 & RePA & 2048 & CP-T & QT & 16 & 5.11 \\
CelebA & RePA & 2048 & CP-T & QG & 32 & 4.42 & Imagenet 32 & RePA & 2048 & CP-T & QT & 32 & 5.09 \\
CelebA & RePA & 2048 & CP-T & QT & 2 & 5.14 & Imagenet 64 & IPA & 128 & CP-T & CLT & 2 & 5.17 \\
CelebA & RePA & 2048 & CP-T & QT & 4 & 4.78 & Imagenet 64 & IPA & 128 & CP-T & CLT & 4 & 4.83 \\
CelebA & RePA & 2048 & CP-T & QT & 8 & 4.67 & Imagenet 64 & IPA & 128 & CP-T & CLT & 8 & 4.75 \\
CelebA & RePA & 2048 & CP-T & QT & 16 & 4.62 & Imagenet 64 & IPA & 128 & CP-T & CLT & 16 & 4.73 \\
CelebA & RePA & 2048 & CP-T & QT & 32 & 4.61 & Imagenet 64 & IPA & 128 & CP-T & CLT & 32 & 4.72 \\
CelebA & RePA & 2048 & CP-T & QT & 64 & 4.58 & Imagenet 64 & IPA & 128 & CP-T & QG & 2 & 5.67 \\
CelebA & RePA & 4096 & CP-T & QG & 2 & 5.02 & Imagenet 64 & IPA & 128 & CP-T & QG & 4 & 5.44 \\
CelebA & RePA & 4096 & CP-T & QG & 4 & 4.57 & Imagenet 64 & IPA & 128 & CP-T & QG & 8 & 5.37 \\
CelebA & RePA & 4096 & CP-T & QG & 8 & 4.44 & Imagenet 64 & IPA & 128 & CP-T & QG & 16 & 5.36 \\
CelebA & RePA & 4096 & CP-T & QG & 16 & 4.39 & Imagenet 64 & IPA & 128 & CP-T & QG & 32 & 5.36 \\
CelebA & RePA & 4096 & CP-T & QT & 2 & 5.12 & Imagenet 64 & IPA & 128 & CP-T & QG & 64 & 5.37 \\
CelebA & RePA & 4096 & CP-T & QT & 4 & 4.71 & Imagenet 64 & IPA & 128 & CP-T & QT & 2 & 5.74 \\
CelebA & RePA & 4096 & CP-T & QT & 8 & 4.59 & Imagenet 64 & IPA & 128 & CP-T & QT & 4 & 5.51 \\
CelebA & RePA & 4096 & CP-T & QT & 16 & 4.55 & Imagenet 64 & IPA & 128 & CP-T & QT & 8 & 5.45 \\
CelebA & RePA & 4096 & CP-T & QT & 32 & 4.52 & Imagenet 64 & IPA & 128 & CP-T & QT & 16 & 5.44 \\
Imagenet 32 & IPA & 128 & CP & CLT & 2 & 5.49 & Imagenet 64 & IPA & 128 & CP-T & QT & 32 & 5.45 \\
Imagenet 32 & IPA & 128 & CP & CLT & 4 & 5.12 & Imagenet 64 & IPA & 128 & CP-T & QT & 64 & 5.45 \\
Imagenet 32 & IPA & 128 & CP & CLT & 8 & 5.09 & Imagenet 64 & IPA & 256 & CP-T & CLT & 2 & 5.10 \\
Imagenet 32 & IPA & 128 & CP & CLT & 16 & 5.07 & Imagenet 64 & IPA & 256 & CP-T & CLT & 4 & 4.74 \\
Imagenet 32 & IPA & 128 & CP & CLT & 32 & 5.06 & Imagenet 64 & IPA & 256 & CP-T & CLT & 8 & 4.68 \\
Imagenet 32 & IPA & 128 & CP & QG & 2 & 5.82 & Imagenet 64 & IPA & 256 & CP-T & CLT & 16 & 4.65 \\
Imagenet 32 & IPA & 128 & CP & QG & 4 & 5.67 & Imagenet 64 & IPA & 256 & CP-T & QG & 2 & 5.56 \\
Imagenet 32 & IPA & 128 & CP & QG & 8 & 5.65 & Imagenet 64 & IPA & 256 & CP-T & QG & 4 & 5.28 \\
Imagenet 32 & IPA & 128 & CP & QG & 16 & 5.64 & Imagenet 64 & IPA & 256 & CP-T & QG & 8 & 5.22 \\
Imagenet 32 & IPA & 128 & CP & QG & 32 & 5.65 & Imagenet 64 & IPA & 256 & CP-T & QG & 16 & 5.20 \\
Imagenet 32 & IPA & 128 & CP & QT & 2 & 5.97 & Imagenet 64 & IPA & 256 & CP-T & QG & 32 & 5.21 \\
Imagenet 32 & IPA & 128 & CP & QT & 4 & 5.75 & Imagenet 64 & IPA & 256 & CP-T & QT & 2 & 5.63 \\
Imagenet 32 & IPA & 128 & CP & QT & 8 & 5.71 & Imagenet 64 & IPA & 256 & CP-T & QT & 4 & 5.38 \\
Imagenet 32 & IPA & 128 & CP & QT & 16 & 5.69 & Imagenet 64 & IPA & 256 & CP-T & QT & 8 & 5.31 \\
Imagenet 32 & IPA & 128 & CP & QT & 32 & 5.70 & Imagenet 64 & IPA & 256 & CP-T & QT & 16 & 5.31 \\
Imagenet 32 & IPA & 128 & CP-T & CLT & 2 & 5.40 & Imagenet 64 & IPA & 256 & CP-T & QT & 32 & 5.32 \\
Imagenet 32 & IPA & 128 & CP-T & CLT & 4 & 5.07 & Imagenet 64 & IPA & 512 & CP-T & CLT & 2 & 5.05 \\
Imagenet 32 & IPA & 128 & CP-T & CLT & 8 & 5.02 & Imagenet 64 & IPA & 512 & CP-T & CLT & 4 & 4.69 \\
Imagenet 32 & IPA & 128 & CP-T & CLT & 16 & 5.01 & Imagenet 64 & IPA & 512 & CP-T & CLT & 8 & 4.62 \\
Imagenet 32 & IPA & 128 & CP-T & CLT & 32 & 4.98 & Imagenet 64 & IPA & 512 & CP-T & CLT & 16 & 4.59 \\
Imagenet 32 & IPA & 128 & CP-T & QG & 2 & 5.76 & Imagenet 64 & IPA & 512 & CP-T & CLT & 32 & 4.59 \\
Imagenet 32 & IPA & 128 & CP-T & QG & 4 & 5.55 & Imagenet 64 & IPA & 512 & CP-T & QG & 2 & 5.47 \\
Imagenet 32 & IPA & 128 & CP-T & QG & 8 & 5.51 & Imagenet 64 & IPA & 512 & CP-T & QG & 4 & 5.16 \\
Imagenet 32 & IPA & 128 & CP-T & QG & 16 & 5.49 & Imagenet 64 & IPA & 512 & CP-T & QG & 8 & 5.09 \\
Imagenet 32 & IPA & 128 & CP-T & QG & 32 & 5.51 & Imagenet 64 & IPA & 512 & CP-T & QG & 16 & 5.08 \\
Imagenet 32 & IPA & 128 & CP-T & QT & 2 & 5.86 & Imagenet 64 & IPA & 512 & CP-T & QT & 2 & 5.55 \\
Imagenet 32 & IPA & 128 & CP-T & QT & 4 & 5.63 & Imagenet 64 & IPA & 512 & CP-T & QT & 4 & 5.28 \\
Imagenet 32 & IPA & 128 & CP-T & QT & 8 & 5.59 & Imagenet 64 & IPA & 512 & CP-T & QT & 8 & 5.22 \\
Imagenet 32 & IPA & 128 & CP-T & QT & 16 & 5.59 & Imagenet 64 & IPA & 512 & CP-T & QT & 16 & 5.21 \\
Imagenet 32 & IPA & 128 & CP-T & QT & 32 & 5.60 & Imagenet 64 & IPA & 512 & CP-T & QT & 32 & 5.22 \\
Imagenet 32 & IPA & 256 & CP & CLT & 2 & 5.44 & Imagenet 64 & IPA & 1024 & CP-T & CLT & 2 & 5.03 \\
Imagenet 32 & IPA & 256 & CP & CLT & 4 & 5.05 & Imagenet 64 & IPA & 1024 & CP-T & CLT & 4 & 4.64 \\
Imagenet 32 & IPA & 256 & CP & CLT & 8 & 5.02 & Imagenet 64 & IPA & 1024 & CP-T & CLT & 8 & 4.57 \\
Imagenet 32 & IPA & 256 & CP & CLT & 16 & 5.00 & Imagenet 64 & IPA & 1024 & CP-T & QG & 2 & 5.40 \\
Imagenet 32 & IPA & 256 & CP & QG & 2 & 5.78 & Imagenet 64 & IPA & 1024 & CP-T & QG & 4 & 5.07 \\
Imagenet 32 & IPA & 256 & CP & QG & 4 & 5.58 & Imagenet 64 & IPA & 1024 & CP-T & QG & 8 & 5.00 \\
Imagenet 32 & IPA & 256 & CP & QG & 8 & 5.51 & Imagenet 64 & IPA & 1024 & CP-T & QT & 2 & 5.49 \\
Imagenet 32 & IPA & 256 & CP & QG & 16 & 5.52 & Imagenet 64 & IPA & 1024 & CP-T & QT & 4 & 5.19 \\
Imagenet 32 & IPA & 256 & CP & QG & 32 & 5.55 & Imagenet 64 & IPA & 1024 & CP-T & QT & 8 & 5.12 \\
Imagenet 32 & IPA & 256 & CP & QT & 2 & 5.86 & Imagenet 64 & IPA & 1024 & CP-T & QT & 16 & 5.11 \\
Imagenet 32 & IPA & 256 & CP & QT & 4 & 5.62 & Imagenet 64 & RePA & 512 & CP-T & CLT & 4 & 4.67 \\
Imagenet 32 & IPA & 256 & CP & QT & 8 & 5.58 & Imagenet 64 & RePA & 512 & CP-T & CLT & 16 & 4.59 \\
Imagenet 32 & IPA & 256 & CP & QT & 16 & 5.58 & Imagenet 64 & RePA & 512 & CP-T & CLT & 64 & 4.58 \\
Imagenet 32 & IPA & 256 & CP & QT & 32 & 5.59 & Imagenet 64 & RePA & 512 & CP-T & QG & 2 & 5.46 \\
Imagenet 32 & IPA & 256 & CP-T & CLT & 2 & 5.34 & Imagenet 64 & RePA & 512 & CP-T & QG & 4 & 5.11 \\
Imagenet 32 & IPA & 256 & CP-T & CLT & 4 & 4.99 & Imagenet 64 & RePA & 512 & CP-T & QG & 8 & 5.02 \\
Imagenet 32 & IPA & 256 & CP-T & CLT & 8 & 4.95 & Imagenet 64 & RePA & 512 & CP-T & QG & 16 & 4.99 \\
Imagenet 32 & IPA & 256 & CP-T & CLT & 16 & 4.94 & Imagenet 64 & RePA & 512 & CP-T & QG & 32 & 4.98 \\
Imagenet 32 & IPA & 256 & CP-T & QG & 2 & 5.68 & Imagenet 64 & RePA & 512 & CP-T & QG & 64 & 4.99 \\
Imagenet 32 & IPA & 256 & CP-T & QG & 4 & 5.42 & Imagenet 64 & RePA & 512 & CP-T & QT & 2 & 5.56 \\
Imagenet 32 & IPA & 256 & CP-T & QG & 8 & 5.37 & Imagenet 64 & RePA & 512 & CP-T & QT & 4 & 5.24 \\
Imagenet 32 & IPA & 256 & CP-T & QG & 16 & 5.38 & Imagenet 64 & RePA & 512 & CP-T & QT & 8 & 5.14 \\
Imagenet 32 & IPA & 256 & CP-T & QG & 32 & 5.41 & Imagenet 64 & RePA & 512 & CP-T & QT & 16 & 5.11 \\
Imagenet 32 & IPA & 256 & CP-T & QT & 2 & 5.75 & Imagenet 64 & RePA & 512 & CP-T & QT & 32 & 5.09 \\
Imagenet 32 & IPA & 256 & CP-T & QT & 4 & 5.53 & Imagenet 64 & RePA & 512 & CP-T & QT & 64 & 5.10 \\
Imagenet 32 & IPA & 256 & CP-T & QT & 8 & 5.48 & Imagenet 64 & RePA & 1024 & CP-T & CLT & 4 & 4.62 \\
Imagenet 32 & IPA & 256 & CP-T & QT & 16 & 5.49 & Imagenet 64 & RePA & 1024 & CP-T & CLT & 16 & 4.53 \\
Imagenet 32 & IPA & 256 & CP-T & QT & 32 & 5.51 & Imagenet 64 & RePA & 1024 & CP-T & CLT & 64 & 4.52 \\
Imagenet 32 & IPA & 512 & CP & CLT & 2 & 5.38 & Imagenet 64 & RePA & 1024 & CP-T & QG & 2 & 5.39 \\
Imagenet 32 & IPA & 512 & CP & CLT & 4 & 4.99 & Imagenet 64 & RePA & 1024 & CP-T & QG & 4 & 4.99 \\
Imagenet 32 & IPA & 512 & CP & CLT & 8 & 4.96 & Imagenet 64 & RePA & 1024 & CP-T & QG & 8 & 4.90 \\
Imagenet 32 & IPA & 512 & CP & CLT & 16 & 4.95 & Imagenet 64 & RePA & 1024 & CP-T & QG & 16 & 4.86 \\
Imagenet 32 & IPA & 512 & CP & CLT & 32 & 4.97 & Imagenet 64 & RePA & 1024 & CP-T & QG & 32 & 4.86 \\
Imagenet 32 & IPA & 512 & CP & QG & 2 & 5.71 & Imagenet 64 & RePA & 1024 & CP-T & QT & 2 & 5.49 \\
Imagenet 32 & IPA & 512 & CP & QG & 4 & 5.47 & Imagenet 64 & RePA & 1024 & CP-T & QT & 4 & 5.13 \\
Imagenet 32 & IPA & 512 & CP & QG & 8 & 5.39 & Imagenet 64 & RePA & 1024 & CP-T & QT & 8 & 5.03 \\
Imagenet 32 & IPA & 512 & CP & QG & 16 & 5.42 & Imagenet 64 & RePA & 1024 & CP-T & QT & 16 & 4.99 \\
Imagenet 32 & IPA & 512 & CP & QG & 32 & 5.47 & Imagenet 64 & RePA & 1024 & CP-T & QT & 32 & 4.98 \\
Imagenet 32 & IPA & 512 & CP & QT & 2 & 5.77 & Imagenet 64 & RePA & 1024 & CP-T & QT & 64 & 4.97 \\
Imagenet 32 & IPA & 512 & CP & QT & 4 & 5.51 & Imagenet 64 & RePA & 2048 & CP-T & QG & 2 & 5.35 \\
Imagenet 32 & IPA & 512 & CP & QT & 8 & 5.46 & Imagenet 64 & RePA & 2048 & CP-T & QG & 4 & 4.92 \\
Imagenet 32 & IPA & 512 & CP & QT & 16 & 5.47 & Imagenet 64 & RePA & 2048 & CP-T & QG & 8 & 4.81 \\
Imagenet 32 & IPA & 512 & CP & QT & 32 & 5.51 & Imagenet 64 & RePA & 2048 & CP-T & QG & 16 & 4.77 \\
Imagenet 32 & IPA & 512 & CP-T & CLT & 2 & 5.30 & Imagenet 64 & RePA & 2048 & CP-T & QT & 2 & 5.43 \\
Imagenet 32 & IPA & 512 & CP-T & CLT & 4 & 4.94 & Imagenet 64 & RePA & 2048 & CP-T & QT & 4 & 5.06 \\
Imagenet 32 & IPA & 512 & CP-T & CLT & 8 & 4.90 & Imagenet 64 & RePA & 2048 & CP-T & QT & 8 & 4.95 \\
Imagenet 32 & IPA & 512 & CP-T & CLT & 16 & 4.89 & Imagenet 64 & RePA & 2048 & CP-T & QT & 16 & 4.89 \\
Imagenet 32 & IPA & 512 & CP-T & QG & 2 & 5.61 & Imagenet 64 & RePA & 2048 & CP-T & QT & 32 & 4.89 \\
Imagenet 32 & IPA & 512 & CP-T & QG & 4 & 5.33 & Imagenet 64 & RePA & 4096 & CP-T & QG & 2 & 5.31 \\
Imagenet 32 & IPA & 512 & CP-T & QG & 8 & 5.28 & Imagenet 64 & RePA & 4096 & CP-T & QG & 4 & 4.86 \\
Imagenet 32 & IPA & 512 & CP-T & QG & 16 & 5.29 & Imagenet 64 & RePA & 4096 & CP-T & QG & 8 & 4.72 \\
Imagenet 32 & IPA & 512 & CP-T & QG & 32 & 5.35 & Imagenet 64 & RePA & 4096 & CP-T & QT & 2 & 5.39 \\
Imagenet 32 & IPA & 512 & CP-T & QT & 2 & 5.69 & Imagenet 64 & RePA & 4096 & CP-T & QT & 4 & 4.98 \\
Imagenet 32 & IPA & 512 & CP-T & QT & 4 & 5.45 & Imagenet 64 & RePA & 4096 & CP-T & QT & 8 & 4.86 \\
Imagenet 32 & IPA & 512 & CP-T & QT & 8 & 5.39 & Imagenet 64 & RePA & 4096 & CP-T & QT & 16 & 4.82 \\
\end{longtable}

}

\begin{figure}
    \centering
    \includegraphics[width=0.32\linewidth]{media/bpd_ds_rg_ipa_repa/imagenet-32__qt.pdf}
    \hfill
    \includegraphics[width=0.32\linewidth]{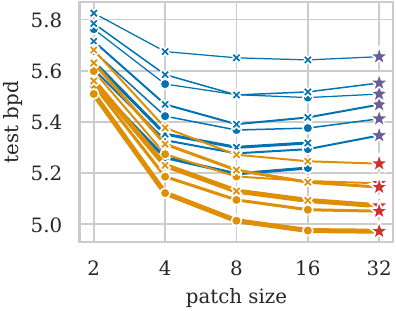}
    \hfill
    \includegraphics[width=0.32\linewidth]{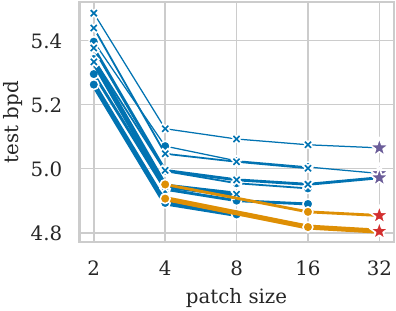}

    \includegraphics[width=0.32\linewidth]{media/bpd_ds_rg_ipa_repa/imagenet-64__qt.pdf}
    \hfill
    \includegraphics[width=0.32\linewidth]{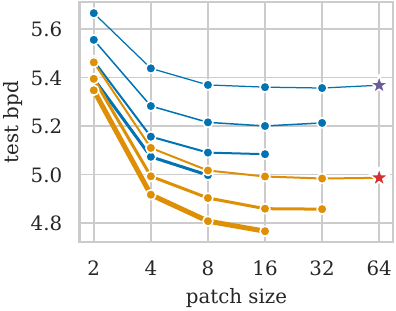}
    \hfill
    \includegraphics[width=0.32\linewidth]{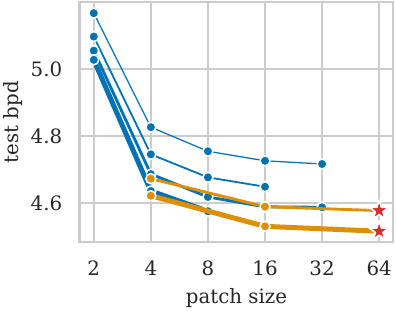}

    \includegraphics[width=0.32\linewidth]{media/bpd_ds_rg_ipa_repa/celeba__qt.pdf}
    \hfill
    \includegraphics[width=0.32\linewidth]{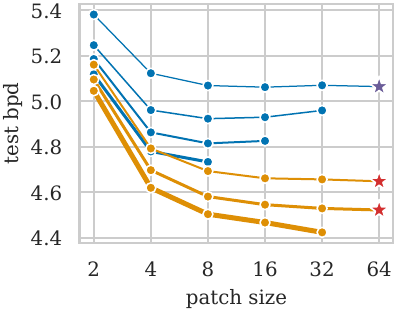}
    \hfill
    \includegraphics[width=0.32\linewidth]{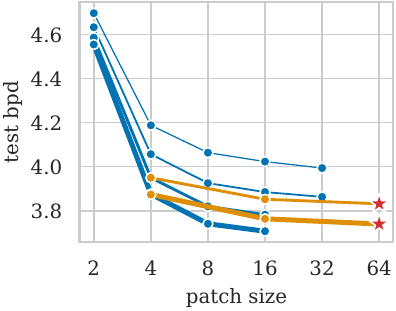}
    \caption{\textbf{\repa improves learnability for density estimation.} We compare the bpd with respect to patch size for QT (left), QG (center), CLT (right) on ImageNet32 (top), ImageNet64 (middle) and CelebA (bottom)
    }
    \label{fig:bpd_vs_patch}
\end{figure}

\subsection{Full results: efficiency Pareto front}

\begin{figure}
    \centering
    \includegraphics[width=0.32\linewidth]{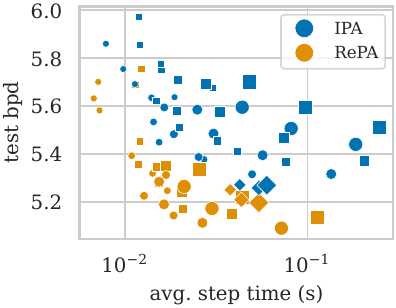}
    \hfill
    \includegraphics[width=0.32\linewidth]{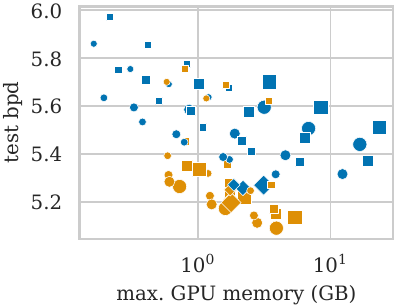}
    \hfill
    \includegraphics[width=0.32\linewidth]{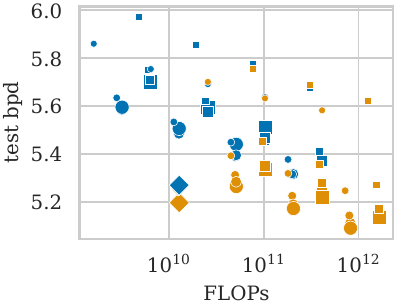}

    \includegraphics[width=0.32\linewidth]{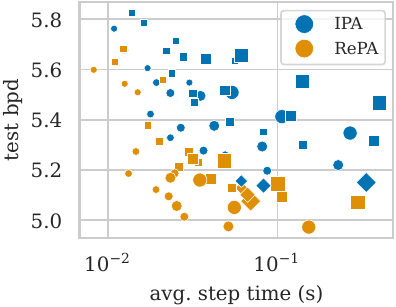}
    \hfill
    \includegraphics[width=0.32\linewidth]{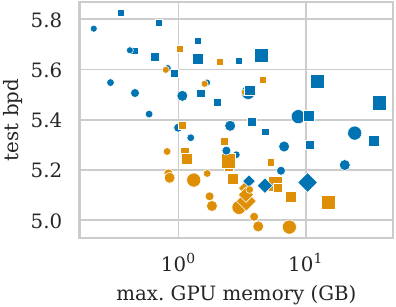}
    \hfill
    \includegraphics[width=0.32\linewidth]{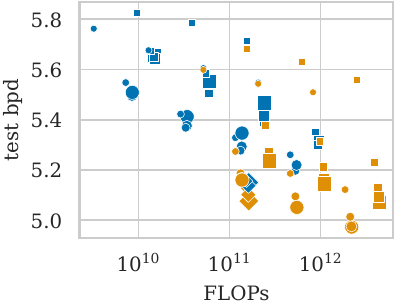}

    \includegraphics[width=0.32\linewidth]{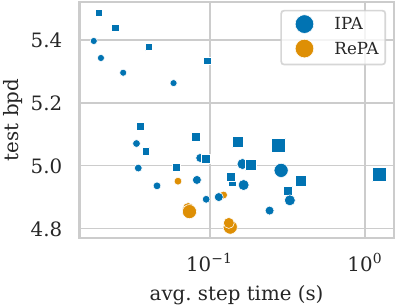}
    \hfill
    \includegraphics[width=0.32\linewidth]{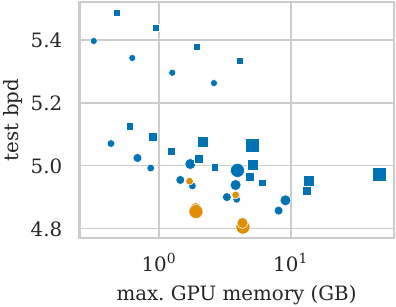}
    \hfill
    \includegraphics[width=0.32\linewidth]{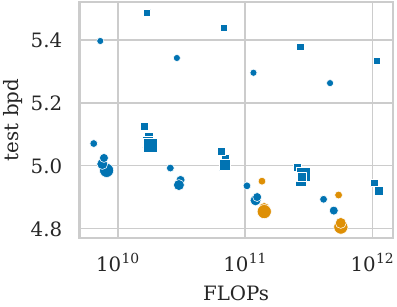}
    \caption{ \textbf{\repa yields a better Pareto front on ImageNet32.}  We compare the bpd with respect to efficiency for QT (top), QG (middle), CLT (bottom) on ImageNet32. We do not distinguish different values of $K$ }
    \label{app:fig:pareto_imagenet32}
\end{figure}

\begin{figure}
    \centering
    \includegraphics[width=0.32\linewidth]{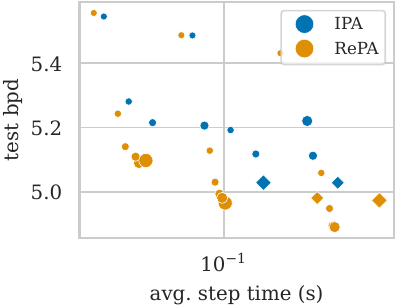}
    \hfill
    \includegraphics[width=0.32\linewidth]{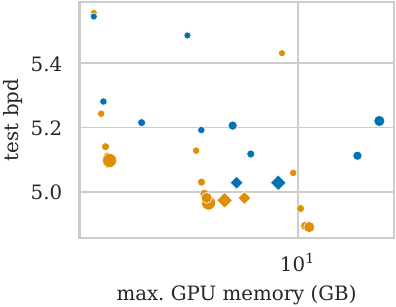}
    \hfill
    \includegraphics[width=0.32\linewidth]{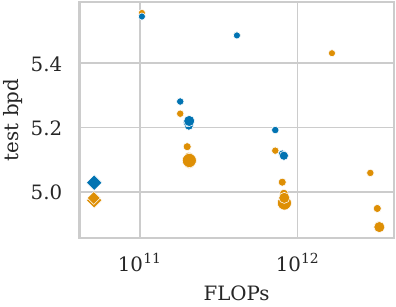}

    \includegraphics[width=0.32\linewidth]{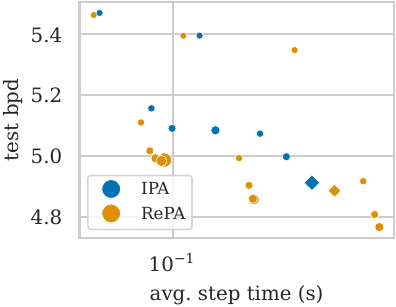}
    \hfill
    \includegraphics[width=0.32\linewidth]{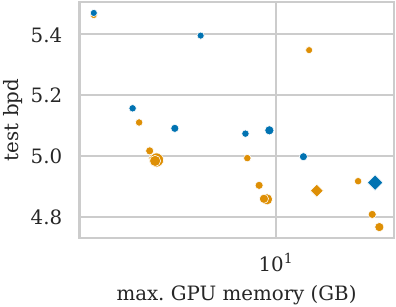}
    \hfill
    \includegraphics[width=0.32\linewidth]{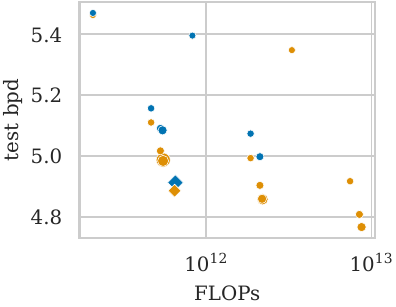}

    \includegraphics[width=0.32\linewidth]{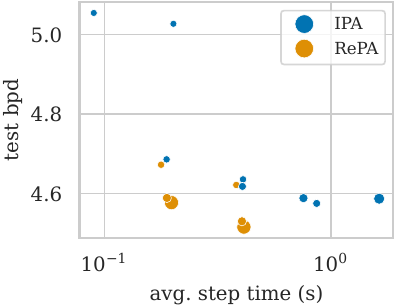}
    \hfill
    \includegraphics[width=0.32\linewidth]{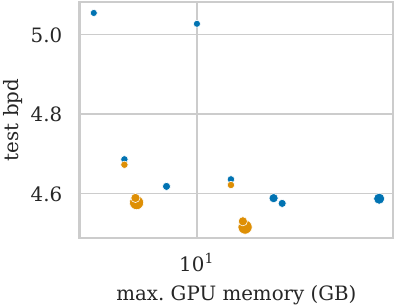}
    \hfill
    \includegraphics[width=0.32\linewidth]{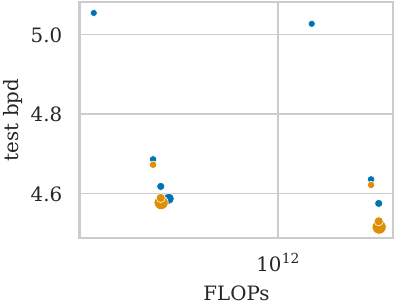}
    \caption{\textbf{\repa yields a better Pareto front on ImageNet64.}  We compare the bpd with respect to efficiency for QT (top), QG (middle), CLT (bottom) on ImageNet64. We do not distinguish different values of $K$ }
    \label{app:fig:pareto_imagenet64}
\end{figure}

\begin{figure}
    \centering
    \includegraphics[width=0.32\linewidth]{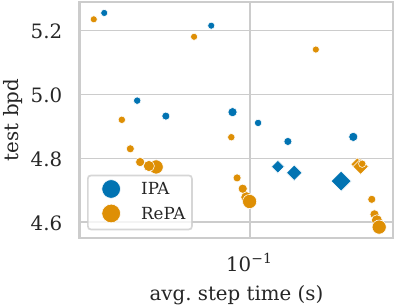}
    \hfill
    \includegraphics[width=0.32\linewidth]{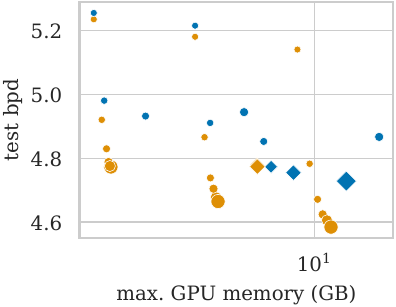}
    \hfill
    \includegraphics[width=0.32\linewidth]{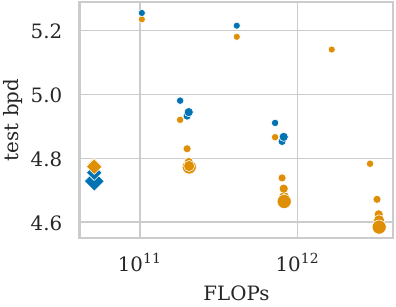}

    \includegraphics[width=0.32\linewidth]{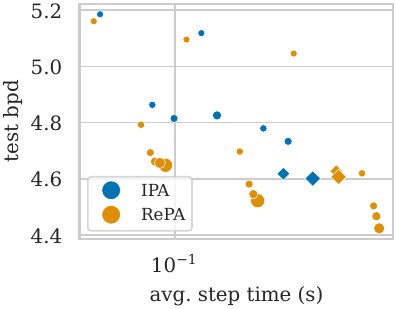}
    \hfill
    \includegraphics[width=0.32\linewidth]{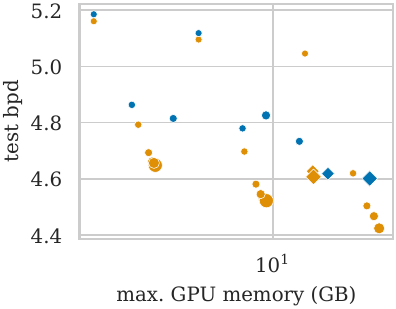}
    \hfill
    \includegraphics[width=0.32\linewidth]{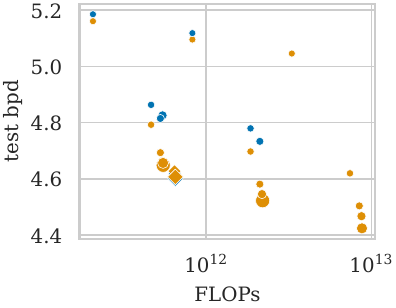}

    \includegraphics[width=0.32\linewidth]{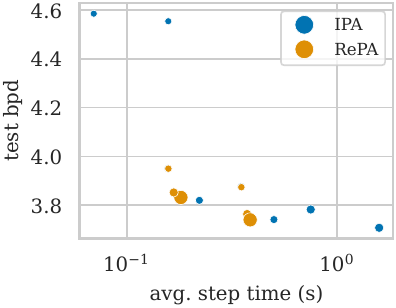}
    \hfill
    \includegraphics[width=0.32\linewidth]{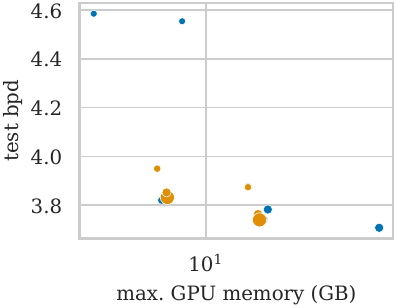}
    \hfill
    \includegraphics[width=0.32\linewidth]{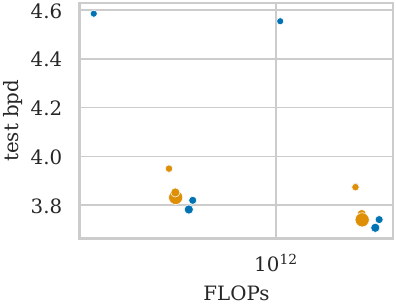}
    \caption{\textbf{\repa yields a better Pareto front on CelebA.} We compare the bpd with respect to efficiency for QT (top), QG (middle), CLT (bottom) on CelebA. We do not distinguish different values of $K$ }
    \label{app:fig:pareto_celeba}
\end{figure}

We report the raw results of our experiments, without filtering to keep the results with best bpd as we did in \cref{fig:fig-1} and \cref{app:fig:bpd_vs_efficiency}, for ImageNet32 in \cref{app:fig:pareto_imagenet32}, ImageNet64 in \cref{app:fig:pareto_imagenet64} and CelebA in \cref{app:fig:pareto_celeba}.

\subsection{\repa and Chow-Liu trees}
\label{app:repa-clt}

CLT RGs are not patch-aligned (\cref{app:rg}), i.e., they cannot be partitioned into a hierarchy where variables exactly correspond to image patches.
To use \repa with them, we use an iterative, bottom-up training strategy inspired by \citet{anji2022scaling-pcs-lvd}. 
Specifically, we alternate two stages: first, learning a shared CLT from the data (or the circuit activations $\Out^{(t)}$ at level $t$), and second, using this RG to overparameterize the PC and train it via MLE --- see \cref{app:alg:trainrepaclt}.
We then move to level $t+1$, repeating this process until we reach level $T$.
While training such a \repa PC by MLE, because each patch-level PC outputs a single probability, we must temporarily assume patch-level independence.
That is, we compute:
\begin{align}
    \indepatch_{D^{T-t}}(\repa(c_{1:t}, \X)),
    \label{eq:ipa-repa}
\end{align}
where $D^{T-t}$ is the size of the patches assumed to be independent.
Note that this local independence assumption changes at each level of the hierarchy, and is discarded when building the subsequent level.
In our experiments, we found it crucial to properly initialize the $K$ sum units introduced in the \repa patch circuits when moving to the next level, ensuring they correctly encode the distribution learned by \cref{eq:ipa-repa} (see \cref{app:subsec:multi-stage-train}). 

\begin{figure}
    \centering
    \includegraphics[width=0.32\linewidth]{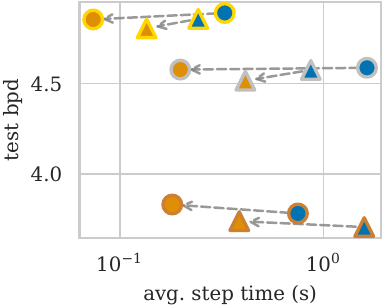}
    \includegraphics[width=0.32\linewidth]{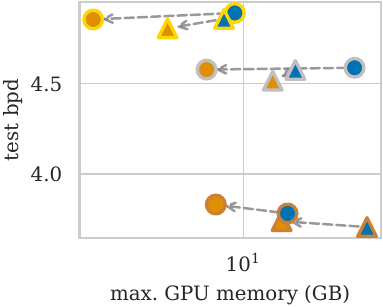}
    \includegraphics[width=0.32\linewidth]{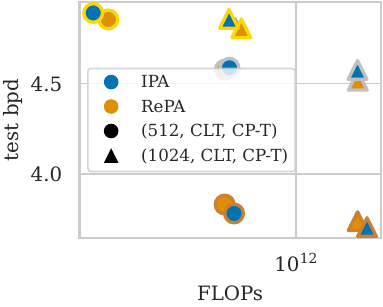}
    \caption{\textbf{Our \repa construction with CLTs reduces training time and memory consumption} for the ImageNet32 (gold border dots), ImageNet64 (silver) and CelebA (bronze) datasets.}
    \label{fig:repa_clt}
\end{figure}

Following this method, we train two \repa models for each dataset, using either $K\!=\!512$ or $K\!=\!1024$. 
Each model comprises $T\!=\!3$ PCs operating on $4\times 4$ patches, as we found $2\times 2$ patches to be too small to learn a meaningful CLT.
Because this setup results in a final \repa PC operating on a $64\times64\times3$ image---which does not match ImageNet32---we use $D=2$ for the final circuit $c_3$.
\cref{app:tab:hyper_grid_repa_clt} provides further details. 
\cref{fig:repa_clt} shows that our \repa CLT models are systematically lighter and faster than their \indepatch counterparts, while maintaining similar FLOPs. 
On the two ImageNet datasets, \repa also achieves better bpd than \indepatch, yielding the best density estimation across all our experiments. 
While these models are comparable in terms of FLOPs and quality, we emphasize that \repa CLTs are not strictly equivalent to the \indepatch CLTs, unlike \repa PCs using QT and QG RGs.  
Ultimately, this experiment shows that \repa is flexible enough to work with non-patch-aligned RGs.

\subsection{\repa and Monarch matrices}
\label{app:repa-monarch}
Monarch layers \citep{zhang_scaling_2025} are CP-T layers using special low-rank Monarch matrices \cite{dao_monarch_2022} as the weight matrices $\W$.
As such, Monarch layers constitute a parameter sharing approach that is orthogonal to our \repa. We refer the reader to \cref{app:monarch} for details about Monarch matrices. 

\begin{figure*}
    \centering\hfill
    \includegraphics[width=0.3\linewidth]{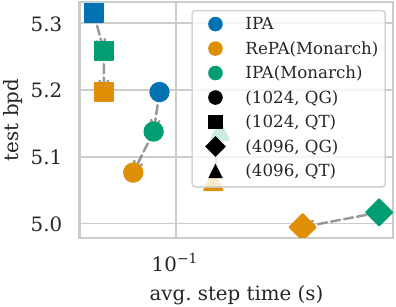}\hfill
    \includegraphics[width=0.3\linewidth]{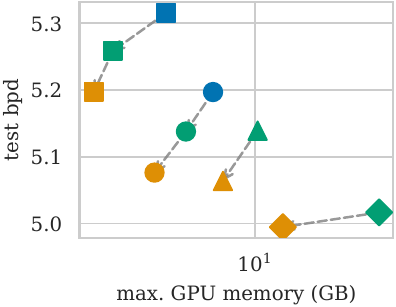}\hfill
    \includegraphics[width=0.3\linewidth]{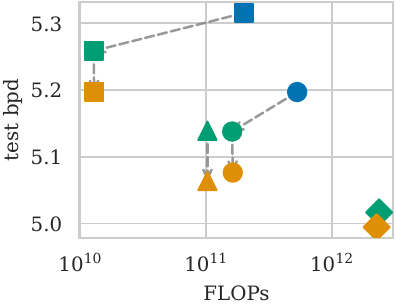}\hfill
    \caption{\textbf{Combining \repa with Monarch layers lets PCs scale even further} as shown for QT and QG RGs and $K=1024,4096$ on ImageNet32. 
    }
    \label{fig:mona_repa}
\end{figure*}

We compare architectures with $K=1024$ using \indepatch, \textsc{IPA}(Monarch) and \textsc{RePA}(Monarch) to evaluate the benefits and contributions of the two parameter sharing techniques. 
We also train additional configurations for Monarch with $K=4096$ to confirm these observations for larger PCs and to exploit the additional efficiency gains provided by Monarch layers.   
\cref{fig:mona_repa} confirms prior findings that Monarch layers can improve estimation quality while reducing FLOPs. 
We highlight that \textsc{RePA}(Monarch) systematically improves performance, demonstrating that these two parameter sharing techniques act complementarily to alleviate the learnability issue. 
Finally, we note that our implementation of Monarch layers does not use custom CUDA kernels \citep{liu2024scaling}; incorporating them could improve efficiency even further.
It is important to mention that our implementation of Monarch is not highly optimized. Therefore, we cannot strictly compare \repa and \indepatch(Monarch).

\subsection{Full results: parameter sharing on \textsc{TabPC}}

We report the full results for all 7 datasets on bpd (\cref{app:fig:full_tabpc_bpd}) and C2ST (\cref{app:fig:full_tabpc_c2st}).

\def\plotwidth{0.24}
\begin{figure}
    \centering
    
\raisebox{40pt}{\rotatebox{90}{\textsc{adult}}}\hspace{8pt}
    \includegraphics[width=\plotwidth\linewidth]{media/tabpc/xgboost_mem_adult.pdf}
    \hfill
    \includegraphics[width=\plotwidth\linewidth]{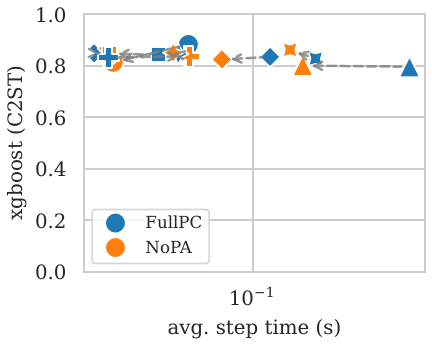}
    \hfill
    \includegraphics[width=\plotwidth\linewidth]{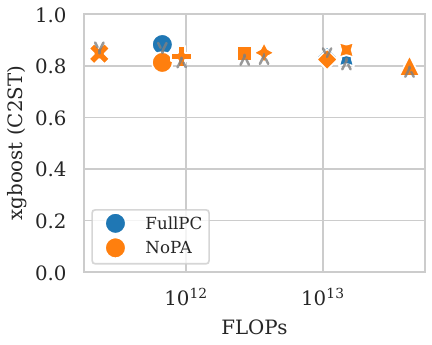}

\raisebox{40pt}{\rotatebox{90}{\textsc{beijing}}}\hspace{8pt}
    \includegraphics[width=\plotwidth\linewidth]{media/tabpc/xgboost_mem_beijing.pdf}
    \hfill
    \includegraphics[width=\plotwidth\linewidth]{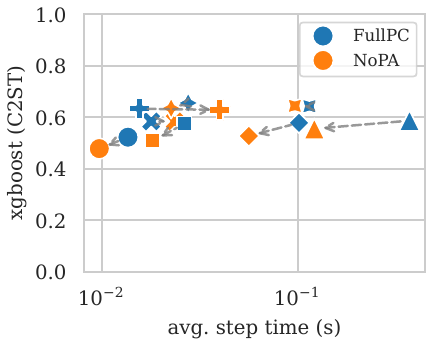}
    \hfill
    \includegraphics[width=\plotwidth\linewidth]{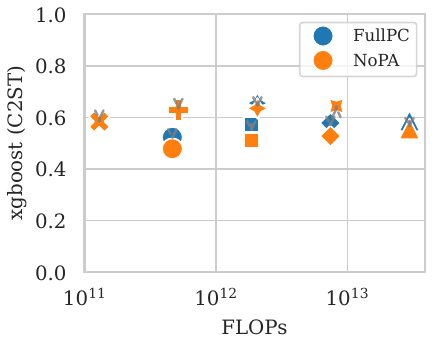}

\raisebox{40pt}{\rotatebox{90}{\textsc{default}}}\hspace{8pt}
    \includegraphics[width=\plotwidth\linewidth]{media/tabpc/xgboost_mem_default.pdf}
    \hfill
    \includegraphics[width=\plotwidth\linewidth]{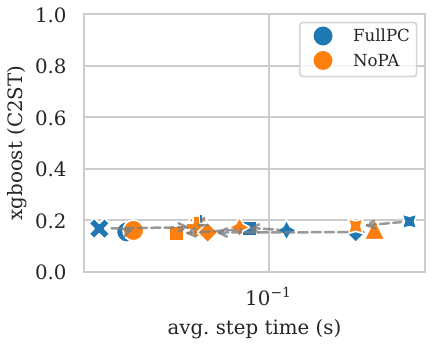}
    \hfill
    \includegraphics[width=\plotwidth\linewidth]{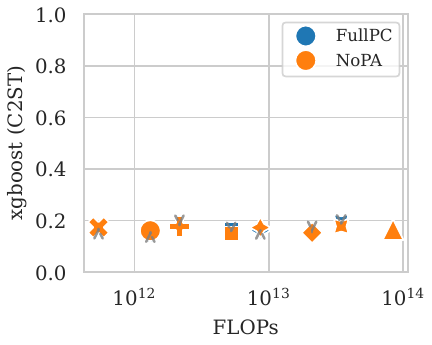}

\raisebox{40pt}{\rotatebox{90}{\textsc{diabetes}}}\hspace{8pt}
    \includegraphics[width=\plotwidth\linewidth]{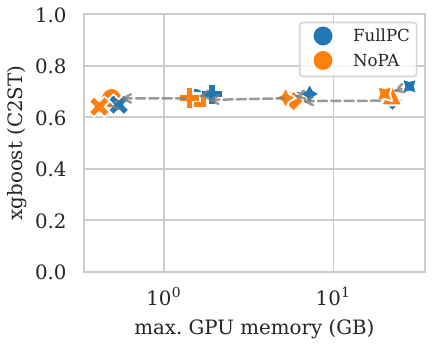}
    \hfill
    \includegraphics[width=\plotwidth\linewidth]{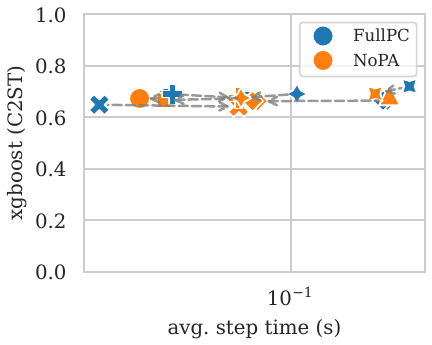}
    \hfill
    \includegraphics[width=\plotwidth\linewidth]{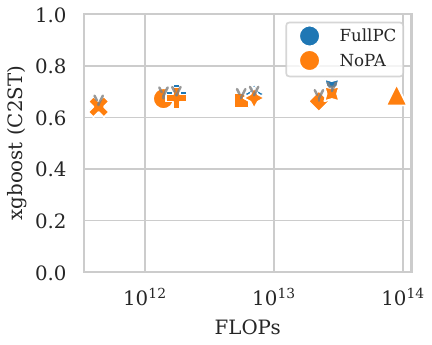}

\raisebox{40pt}{\rotatebox{90}{\textsc{magic}}}\hspace{8pt}
    \includegraphics[width=\plotwidth\linewidth]{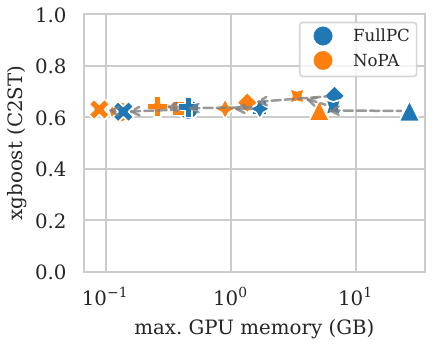}
    \hfill
    \includegraphics[width=\plotwidth\linewidth]{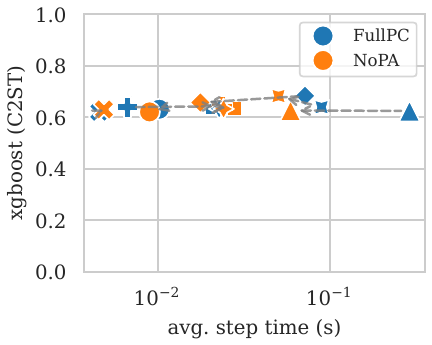}
    \hfill
    \includegraphics[width=\plotwidth\linewidth]{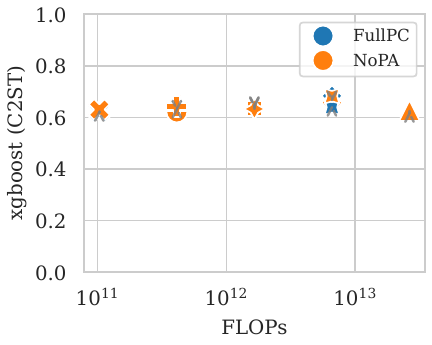}

\raisebox{40pt}{\rotatebox{90}{\textsc{news}}}\hspace{8pt}
    \includegraphics[width=\plotwidth\linewidth]{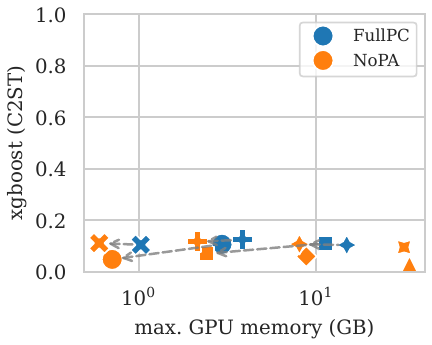}
    \hfill
    \includegraphics[width=\plotwidth\linewidth]{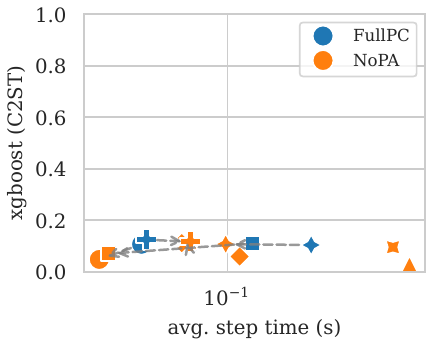}
    \hfill
    \includegraphics[width=\plotwidth\linewidth]{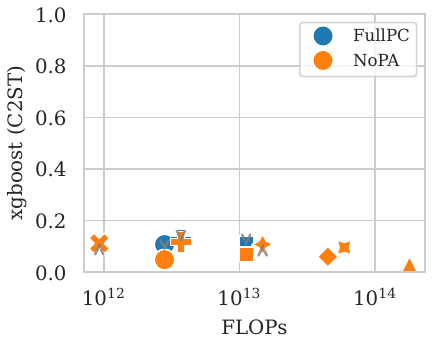}

\raisebox{40pt}{\rotatebox{90}{\textsc{shoppers}}}\hspace{8pt}
    \includegraphics[width=\plotwidth\linewidth]{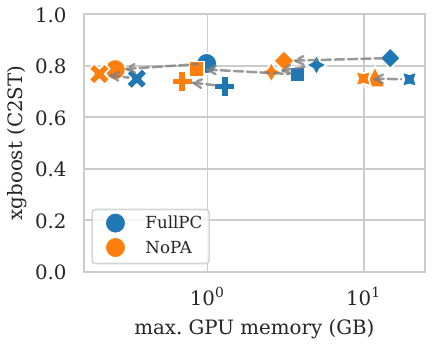}
    \hfill
    \includegraphics[width=\plotwidth\linewidth]{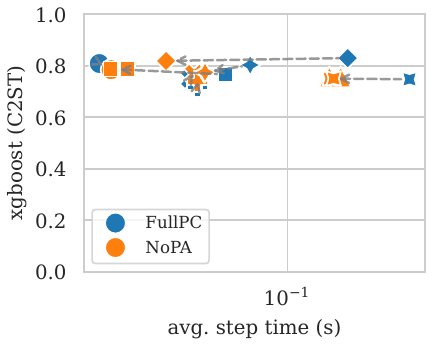}
    \hfill
    \includegraphics[width=\plotwidth\linewidth]{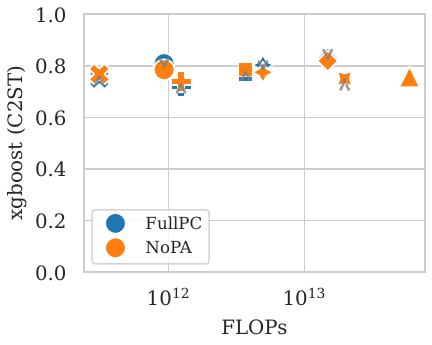}
    
    \caption{\textbf{\nopa yields more efficient models with little decrease in C2ST with XGBoost.} Each row is a dataset in this order: Adult, Beijing, Default, Diabetes, Magic, News, Shoppers.}
    \label{app:fig:full_tabpc_c2st}
\end{figure}

\begin{figure}
    \centering
    \raisebox{40pt}{\rotatebox{90}{\textsc{adult}}}\hspace{8pt}\includegraphics[width=\plotwidth\linewidth]{media/tabpc/bpd_mem_adult.pdf}
    \hfill
    \includegraphics[width=\plotwidth\linewidth]{media/tabpc/bpd_time_adult.pdf}
    \hfill
    \includegraphics[width=\plotwidth\linewidth]{media/tabpc/bpd_flops_adult.pdf}

\raisebox{40pt}{\rotatebox{90}{\textsc{beijing}}}\hspace{8pt}
    \includegraphics[width=\plotwidth\linewidth]{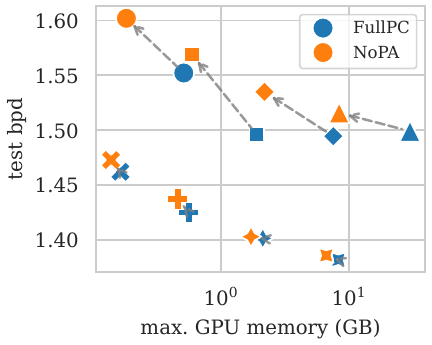}
    \hfill
    \includegraphics[width=\plotwidth\linewidth]{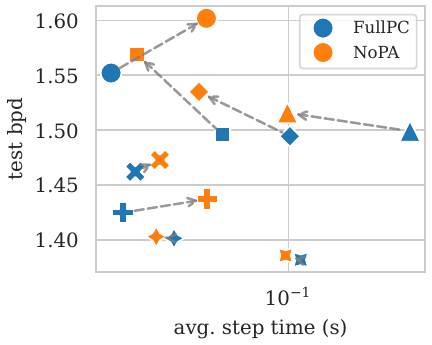}
    \hfill
    \includegraphics[width=\plotwidth\linewidth]{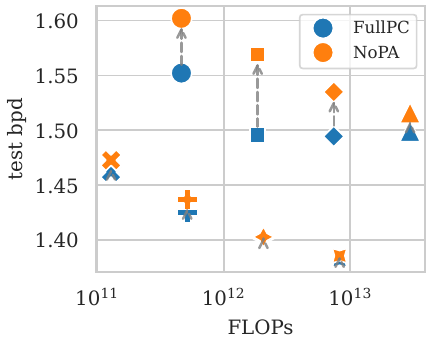}

\raisebox{40pt}{\rotatebox{90}{\textsc{default}}}\hspace{8pt}
    \includegraphics[width=\plotwidth\linewidth]{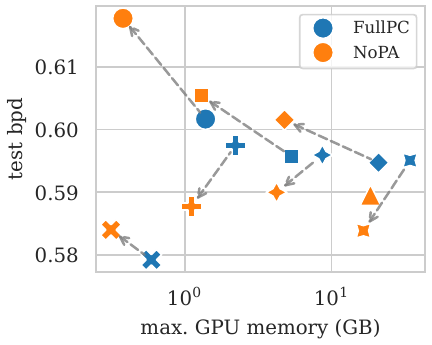}
    \hfill
    \includegraphics[width=\plotwidth\linewidth]{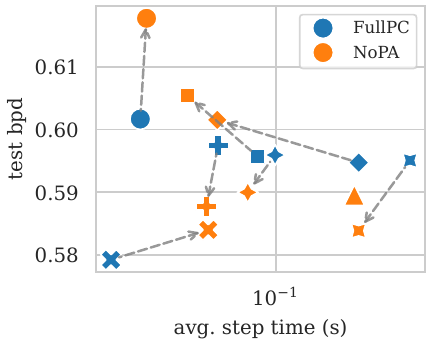}
    \hfill
    \includegraphics[width=\plotwidth\linewidth]{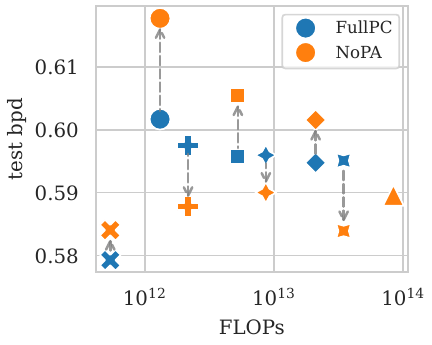}

\raisebox{40pt}{\rotatebox{90}{\textsc{diabetes}}}\hspace{8pt}
    \includegraphics[width=\plotwidth\linewidth]{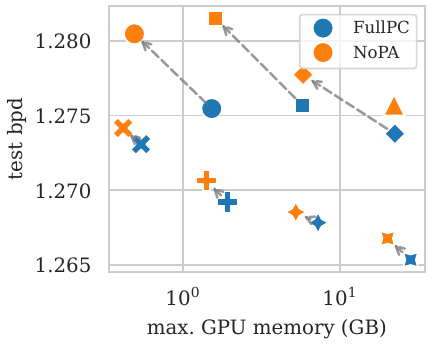}
    \hfill
    \includegraphics[width=\plotwidth\linewidth]{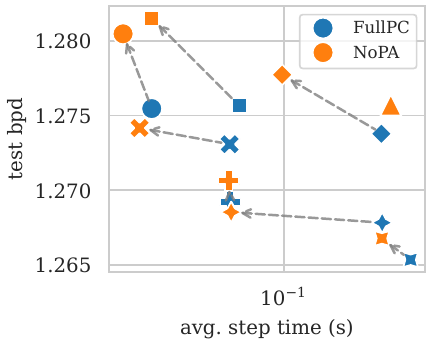}
    \hfill
    \includegraphics[width=\plotwidth\linewidth]{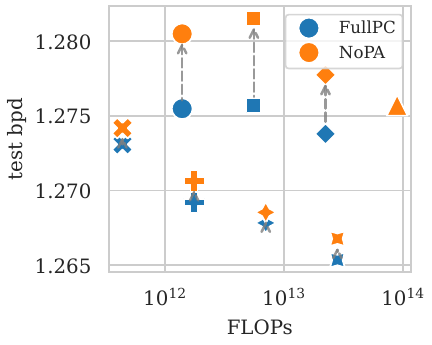}

\raisebox{40pt}{\rotatebox{90}{\textsc{magic}}}\hspace{8pt}
    \includegraphics[width=\plotwidth\linewidth]{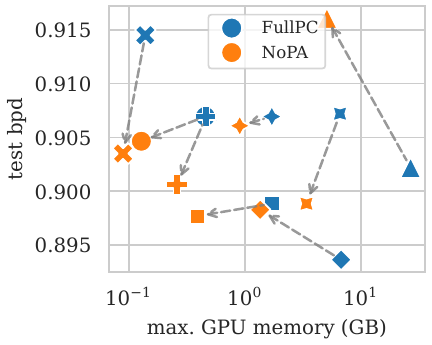}
    \hfill
    \includegraphics[width=\plotwidth\linewidth]{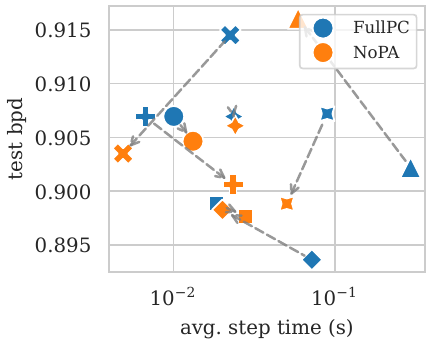}
    \hfill
    \includegraphics[width=\plotwidth\linewidth]{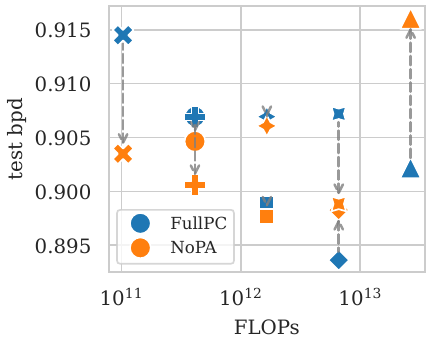}

\raisebox{40pt}{\rotatebox{90}{\textsc{news}}}\hspace{8pt}
    \includegraphics[width=\plotwidth\linewidth]{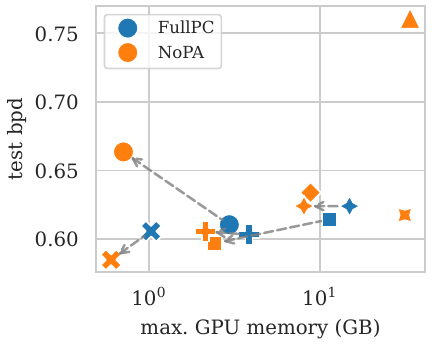}
    \hfill
    \includegraphics[width=\plotwidth\linewidth]{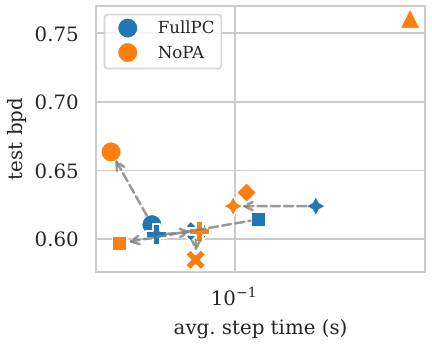}
    \hfill
    \includegraphics[width=\plotwidth\linewidth]{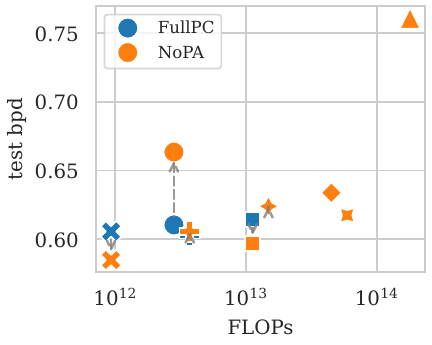}

\raisebox{40pt}{\rotatebox{90}{\textsc{shoppers}}}\hspace{8pt}
    \includegraphics[width=\plotwidth\linewidth]{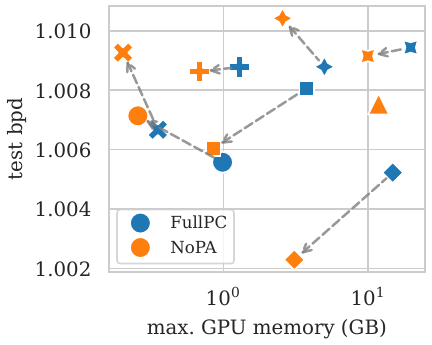}
    \hfill
    \includegraphics[width=\plotwidth\linewidth]{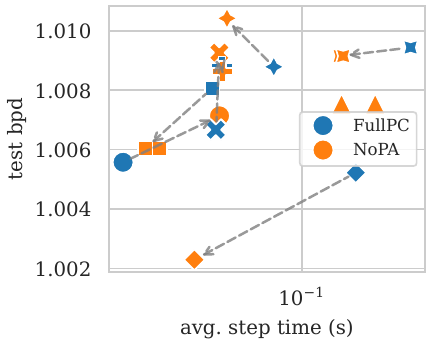}
    \hfill
    \includegraphics[width=\plotwidth\linewidth]{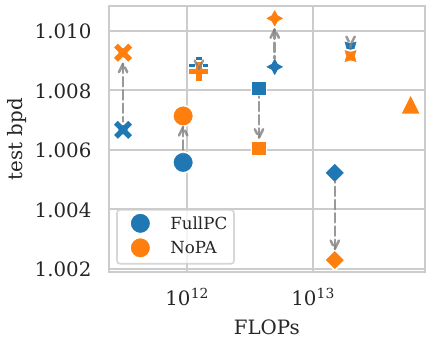}
    
    \caption{\textbf{\nopa yields more efficient models with little increase in bpd.} Each row is a dataset in this order: Adult, Beijing, Default, Diabetes, Magic, News, Shoppers.}
    \label{app:fig:full_tabpc_bpd}
\end{figure}

\subsection{Full results: samples}

\clearpage
\section*{NeurIPS Paper Checklist}

\begin{enumerate}

\item {\bf Claims}
    \item[] Question: Do the main claims made in the abstract and introduction accurately reflect the paper's contributions and scope?
    \item[] Answer: \answerYes{} %
    \item[] Justification: The abstract claims that we show a learnability issue which is explained in \cref{sec:indepatch} and supported by the results in \cref{fig:bpd_vs_patch2}. It also claims that we introduce a new method with efficiency and quality improvement, detailed in \cref{sec:repa} and supported by the results in \cref{subsec:efficiency_repa}.

\item {\bf Limitations}
    \item[] Question: Does the paper discuss the limitations of the work performed by the authors?
    \item[] Answer:  \answerYes{}
    \item[] Justification: We mention all significant limitations in the conclusion of the paper.

\item {\bf Theory assumptions and proofs}
    \item[] Question: For each theoretical result, does the paper provide the full set of assumptions and a complete (and correct) proof?
    \item[] Answer: \answerNA{} %
    \item[] Justification: No theoretical results are presented.

    \item {\bf Experimental result reproducibility}
    \item[] Question: Does the paper fully disclose all the information needed to reproduce the main experimental results of the paper to the extent that it affects the main claims and/or conclusions of the paper (regardless of whether the code and data are provided or not)?
    \item[] Answer: \answerYes{} %
    \item[] Justification: We provide full formulation of \repa in \cref{sec:repa} as well as the training procedure to train \repa(CLT) in \cref{app:repa-clt}. Furthermore, these are supported by algorithms in \cref{app:patch_algos}. 

\item {\bf Open access to data and code}
    \item[] Question: Does the paper provide open access to the data and code, with sufficient instructions to faithfully reproduce the main experimental results, as described in supplemental material?
    \item[] Answer: \answerYes{} %
    \item[] Justification: This paper only uses publicly available datasets and provides the necessary transforms in \cref{app:exp-details}. The code will be released on GitHub upon acceptance with instructions and scripts to create the environment, process the datasets and run the experiments.

\item {\bf Experimental setting/details}
    \item[] Question: Does the paper specify all the training and test details (e.g., data splits, hyperparameters, how they were chosen, type of optimizer) necessary to understand the results?
    \item[] Answer: \answerYes{}
    \item[] Justification: We also disclose the full set of hyperparameters for each model in \cref{app:hyperparams} as well as the general training setting in \cref{sec:experiments}. We specify the data split in \cref{app:dataset}.

\item {\bf Experiment statistical significance}
    \item[] Question: Does the paper report error bars suitably and correctly defined or other appropriate information about the statistical significance of the experiments?
    \item[] Answer: \answerYes{}
    \item[] Justification: 
    We run 6 seeds of our best models to demonstrate the low variance in the results. Furthermore, rather than running several replicas of a single experiment configuration, here we are interested in mapping a very large (444 distinct configurations) experimental configuration space, and use this population to extract a statistical trend. 

\item {\bf Experiments compute resources}
    \item[] Question: For each experiment, does the paper provide sufficient information on the computer resources (type of compute workers, memory, time of execution) needed to reproduce the experiments?
    \item[] Answer: \answerYes{}
    \item[] Justification: The hardware used to run the model is disclosed in \cref{app:hw}.
    
\item {\bf Code of ethics}
    \item[] Question: Does the research conducted in the paper conform, in every respect, with the NeurIPS Code of Ethics \url{https://neurips.cc/public/EthicsGuidelines}?
    \item[] Answer: \answerYes{}
    \item[] Justification: Our paper does not include any research on live subjects or new data and we do not anticipate harmful consequences from this work.

\item {\bf Broader impacts}
    \item[] Question: Does the paper discuss both potential positive societal impacts and negative societal impacts of the work performed?
    \item[] Answer:  \answerNA{}.
    \item[] Justification: No clear societal impact to discuss.
    
\item {\bf Safeguards}
    \item[] Question: Does the paper describe safeguards that have been put in place for responsible release of data or models that have a high risk for misuse (e.g., pre-trained language models, image generators, or scraped datasets)?
    \item[] Answer: \answerNA{}.
    \item[] Justification: Our paper does not pose such risks.

\item {\bf Licenses for existing assets}
    \item[] Question: Are the creators or original owners of assets (e.g., code, data, models), used in the paper, properly credited and are the license and terms of use explicitly mentioned and properly respected?
    \item[] Answer: \answerYes{}
    \item[] Justification: CelebA and ImageNet are both available for non-commercial research purposes and are properly cited.

\item {\bf New assets}
    \item[] Question: Are new assets introduced in the paper well documented and is the documentation provided alongside the assets?
    \item[] Answer:  \answerYes{}
    \item[] Justification: All the models will be released upon acceptance.

\item {\bf Crowdsourcing and research with human subjects}
    \item[] Question: For crowdsourcing experiments and research with human subjects, does the paper include the full text of instructions given to participants and screenshots, if applicable, as well as details about compensation (if any)? 
    \item[] Answer: \answerNA{}
    \item[] Justification: this paper does not involve human subjects or crowdsourcing.

\item {\bf Institutional review board (IRB) approvals or equivalent for research with human subjects}
    \item[] Question: Does the paper describe potential risks incurred by study participants, whether such risks were disclosed to the subjects, and whether Institutional Review Board (IRB) approvals (or an equivalent approval/review based on the requirements of your country or institution) were obtained?
    \item[] Answer: \answerNA{}
    \item[] Justification: this paper does not involve human subjects or crowdsourcing.

\item {\bf Declaration of LLM usage}
    \item[] Question: Does the paper describe the usage of LLMs if it is an important, original, or non-standard component of the core methods in this research? Note that if the LLM is used only for writing, editing, or formatting purposes and does \emph{not} impact the core methodology, scientific rigor, or originality of the research, declaration is not required.
    \item[] Answer: \answerNA{}
    \item[] Justification: this paper does not involve any LLMs.

\end{enumerate}

\end{document}